\documentclass{sig-alternate-05-2015}

\begin{document}

\setcopyright{acmcopyright}
\doi{10.475/123_4}

\isbn{123-4567-24-567/08/06}

\conferenceinfo{PLDI '13}{June 16--19, 2013, Seattle, WA, USA}

\acmPrice{\$15.00}

%
\conferenceinfo{WIMS}{'2016 N\^{i}mes, France}

\title{Adaptive Task Assignment in Online~Learning~Environments}
%
%
%
%
%

%
\numberofauthors{4} 
\author{
\alignauthor
Per-Arne Andersen\\
       \affaddr{Department of ICT}\\
       \affaddr{University of Agder}\\
       \affaddr{Grimstad, Norway}\\
\alignauthor
Christian Kr\r{a}kevik\\
       \affaddr{Department of ICT}\\
       \affaddr{University of Agder}\\
       \affaddr{Grimstad, Norway}\\
\and 
\alignauthor
Morten Goodwin\\
       \affaddr{Department of ICT}\\
       \affaddr{University of Agder}\\
       \affaddr{Grimstad, Norway}\\
       \email{morten.goodwin@uia.no}\\
\alignauthor
Anis Yazidi\\
       \affaddr{Department of Computer Science}\\
       \affaddr{Oslo and Akershus University College}\\
       \affaddr{Oslo, Norway}\\
       \email{anis.yazidi@hioa.no}\\
}



\maketitle
\begin{abstract}

With the increasing popularity of \emph{online learning}, intelligent tutoring systems are regaining increased attention.
In this paper, we introduce adaptive algorithms for personalized assignment of learning tasks to student so that to improve his performance in online learning environments. As main contribution of this paper, we propose a a novel Skill-Based Task Selector (SBTS) algorithm which is able to approximate a student's skill level based on his performance and consequently suggest adequate assignments. The SBTS is inspired by the class of multi-armed bandit algorithms. However, in contrast to standard multi-armed bandit approaches, the SBTS aims at acquiring two criteria related to student learning, namely: which topics should the student work on, and what level of difficulty should the task be. The SBTS centers on innovative reward and punishment schemes in a task and skill matrix based on the student behaviour.

To verify the algorithm, the complex student behaviour is modelled using a neighbour node selection approach based on empirical estimations of a students learning curve. The algorithm is evaluated with a practical scenario from a basic java programming course. 
The SBTS is able to quickly and accurately adapt to the composite student competency --- even with a multitude of student models. 
\end{abstract}

%
%

%
%

%
%
\printccsdesc


\keywords{Online Learning; Intelligent Tutoring System; Adaptive Learning}

\section{Introduction}

With the dramatic increase of the world population, \emph{online learning} is becoming a significant driving force in today's educational systems. It is common in many universities to find classes of several hundred students with only one teacher. This scarcity of resources makes personalised ``one to one" teaching challenging, or practically impossible. Students may struggle to fulfill their full potential because the assigned tasks are generic and not tailored to their specific needs and skill level. Several studies show that personalised learning is the key to increased fulfillment of potential \cite{plins}. A possible solution to the latter problem is resorting to the advances in \textit{AI} so that to personalize the teaching process. Loosely, \textit{AI} is defined as:
``\textit{The automation of activities that we associate with human thinking, activities such as decision-making, problem solving and learning" }(Belleman, 1978)\cite{aibook}.

With the developments and discovery of \textit{AI}, ideas about a learning system or tutoring system began to emerge. Some of the first mentions of an Intelligent Tutoring System (\textit{ITS}) dates back to 1982 \cite{itsold}, where D. Sleeman and J.S Brown discuss a system designed to help students reach their full potential in a limited amount of time. This gained a growing number of supporters, adding funding and research to the field. A few years later. Bloom et al. \cite{bloom19842} published a study demonstrating that individual tutoring is twice as effective as group teaching. This boosted the interest for researches to pursue working on the ultimate goal of providing a system as good as a teacher and to supply each student with his/her own virtual tutor \cite{aits}.

In recent years, several systems have been developed in all kinds of relevant areas. Computer technology has matured and the cost of hardware has decreased while throughput and efficiency have increased. Online teaching courses like \textit{Kahn Academy}, digital handin tools like \textit{Fronter} and plagiarism controls like \textit{Ephorus (Fronter)} have emerged. True \textit{ITS} also exist with open tools like \textit{Codeacademy} and other e-learning platforms.

An \textit{ITS} should ``provide immediate and customized instruction or feedback to learners" \cite{itsbook}. In this paper, we provide an algorithm that aspires to fulfil the latter statement for the purpose of task selection, namely the skill-based task selection (SBTS).

The proposed solution is inspired by the multi-armed bandit algorithm (hereby \textit{MAB}). In simple terms, MAB is a black-box algorithm that searches for the optimal strategy from a set of actions. 

The use of this algorithm enables the developers to retain from gathering data to create a model of the environment and student, as the \textit{MAB} is responsible for finding the optimal solution by itself in an online manner. 

\subsection{Context and motivation of our study}

Teaching programming to large student groups is challenging \cite{schulte2006teachers,milne2002difficulties,barr2015advice,jacobs2015experiences}. One of the reasons for this is the group diversity which typically leads to teaching methods fitting well for a subgroup of the students but not for the whole group. Additionally, the number of students makes it challenging for teachers to spend sufficient time with each student in order to facilitate good learning.

When students do a programming assignment in most universities, they work in an offline process. First, they upload their hand-in to a Learning Management System (LMS) and await approval from the teacher. This asynchronous environment has several challenges, namely that it: (1) Forces a long waiting period between student's work and feedback. (2) Makes the teacher do robot-like tasks such as correcting many assignments instead of spending quality time with the students. (3) The work has very little resemblance to the industry practices for which the universities are supposed to prepare students.

Test-driven development is the de facto standard for software development and testing in the industry today. This centers on writing small pieces of code to test the software, and in turn running the tests automatically to verify the correctness of ongoing and delivered products. T-FLIP\footnote{http://tflip.uia.no --- Partially funded by Norgesuniversitetet} is a pilot Norwegian Project that aims at, among others, making the industry like environment available when teaching programming.

The current research falls under T-Flip for enabling personalized learning at scale. As a case study we will consider a programming course at the University of Agder, Norway.
The University enrolls more than 150 yearly in a basic programming course called \textit{DAT-101}. 
The official description states that the goal of the project is to establish new tutoring technology for students enrolled in programming courses \cite{goodwin2016teaching}\cite{goodwin2015educating}. The current state of this project is a automated hand-in solution where students are able to push their assignments to a continuous integration software called \textit{bamboo}. Bamboo executes unit-tests on each of the  assignments and determines if the student has passed or not. In retrospect with an actual \textit{ITS} system, this can be interpreted as a part of the core. With a few new components this can be developed further and incorporated into a complete system. In its current state, the teacher needs to give tasks and make unit tests manually. 

This paper outlines the creation of an intelligent algorithm responsible for individual tutoring. The algorithm shall adapt to the individual student, giving the optimal tasks for each student in every case. The goal with such an algorithm is to fully maximize the learning potential of each individual student by giving the students the opportunity to learn in a personalized environment. There are several approaches for such an algorithm, and a few of them are discussed in this paper.


\section{State of Art}
In this section, relevant studies and papers are discussed to give the reader an overview over the current state-of-art. Several papers on this topic exist dating back several years. The literature reviewed in this section is limited to content published (preferably) after 2005. Some of the used literature is older, but this applies for content describing psychological theories or well defined machine learning techniques. 

There are several approaches to create an \textit{ITS}. In the most recent papers, we are presented with a mix of different artificial intelligence approaches to solve the problem. Common for most of the papers reviewed is the need for some basic components. One of these components is a way to model the student itself, the system needs to be able to adapt to the individual students and understand the different properties like learning-rate, previous experience and other variables. An approach for such a model \textit{(from now referred to as the \textbf{student model})} is represented in some form in every paper \cite{baynet}\cite{mabits}\cite{oopt}\cite{umap} reviewed.

The use of the student model in recent papers suggests that this approach is fairly common  in the field of \textit{ITS}. Even though the model itself is fairly common, the implementation varies greatly. As an example of this, we can review the work reported in \cite{mabits} where the authors resort to a combination of a student model and a cognitive model to create a tutoring model. The tutoring model is created as an alternative to the student model. With this approach, the authors try to eliminate the need for a strongly typed student model. The goal is to adapt to each different student with as little information as possible. The use of a \textit{MAB} algorithm enables the system to find the optimal learning trajectory for a specific student with as little information as possible. A disadvantage with this approach is that the authors assume that tasks should be carried out in an order. They assume that after task A1 either A2 or B1 need to follow. If a student moves to B1, they can never move ``back" to any A-task. This is in most cases a simplification of the learning process, since students should be able to work on several categories and repeat previous categories. As an example, just because a student has moved on from if-sentences to for-loops, it is not correct to say that they should never practice on if-sentences again.

The authors propose the use of \textit{POMDPs} \textit{(partially observable Markov decision process)} as an optimization of teaching. This method requires the system to assume all students learn in the same way. It is also stated that this approach can be optimal, but requires good student and cognitive models. In most cases these methods are based on \textit{KTM} \textit{(knowledge tracing-methods)} which tries to estimate student knowledge on existing parameters. In most cases the lack of data causes this form of modelling to be inaccurate. The paper also suggests the use of \textit{POMDPs} is mostly used on a population of students, not individuals, and this approach has been proven to be suboptimal in an \textit{ITS} setting \cite{iism}\cite{mabits}.

On the other hand, several improved versions of the \textit{KTM} exist. An example of this is the \textit{BKT} \textit{(bayesian knowledge tracing)} with skill-specific parameters for each student. There are strong indicators that \textit{BKT} models accounting for the student variance is superior to the \textit{BK} model \cite{bktbook}\cite{ibktm}. This partially eliminates the criticism given by \cite{mabits}.

A decent number of reports indicate that intrinsically motivated students perform better. This demands that a good \textit{ITS} keeps motivating the student trough the whole experience. Lumsden, Linda S et al. \cite{smtl} investigated the optimal strategy for motivation, and found that one of the main keystones for a motivational experience is task mastery. This is backed up by \cite{mabits} who proposes a solution where the student is presented with tasks that are neither too easy nor too hard, but slightly beyond their current abilities. The paper argues that this concords with theories of intrinsic motivation.

The use of \textit{optimal-difficulty} should work for most students, thus, in this paper, we want to build on this idea, and  add a learning-rate variable to catch students outside of the normal range. We propose a solution where each student starts with a predefined \textit{optimal-difficulty} based on the findings in \cite{mabits} which adapts over time and changes the learning-rate based on the student answers. Some students may be easier motivated with challenging tasks, and the overall learning outcome may be more effective for these students. On the other hand we find students struggling with the default or \textit{optimal-difficulty}. In these cases the learning-rate should be decreased, allowing these students to participate at a slower pace.

There are several possible routes to create an \textit{ITS}. We have looked at several candidates in this study, including \textit{multi armed bandits} \cite{mabits}, \textit{bayesian-networks} \cite{baynet} and \textit{neural-networks} \cite{alen}, each with its own advantages. As mentioned earlier the \textit{student model} is an important part of this \textit{ITS}. In the latter reviewed papers, the \textit{neural network} and \textit{bayesian-network} both relied on comprehensive student models, with a solid core of data to make assumptions and decisions. These systems are shown to be reliable and good, but comprehensive data models are required so as to achieve optimal operation \cite{mabits}. With the use of \textit{MAB} it is possible to eliminate the need for prior-knowledge about the students. The \textit{MAB} is efficient, and it requires a weaker link between the student and the cognitive model. As the participants in this study lack access to prior relevant data, some versions of \textit{MAB} seem to be a good candidate. 
The paper \cite{mabits} proposes a \textit{MAB} for seven to eight years old school-children learning to decompose numbers while manipulating money. There is no guarantee a similar approach is viable for use in the context of programming-exercises.

A limited number of available papers describe the use of \textit{ITS} in programming courses. As representative studies, we identified \textit{Java Sensei} \cite{alen} (sep 2015) and \textit{ASK-ELLE} \cite{askelle} (2012). Each using a different \textit{machine learning approach}. \textit{Java Sensei} uses a combination of \textit{neural-network} strategies and emotion sensors to register information and making decisions based on input. \textit{ASK-ELLE} \textit{ITS} utilizes a domain reasoner using a Haskel Compiler called \textit{Helios}, this compiler was developed to give feedback on wrong syntax. The system requires each student to complete a given task, but helps the student accomplish the tasks by giving hints and examples relevant to found error(s).

Even though a generic solution is presented in \cite{mabits} relying on \textit{MAB}, actual implementations for
such a system in the context of programming has not been found. In this study we want to propose a solution utilizing the current knowledge of \textit{MAB} to create a programming tutoring system.


\section{Approach}
The algorithm is based on the principles of the \textit{MAB} algorithm. It utilizes a punish / reward system to balance exploration and exploitation in order to find a suiting task for the student. The algorithm consists of the following components.
\begin{description}
  \item[Knowledge matrix] Contains information about student anticipated knowledge.
  \item[Punishment and rewards] All the functions concerning learning rate, punishment and rewards.
  \item[Task generation] Responsible for generating task-sets.
\end{description}

\begin{figure}[!ht]
        \centering
        \includegraphics[width=8cm]{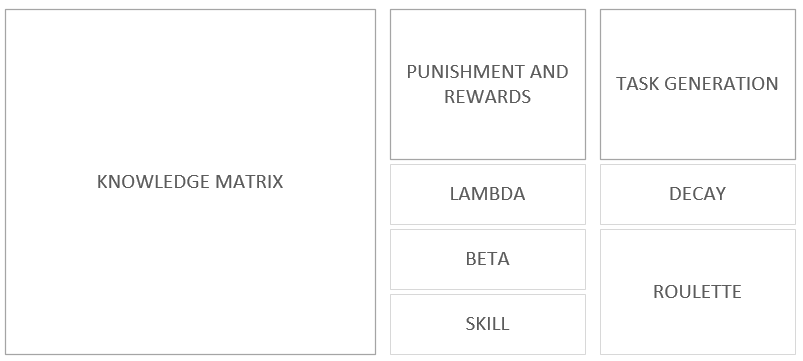}
        \caption{A visual representation of the algorithm components}
        \label{fig:Sys-overview}
    \end{figure}

\subsection{Skill-based Task Selection (SBTS)}
The Skill-based Task Selection (SBTS) aims at estimating which tasks a student should engage in.

The algorithm stores the student knowledge estimation in a task and skill matrix. The prototype of this algorithm has eight rows and ten columns in the matrix. Each row describes a topic in the context of programming. We chose some logical names for these rows: \\\textit{if, for, while, methods, classes, exceptions, gui and reflection}. A column represents the skill-level or difficulty of each individual topic. In this implementation each row-topic has ten different levels of difficulty. The topics are sorted by difficulty, from \textit{if} to \textit{reflection}. It should be noted that the algorithm does not have any knowledge on what the topics \textit{if}, \textit{for}, $\dots$ \textit{reflection} are. It only has knowledge about the order, meaning that \textit{for} is a more advanced level than \textit{if}, and so on. This means that any other topics can be chosen without having an effect on the algorithm performance.

 Note that this is notably different than previous work using \textit{MAB} \cite{mabits} since our approach is probabilistic, it allows choosing all categories according to some probability in line with how confident we are that the student has mastered the area. In contrast to earlier work, SBTS is able to make the student revert back to if-sentences if the student needs more practice.

\begin{figure}[!ht]
    \centering
    \includegraphics[width=8cm]{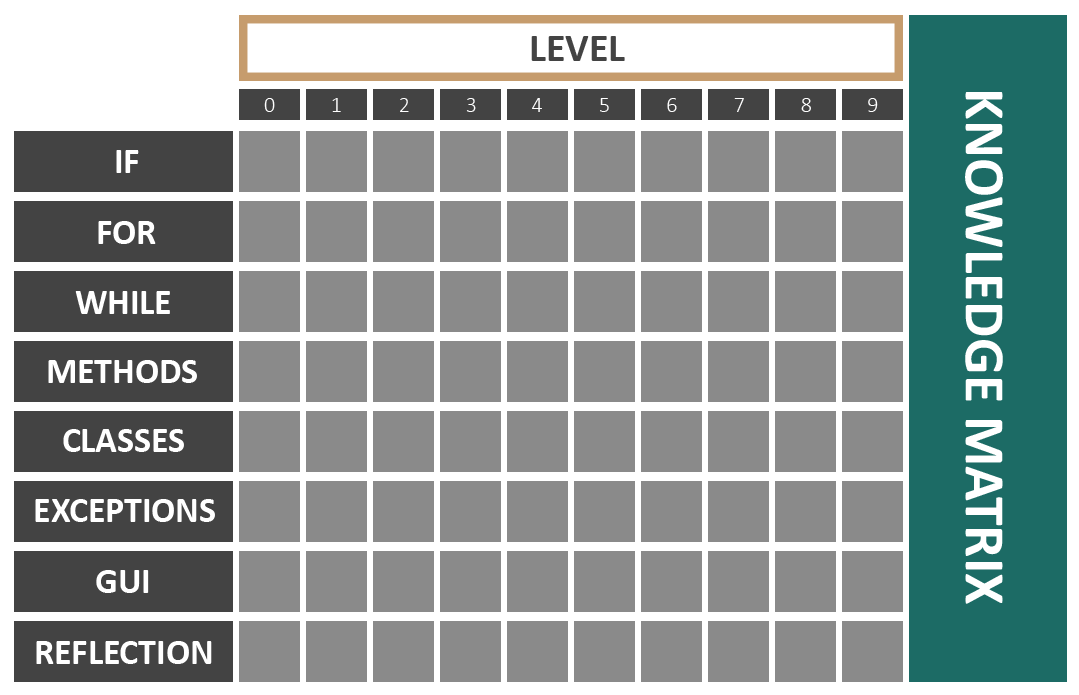}
    \caption{A visual representation of the matrix}
    \label{fig:matrix-overview}
\end{figure}

The knowledge matrix is the core of the algorithm, acting as the reference for every calculation done by rest of the algorithm logic. The goal is to lead the students progress from top to bottom through suited task selections. The algorithm should guide the student diagonally through the knowledge matrix.

Each cell in the knowledge matrix contains a probability value, this value indicates how relevant this task is for the student, higher probability increases the chance for the task to be selected. The elements of the matrix must always sum to \textbf{1.0}.

\subsection{Punishment and rewards}

The algorithm utilizes reward and punish functions in order to adapt to each individual student. This is done by decreasing or increasing the percentage value in each cell in the knowledge matrix. The cell selection is based on a static pattern. This patten can be configured to include additional cells in either direction. By default, the algorithm includes the selected cell and in addition, a single neighbor cell in the X and Y axis. For the reward function, these cells are on the right hand side of the matrix, while punish includes the left hand side cells.

\begin{figure}[!ht]
        \centering
        \includegraphics[width=8cm]{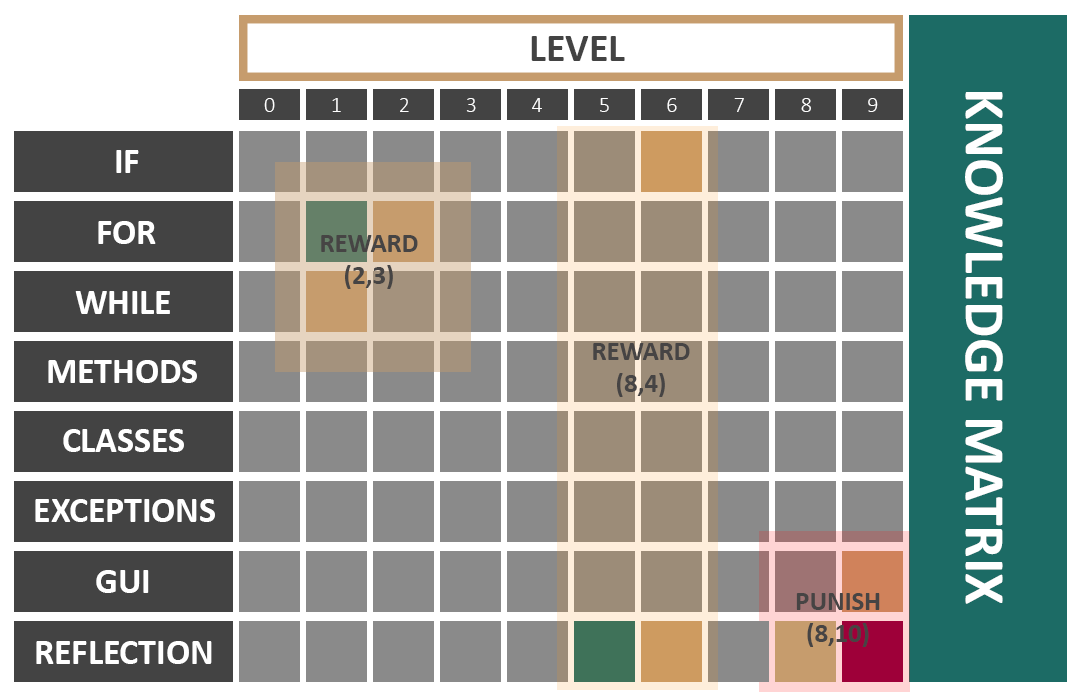}
        \caption{A visual representation of punish and reward}
        \label{fig:matrix-overview}
\end{figure}

Figure \ref{fig:matrix-overview} shows a visual representation of the behaviour in case of reward and punish. The algorithm pushes the student ``down and right'' during reward, and ``up and left'' during punish. This behaviour forces the learning-pattern into a diagonally shaped pattern from (0,IF) to (9,REFLECTION).

\subsection{Skill level and adaption}
A key feature of the algorithm is the adaption process. The algorithm should behave differently from student to student. This is defined with the use of independent skill levels. The skill level is an approximation of the students knowledge, and is represented as a decimal value. The variable is used for the calculation of reward and punishment severity. If a student tries to solve a hard task the outcome would be a much higher increase in progress, compared to solving an easier task.

The skill level is defined by the progress in the knowledge matrix, and can be interpreted as the distribution of percentage in the matrix.

\begin{figure}[!ht]
        \centering
        \includegraphics[width=8cm]{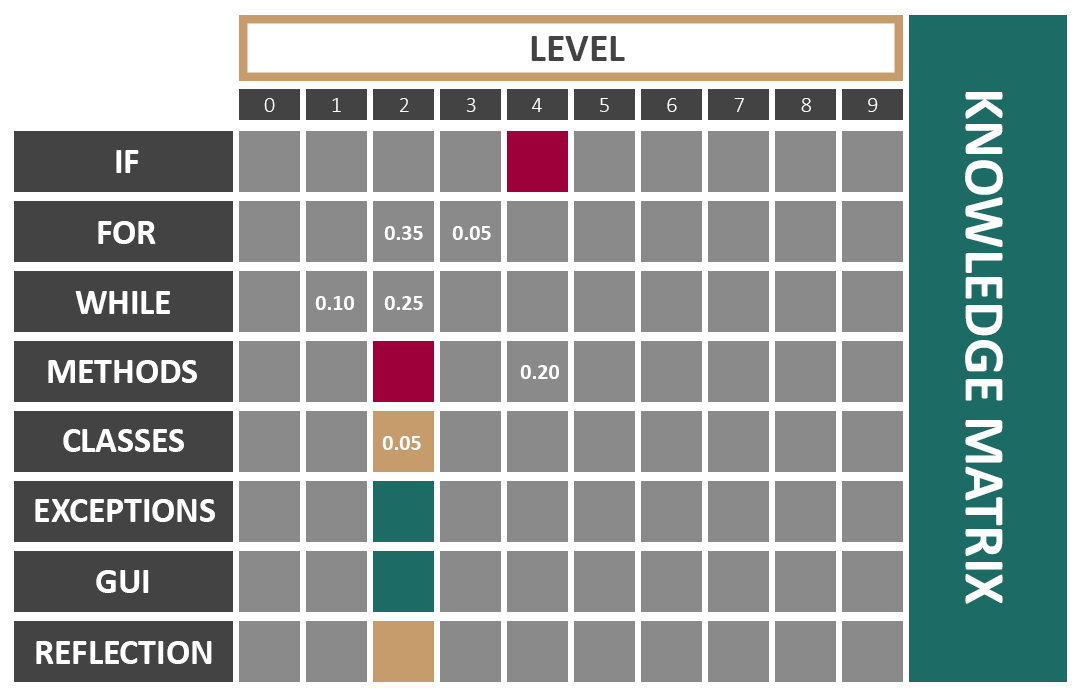}
        \caption{Skill level calculation}
        \label{fig:skill-level}
    \end{figure}



In the example above (Figure \ref{fig:skill-level}), the colored area represent the expected knowledge area based on the current skill level. Green cells represent the best selection, with less relevant tasks to both ends. 
It is also possible for the algorithm to inaccurately calculate the skill level. In the above case, the skill level is considerably higher than the best selection. This is caused by cell (4, METHODS) which has a 20 percent chance of occurring, hence increasing the skill level.
The algorithm handles such problems with an exponential function to either give a large reward or punish depending on the student's answer.

\subsubsection{Beta}
The $\beta$ constant is responsible for providing the right amount of penalty or reward based on the current task. Each task has an independent task-skill. This value represents the expected skill level required to solve the task. 

\begin{equation}
cell\_skill = \frac{cell\_column + 1}{len(num\_rows)} \times row\_number
\end{equation}

With this independent skill requirement, it is possible to calculate different rewards or punishment based on how hard the task is opposed to the current approximated skill level of the student.

     \begin{center}
            $\beta = 1 * x^2 + 0.5$ \\
            \tiny{\textit{$x = task\_skill - user\_skill$}}
    \end{center}

    \begin{figure}[!ht]
        \centering
        \includegraphics[width=8cm]{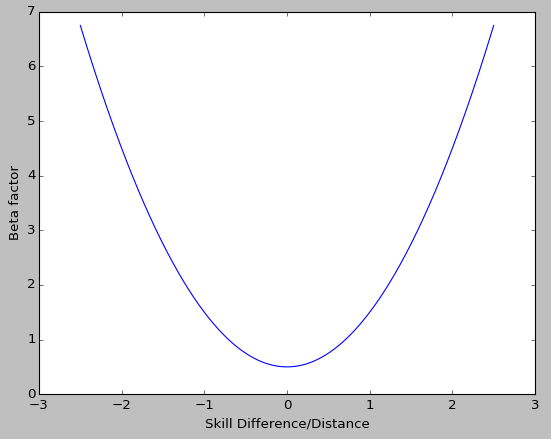}
        \caption{Beta factor determination}
        \label{fig:beta-factor}
    \end{figure}

This formula is selected because of its characteristic paraboloid shape, we will always get a positive number on either side of the graph, where a big difference between task\_skill and user\_skill yields a high reward, or a high punishment. This ensures that the algorithm adapts quickly to good students, and punish quickly if the tasks are too hard.

\subsubsection{Lambda}
The $\lambda$ is a constant used as a learning-speed variable. This variable can take any value between 0.0-1.0. Higher values yield a higher matrix percentage reward / punishment. The $\lambda$ is the last piece of the complete reward / punish formula.

        \begin{center}
            $ new\_prob = old\_prob + (\lambda * \beta * -old\_prob)$ \\
            diff = $(old\_prob - new\_prob) / len(reward\_cells)$\\
            
            \tiny{\textit{Punishment generation}}
        \end{center}

We punish the selected cell by $new\_prob$ and proceed by giving either the left or right neighbour cells a new probability based on the answer (correct / wrong) as in Figure \ref{fig:skill-level}. The sum given is the \textit{diff} variable shown in the above equation.

With these four different variables, it is possible to tune the algorithm to fit a range of different learning scenarios. A slower rate (lower values) would cause the algorithm to demand a higher accomplishment rate or more tasks to be solved before finishing, or reaching level ten.

\subsubsection{Task generation}
In our solution, each task is a part of a task-set. Each task-set contains ten tasks. This makes a task-set similar to a test. The generation of a set is done with the help of the knowledge-matrix, higher percentage topics (cells) give a higher chance to be chosen. A decay function is also implemented, reducing the chances for the same topic to be selected twice or more in a single task-set. The selection phase is done by the help of a random variable from \textit{0.0} to \textit{1.0}, this value is then checked against every cell with a value higher than \textit{0}. If the random value is lower than the cell value, the task is selected and added to the task-set.

\textbf{For reference}: It does not matter if the student gets a single task, or a task-set. The algorithm handles both scenarios.

\subsection{Environment}
The environment is needed to create an accurate student simulation. This can be considered as the environment the SBTS needs to interact with.

The student is our reference for testing. We have three different implementations, each with a unique approach. The student solves a dynamic amount of tasks which produces a graph with information showing how far he or she came. In most tests this graph includes several students, and many iterations so that to get an average smooth graph. 

     \begin{figure}[!ht]
            \centering
            \includegraphics[width=8cm]{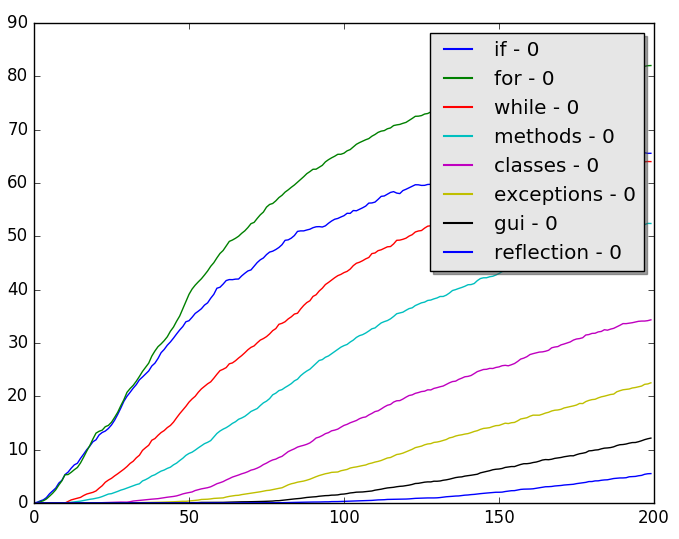}
            \caption{Overview of a single graph for a skill-level}
            \label{fig:single-graph}
        \end{figure}
        
The graph above depicts the success of 1000 students for 200 task-sets iterated 1000 times. The graph represents skill \textit{0} from the knowledge matrix. This means that it only covers the average learning success for each topic on level zero. To get an overview of all the levels we need to look at nine of these graphs, one for each category. This is done because it makes it possible to identify how the students performed in each category, and how far most of the students reached in the skill tree from zero trough seven. 

\subsubsection{Student skill}
In the non-static versions of the student implementation, a similar matrix like the knowledge-matrix is used. This matrix describes the student-actual knowledge on each topic. This value is used by the student-AI to determine if the student should pass, or not pass a given topic. 

Each topic gets a value between \textit{0.0} and \textit{1.0}, and describes the current state of student knowledge. If no knowledge is present, the student utilizes the exploration method to get a percentage chance to pass the task. If the task is passed, the knowledge value increases like in the knowledge-matrix.


\subsubsection{Static}
The static student has a constant probability $p$ of rewarding or punishing a task. This is the most primitive student model possible, but can give insight in how the algorithm reacts to the different types of students.

Comparing the static model to a real life example is acceptable as it could be used to calculate percentage of correct answers for all students at the end of the semester, however there would be a difference in how the student progressed through the graph. This however should not matter as 80\% correct answers should equal 80\% in the final grading.

\subsubsection{Static epsilon}
This method utilizes a \textbf{MAB} which uses the epsilon greedy strategy. Epsilon greedy is a value which may or may not change during the tests. A random value is generated, and it is evaluated against the epsilon value. If the random value is lower than epsilon, exploration will be done, if not, exploitation.

The static epsilon never changes its value which is beneficial to see how a student is doing based on the expected competence.
However, this is not applicable for a real-world test as students normally progress when doing tasks (Learning).

\subsubsection{Dynamic}
This algorithm uses MAB with the epsilon greedy strategy as well, but it decreases its value per iteration.
The idea behind this strategy is to get as much information about the student as possible in the initial phase of the algorithm, and later on exploit this knowledge.

This is more like a real-world student, but may come to fall short after the exploration face is done. This is because the strategy does not take into account that the student may learn faster than the algorithm expects.

The epsilon value is set to 70\% at the beginning and decreases exponentially, and will after 100 iterations only use exploitation. This means that a student must do 100 tasks in order for the algorithm to stop ``guessing" which are the best set of tasks for the selected student.

    
    \section{Discussion}
   In this section,  our findings and results are presented. These is a combination of graphs and empirical values showing that our algorithm is capable of providing a student with incremental tasks, and ``learn" the AI to reach different skill levels. Putting it differently, how well is the SBTS able to adapt to the complex environment?
   

    The first priority of the testing phase was to validate the algorithm. To do this we conducted a test
    graphing the progress for a student trough the whole matrix. The result of this test is depicted below. This test shows relation between skill-level and the success percentage of the student. We can validate the algorithm using this, because it shows a growth in the skill-level depending on the chance of of student's success.
    
        \begin{figure}[!ht]
            \centering
            \includegraphics[width=8 cm]{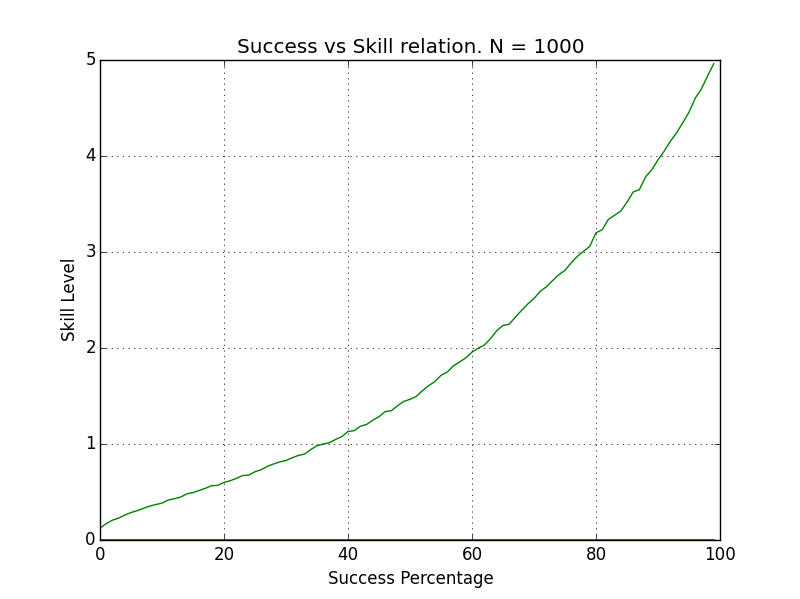}
            \caption{Beta factor determination}
            \label{fig:beta-factor}
        \end{figure}

    The different percentages coheres well with a grade scale. For example the higher difficulty tiers \textit{4-5} require an average success percentage of \textit{80-100}. Notice that this graph only includes ten task-sets (100 tasks). This is the reason for the limited y-axis.
    
    

    \subsection{Case 1 - Static Environment}
     \textbf{Parameters - 1000 students, 200 task-sets, 100 iterations}

    The static implementation is rather simple. It utilizes a common random generation of a double between 0.0 and 1.0. This number is compared to a static variable, and if the random number is higher than the static number the student fails. This gave us the possibility to explore how far a good student and a bad student would progress without any interference from each other. Note that this test does not account for experience gained when doing the tasks, and should only be used as a reference for how far this particular student can come based on a percentage. It also gives us the possibility to check how the algorithm behaves on a static model. Two graphs are presented, one with a static percentage of \textit{70\%} (good student) and \textit{20\%} (bad student).
    
    Figure \ref{fig:Good-static-student} and \ref{fig:Bad-static-student} show the average of bad and good students. The students progress by moving from category to category and skill level to skill level. Each skill level is represented by a graph. This is in many ways a visual representation of Figure \ref{fig:matrix-overview} over time.

    As Figure \ref{fig:Good-static-student} and Figure \ref{fig:Bad-static-student} show, the highest percentage students progress the most, as expected. This test also shows that the algorithm behaves correctly, as the slopes are formed as an exponential increase through all the different graphs. The only difference is how far, and how quick the student reached the higher tiers of difficulty.

{\renewcommand{\arraystretch}{2.0}
\setlength{\textfloatsep}{10pt plus 1.0pt minus 2.0pt}
\begin{figure*}[ht]
\begin{tabular}{*{5}{c}}
  \includegraphics[width=35mm]{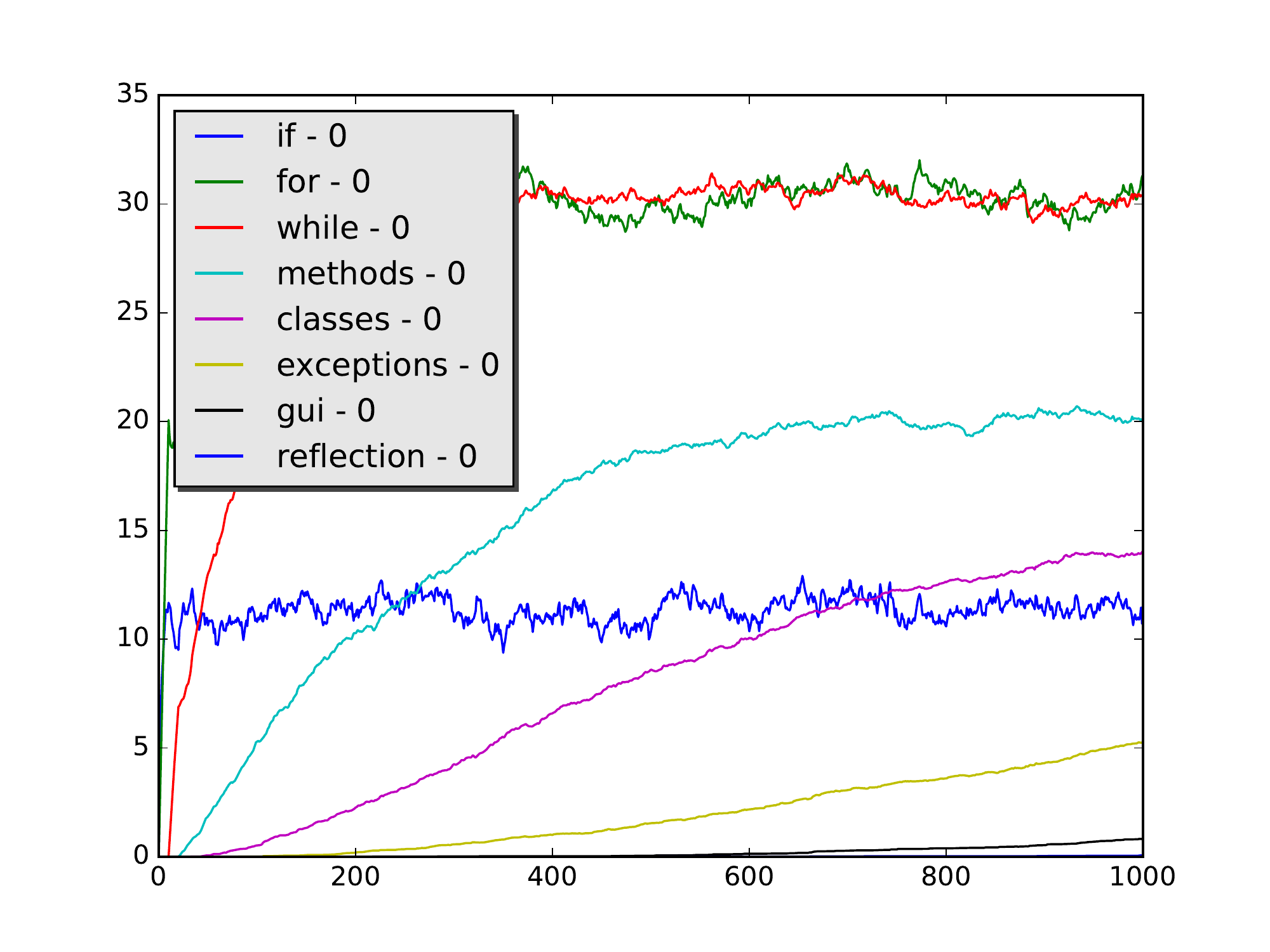} &
  \hspace{-1.6em}\includegraphics[width=35mm]{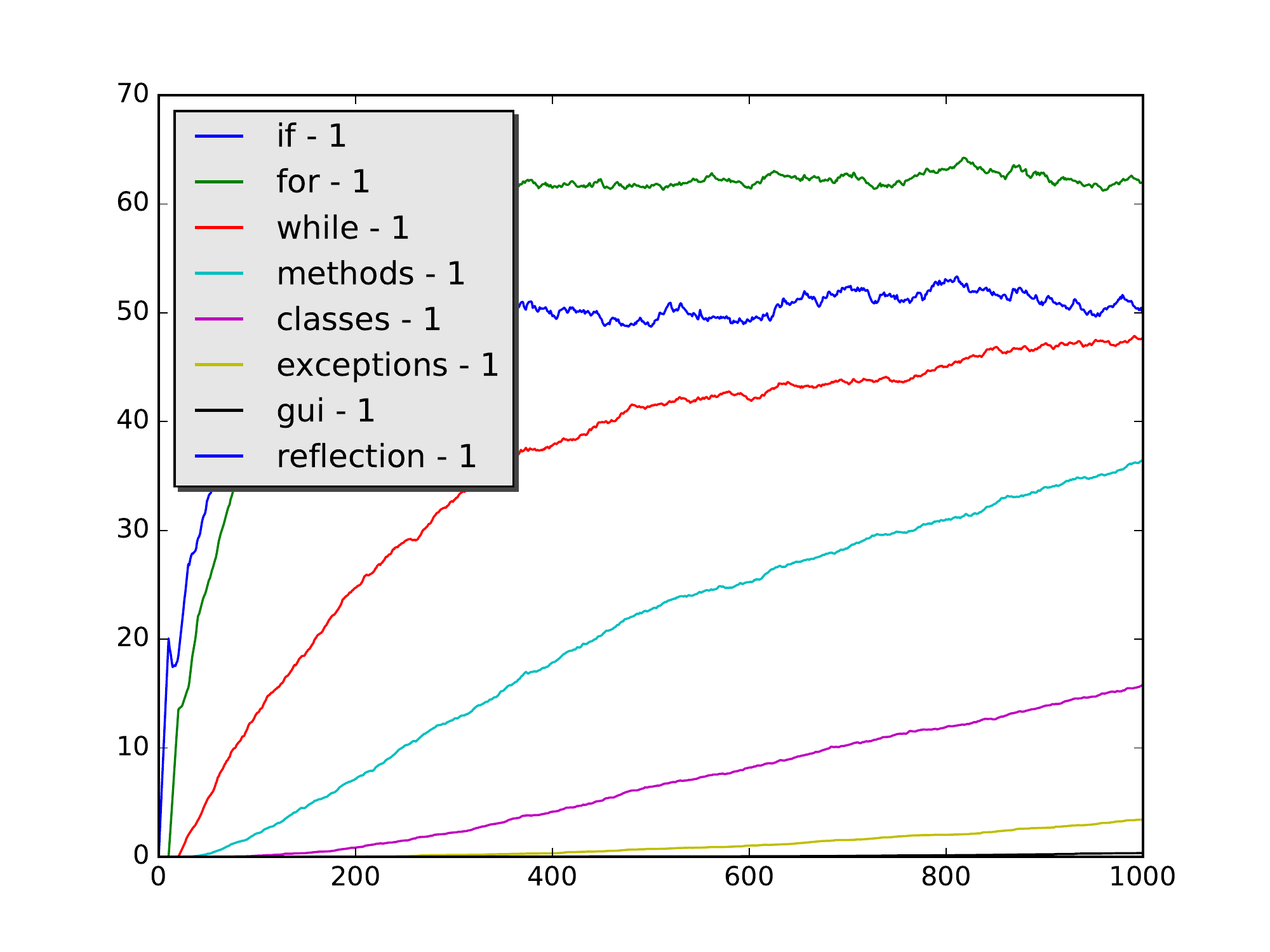} &
  \hspace{-1.6em}\includegraphics[width=35mm]{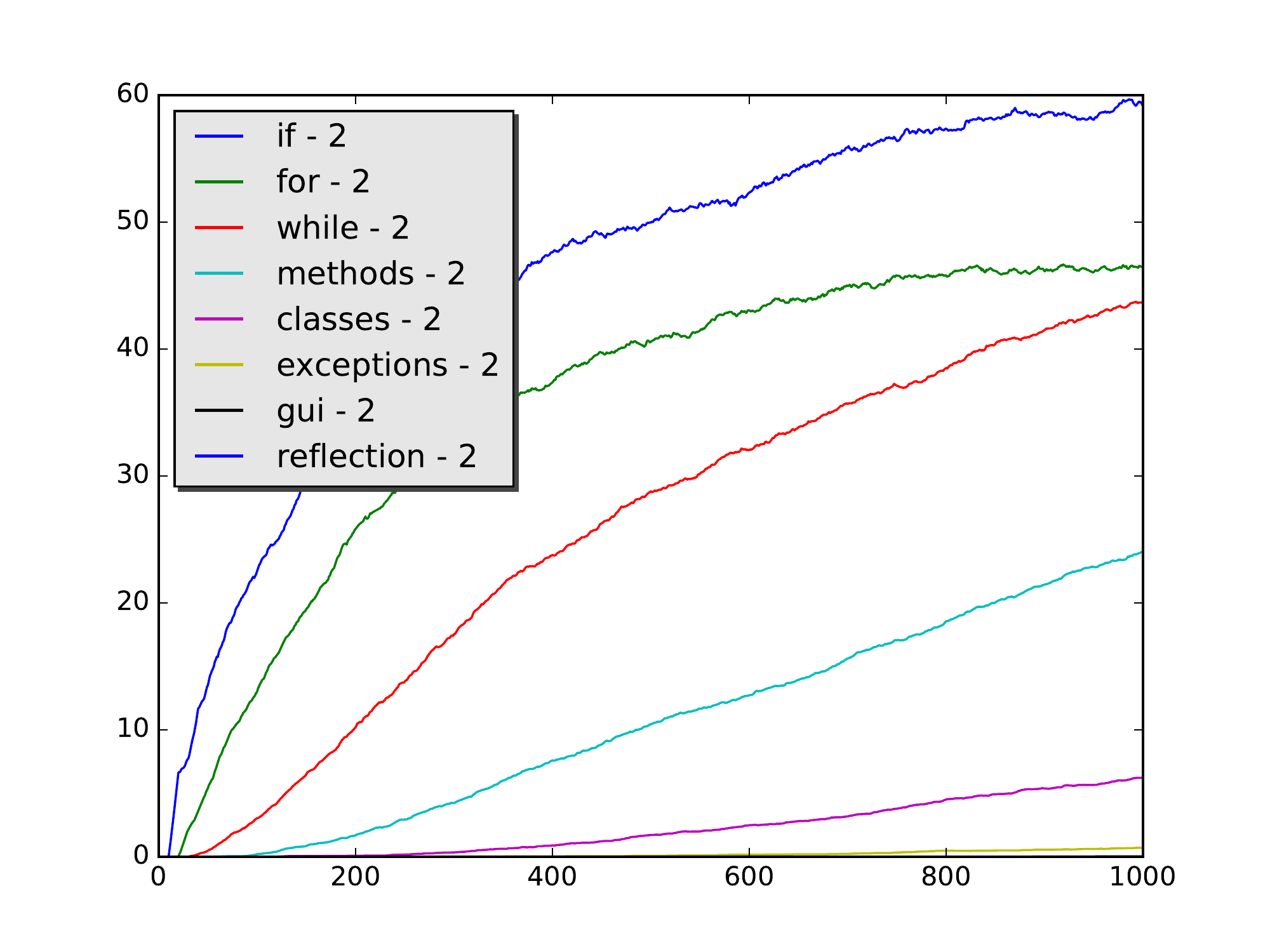} &
  \hspace{-1.6em}\includegraphics[width=35mm]{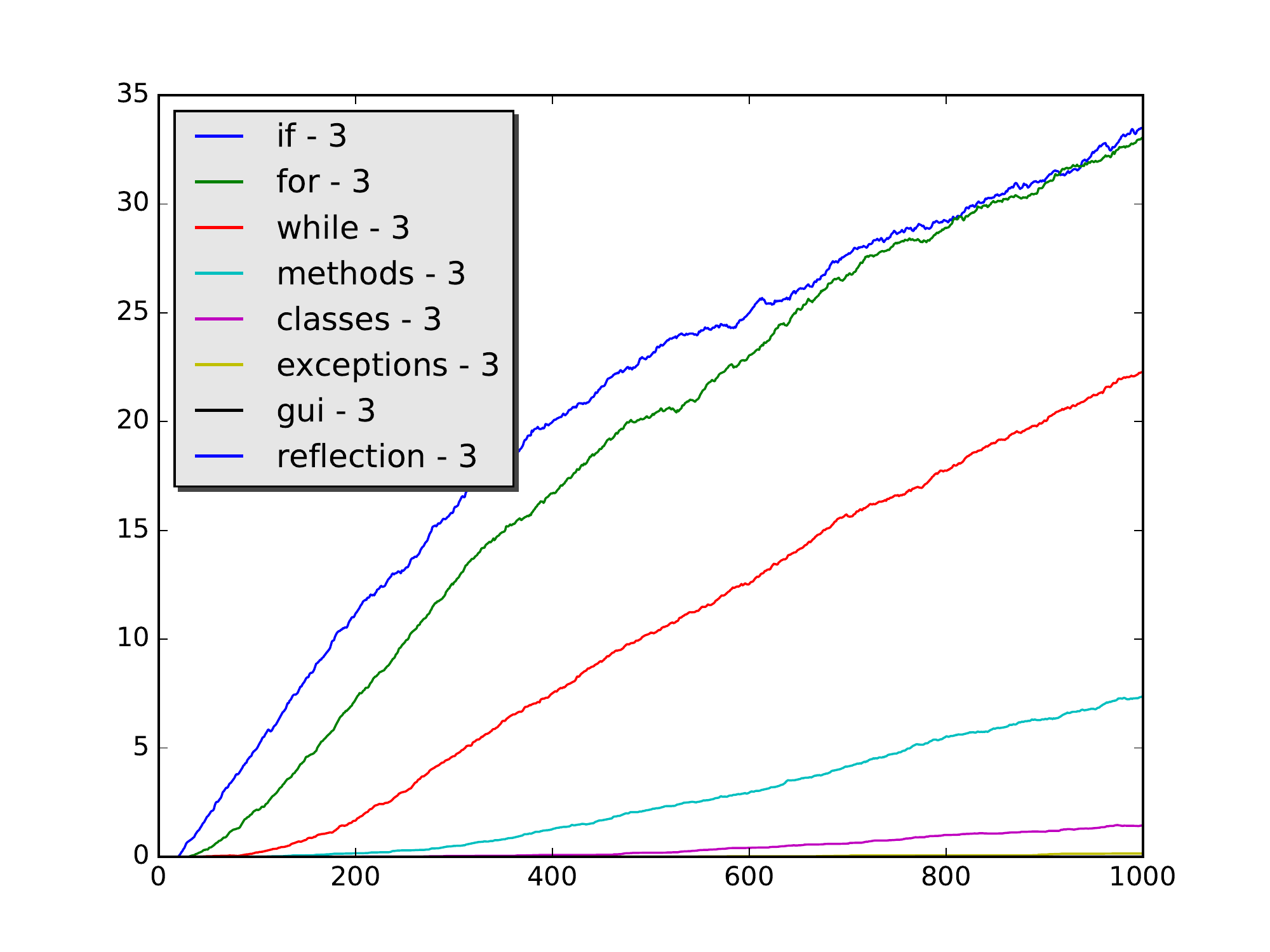} &
  \hspace{-1.6em}\includegraphics[width=35mm]{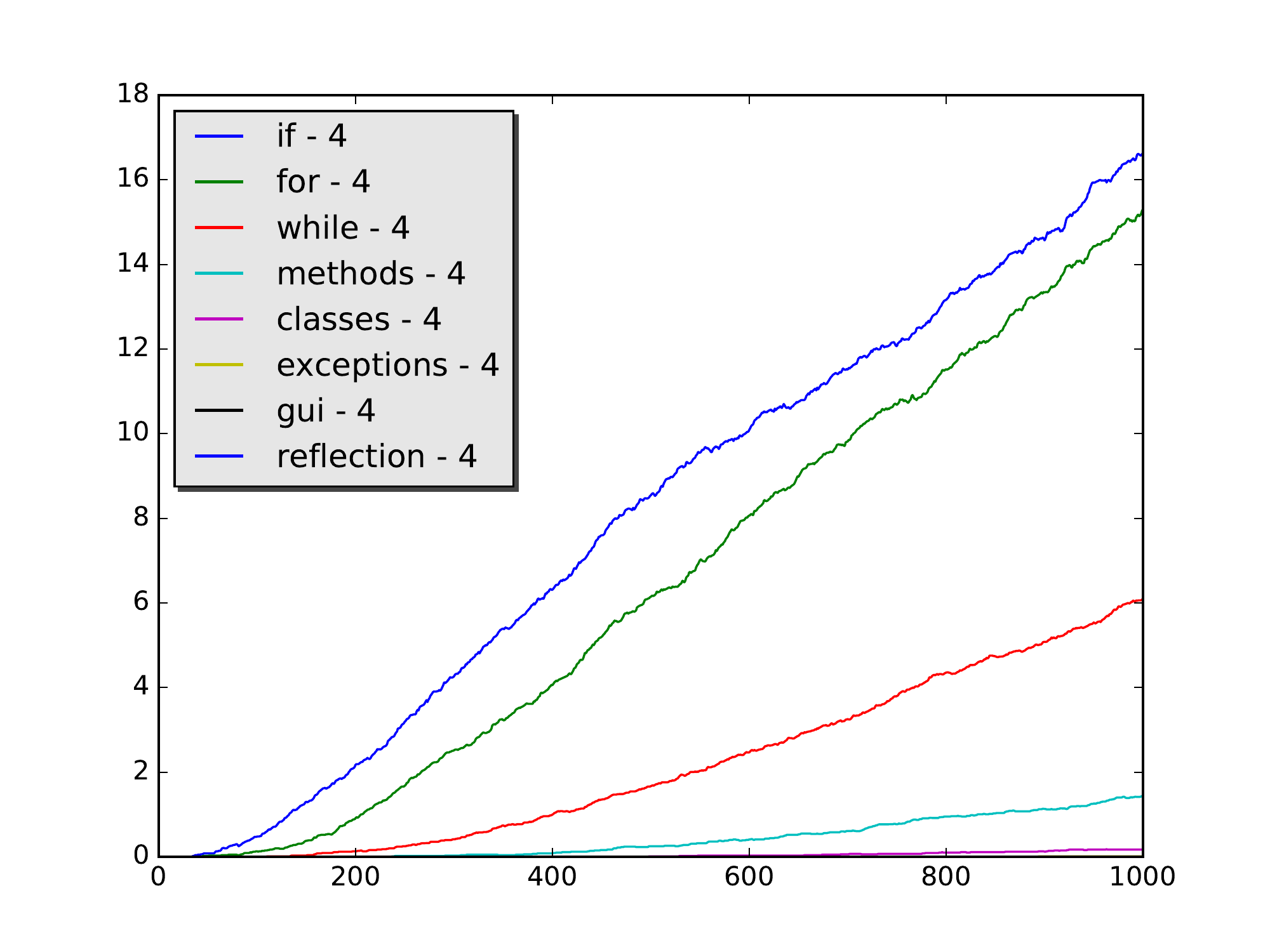} \\[-12pt]
  Level=1&
  \hspace{-1.6em} Level=2&
  \hspace{-1.6em} Level=3&
  \hspace{-1.6em} Level=4&
  \hspace{-1.6em} Level=5\\
  \includegraphics[width=35mm]{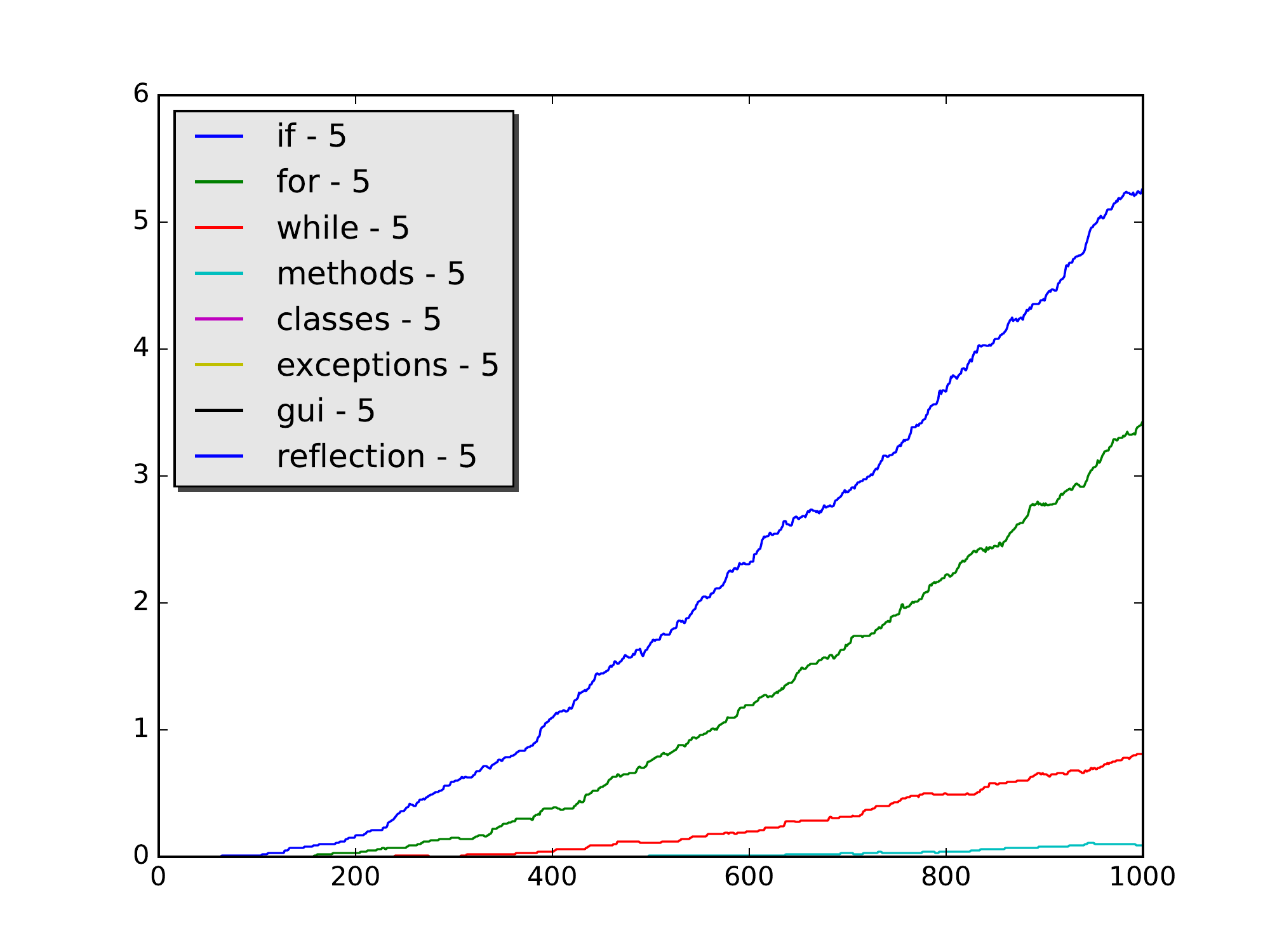} &
  \hspace{-1.6em}\includegraphics[width=35mm]{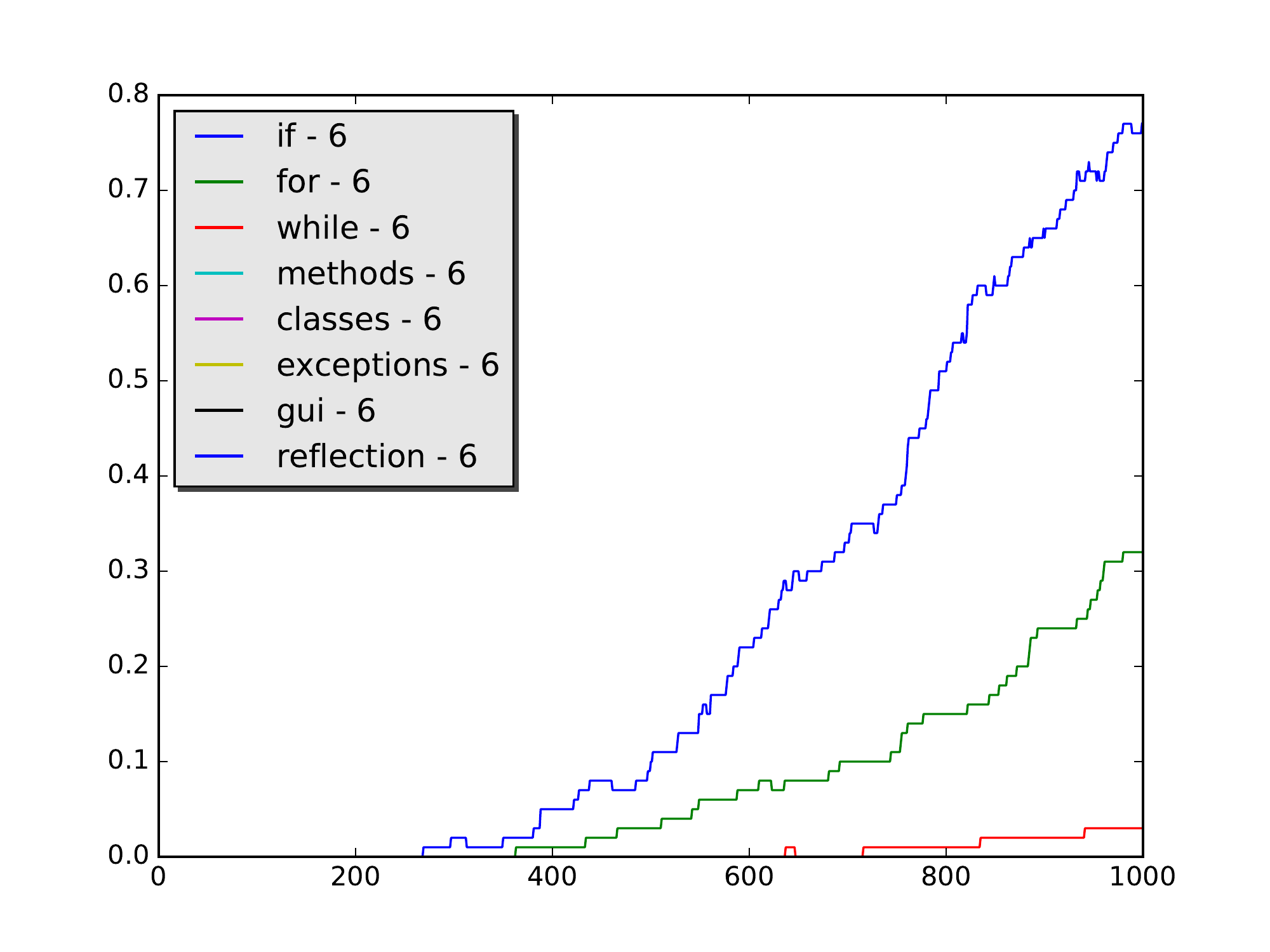} &
  \hspace{-1.6em}\includegraphics[width=35mm]{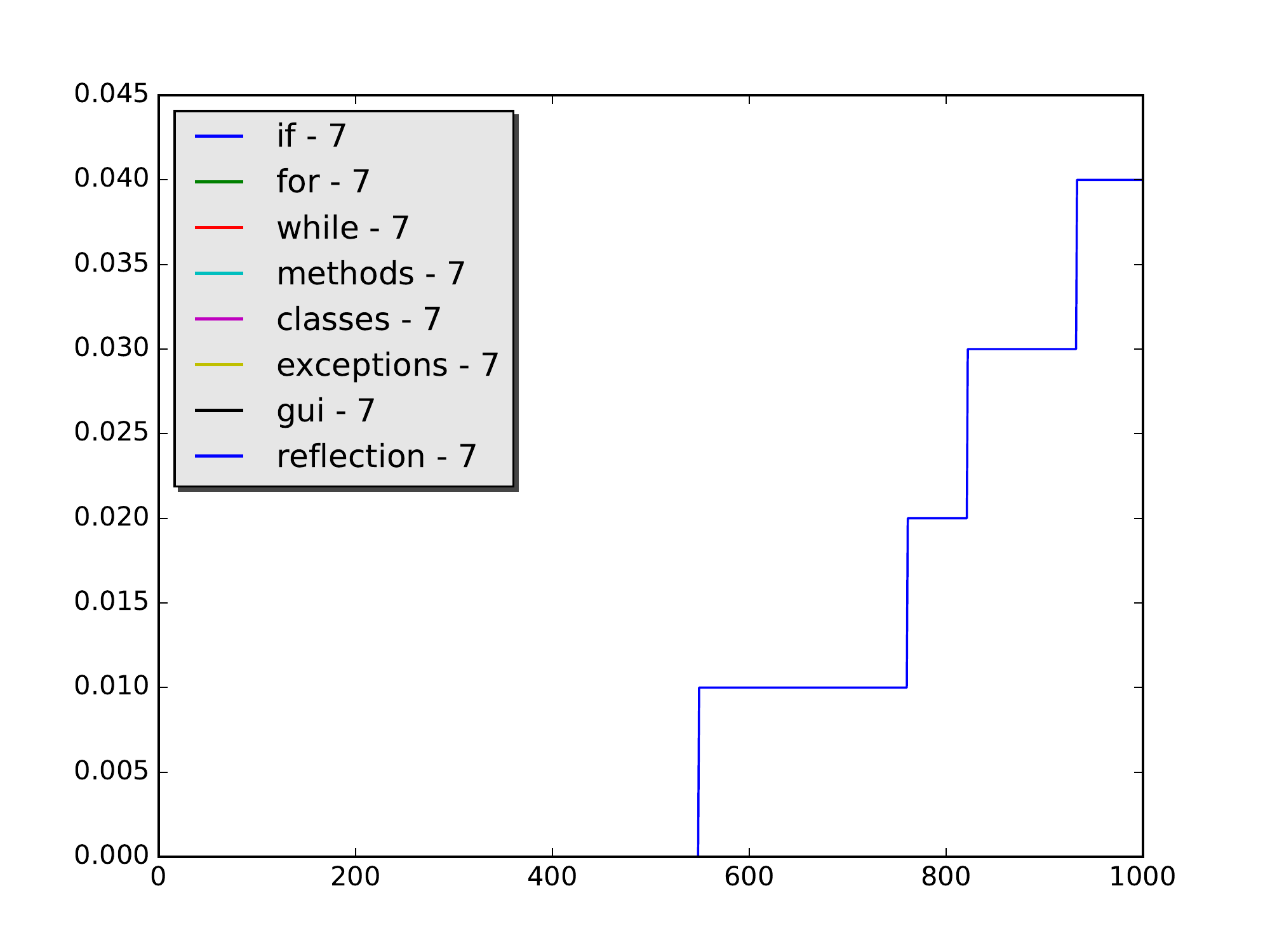} &
  \hspace{-1.6em}\includegraphics[width=35mm]{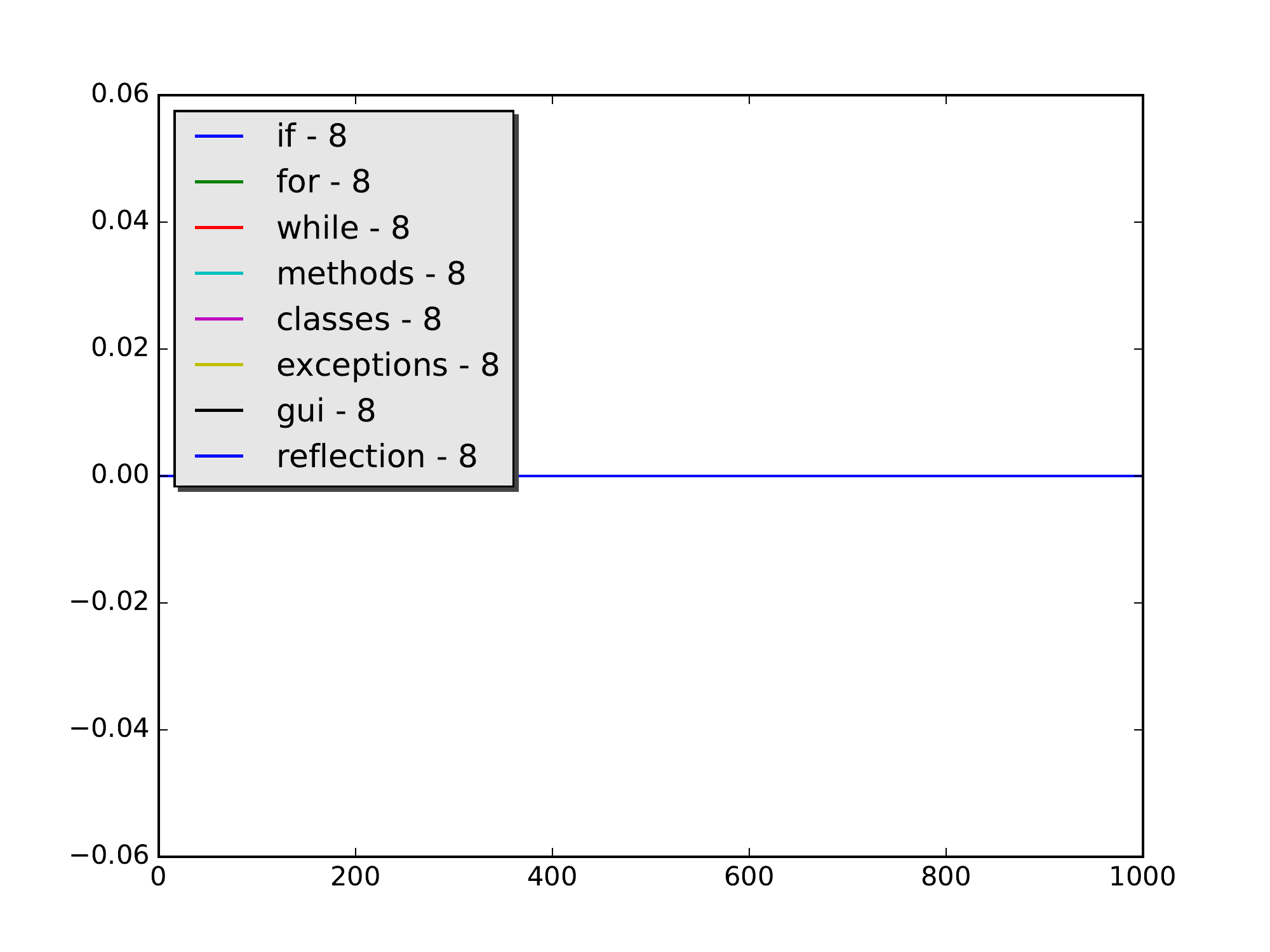} &
  \hspace{-1.6em}\includegraphics[width=35mm]{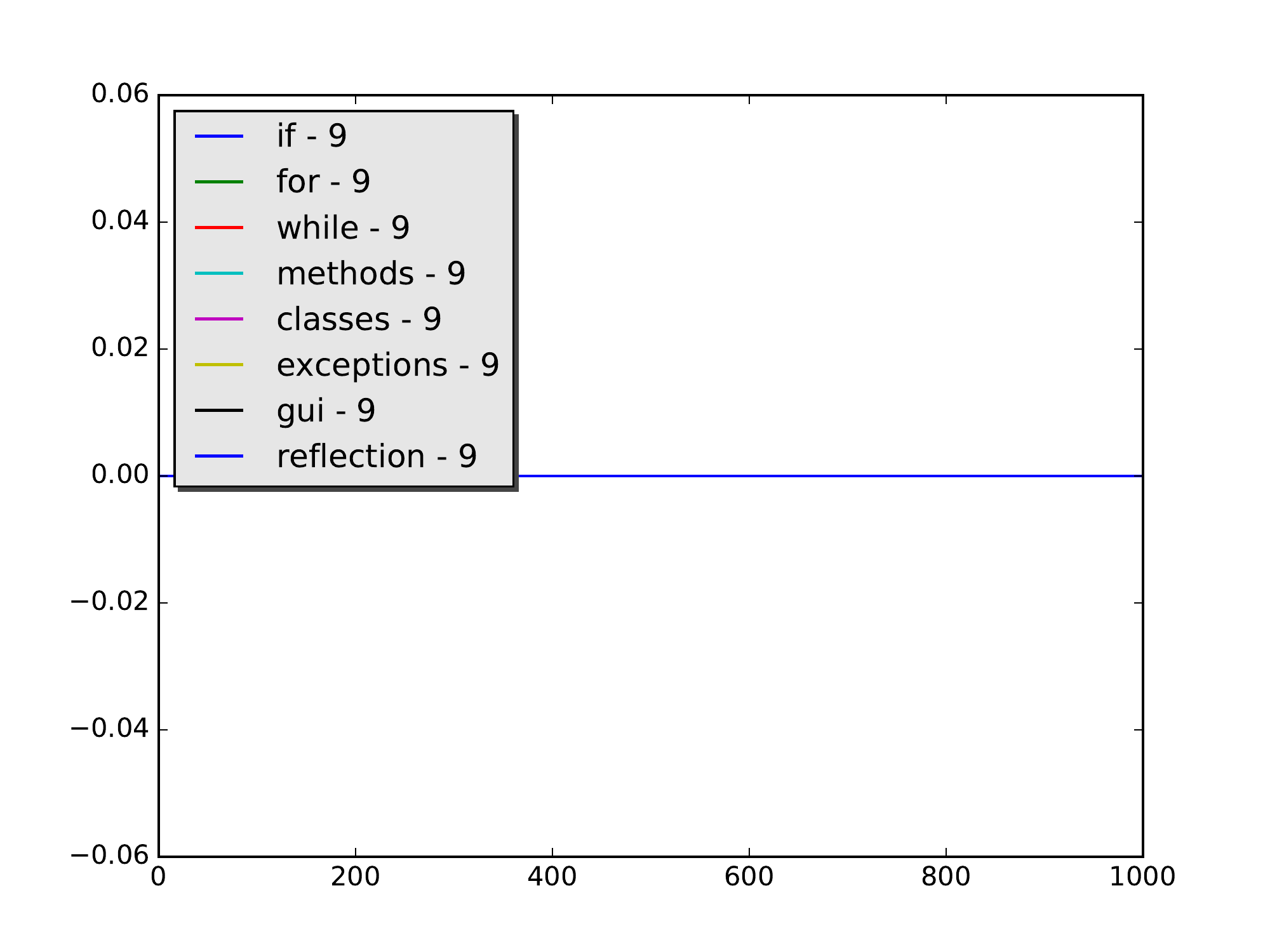} \\[-12pt]
  Level=6&
  \hspace{-1.6em} Level=7&
  \hspace{-1.6em} Level=8&
  \hspace{-1.6em} Level=9&
  \hspace{-1.6em} Level=10\\

\end{tabular}

\caption{Static Environment - Bad Student}
\label{fig:Bad-static-student}
\end{figure*}
}

{\renewcommand{\arraystretch}{2.0}
\setlength{\textfloatsep}{10pt plus 1.0pt minus 2.0pt}
\begin{figure*}[ht]
\begin{tabular}{*{5}{c}}
  \includegraphics[width=35mm]{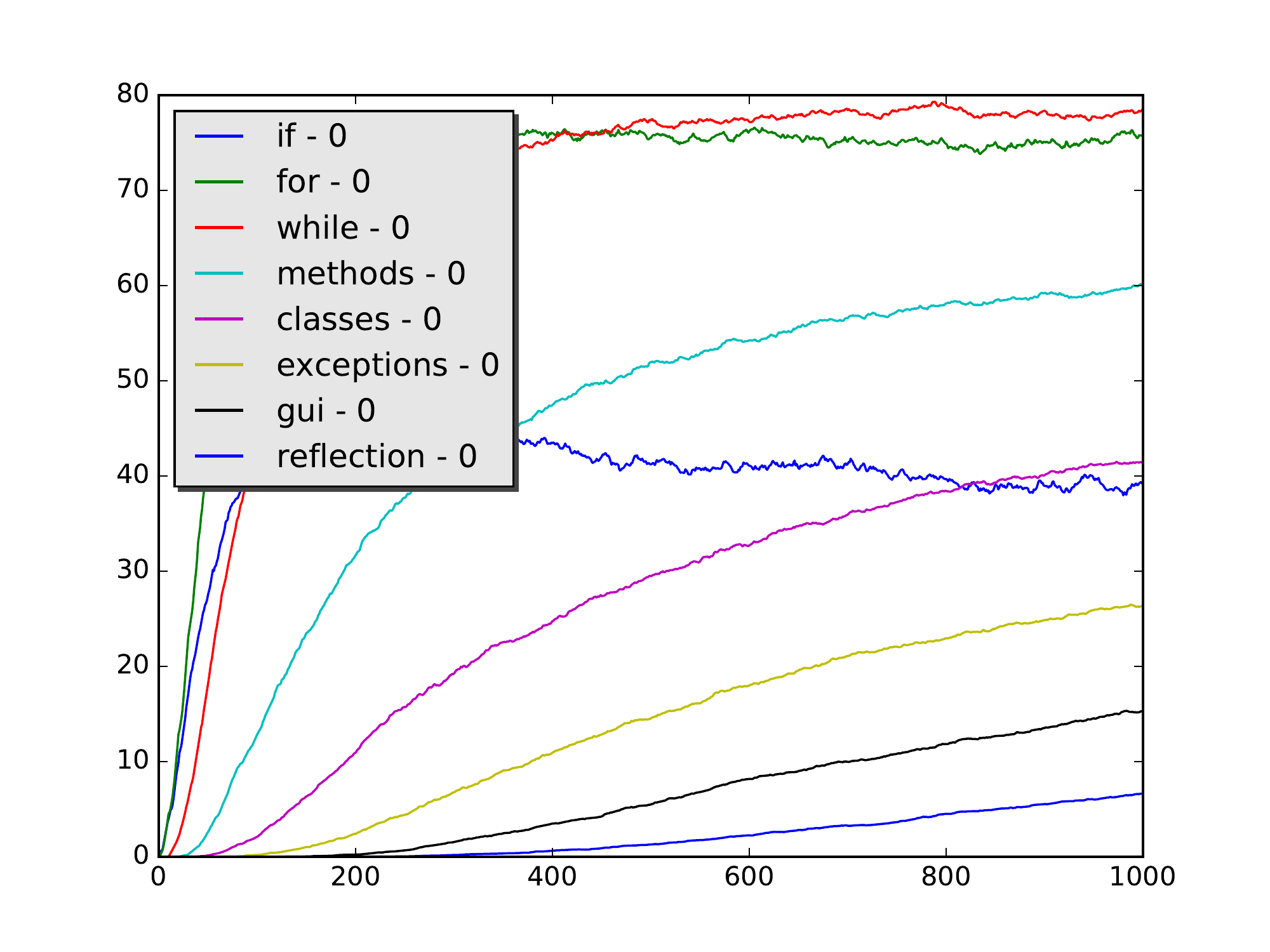} &
  \hspace{-1.6em}\includegraphics[width=35mm]{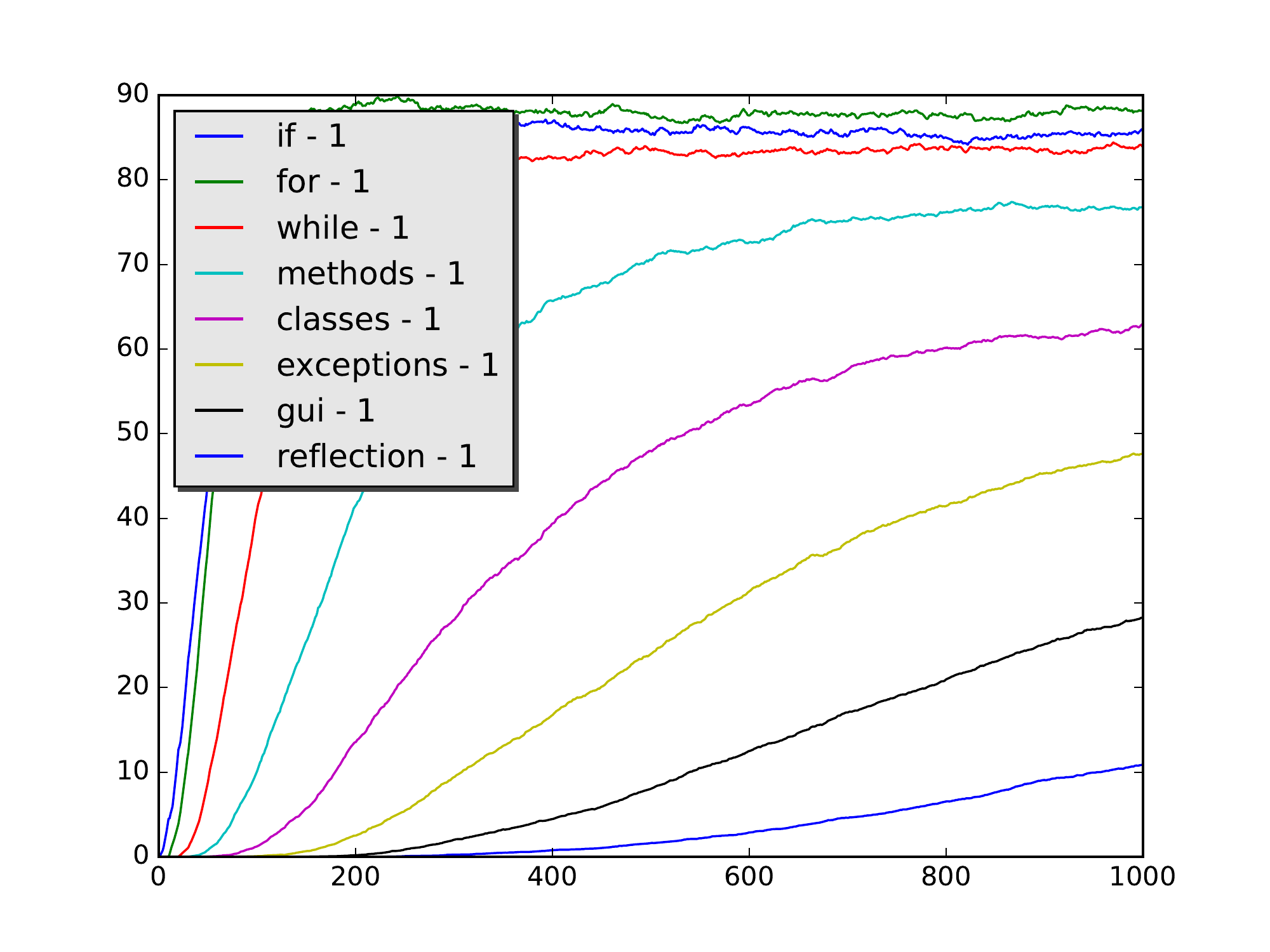} &
  \hspace{-1.6em}\includegraphics[width=35mm]{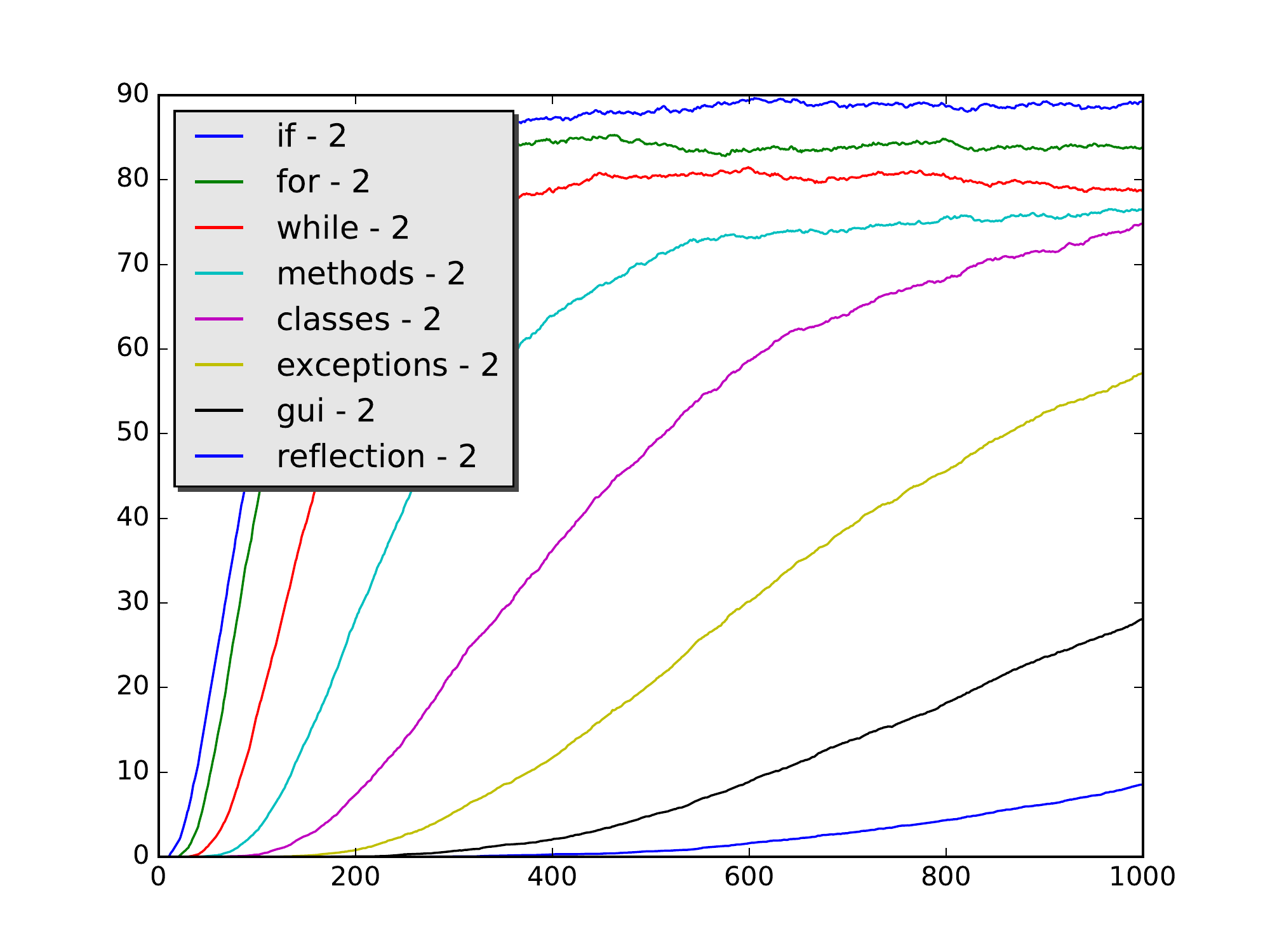} &
  \hspace{-1.6em}\includegraphics[width=35mm]{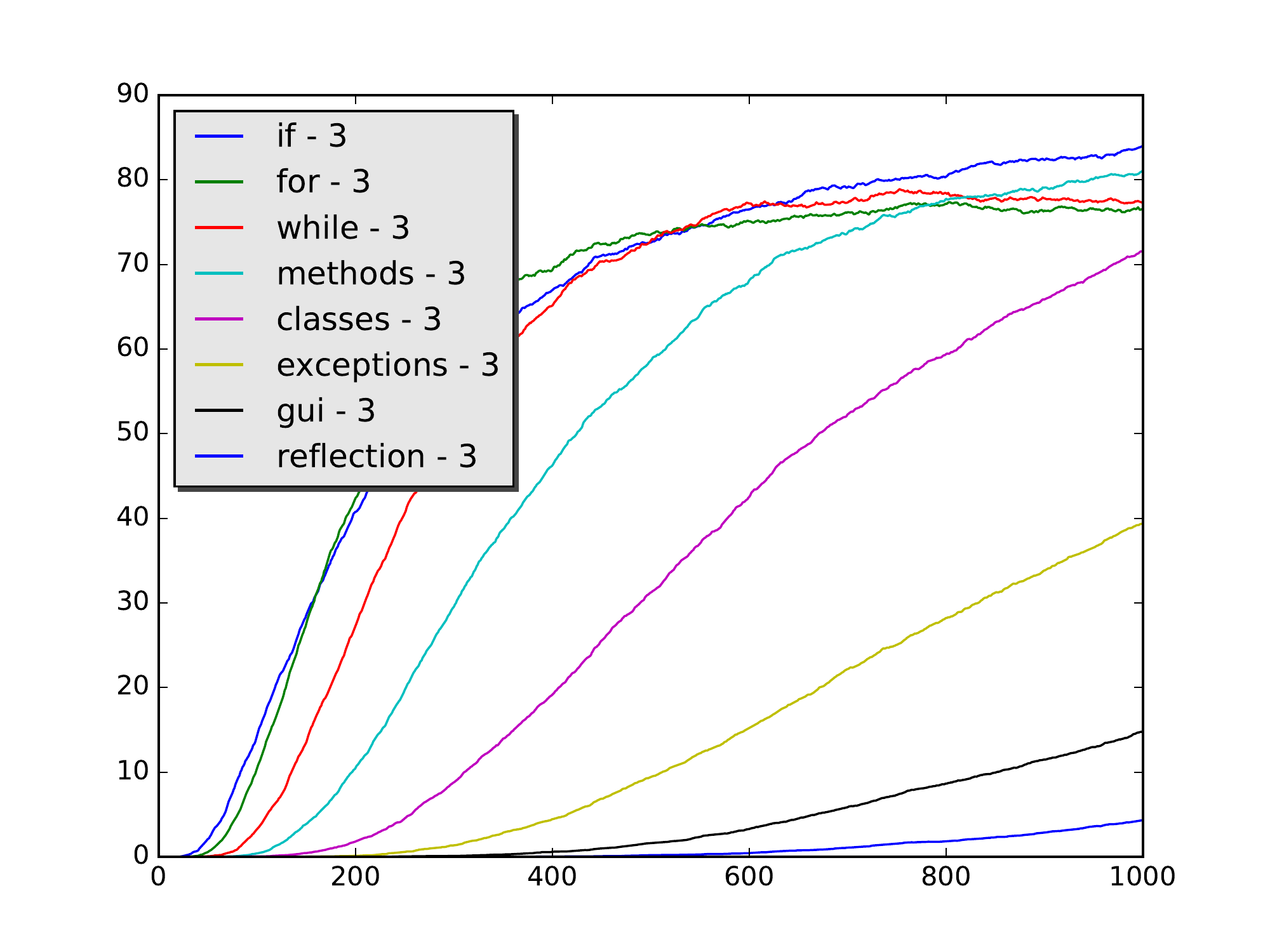} &
  \hspace{-1.6em}\includegraphics[width=35mm]{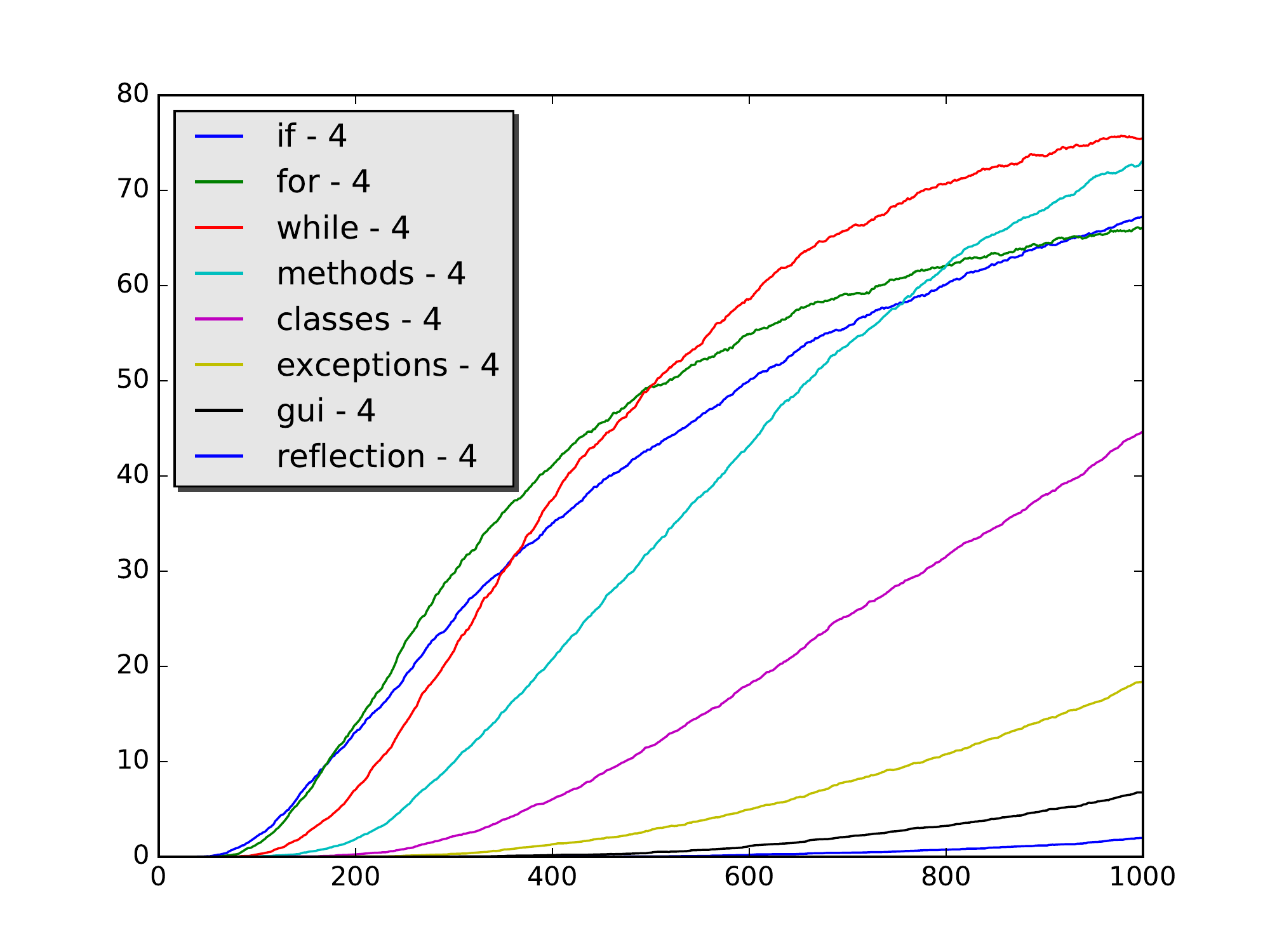} \\[-12pt]
  Level=1&
  \hspace{-1.6em} Level=2&
  \hspace{-1.6em} Level=3&
  \hspace{-1.6em} Level=4&
  \hspace{-1.6em} Level=5\\
  \includegraphics[width=35mm]{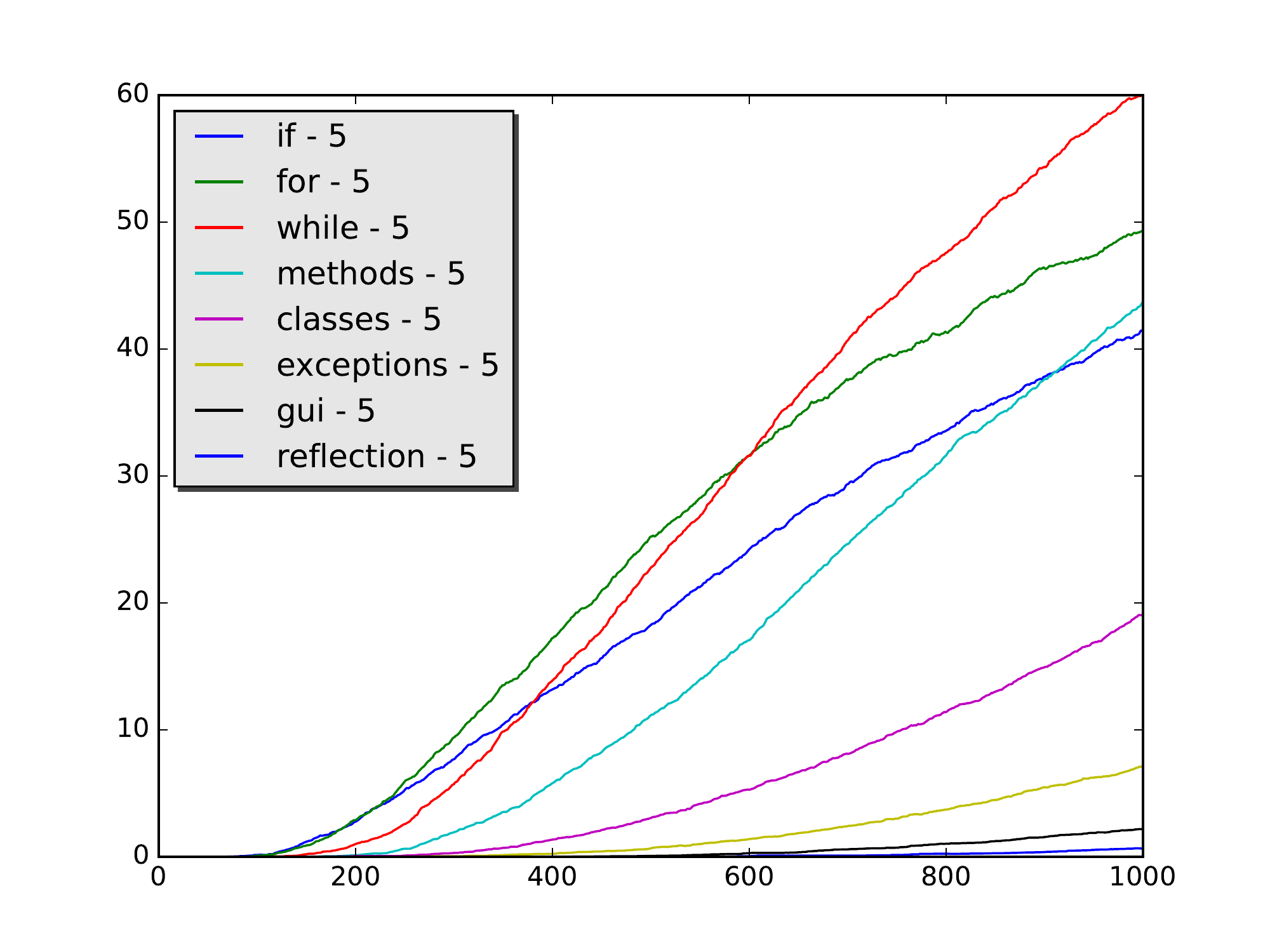} &
  \hspace{-1.6em}\includegraphics[width=35mm]{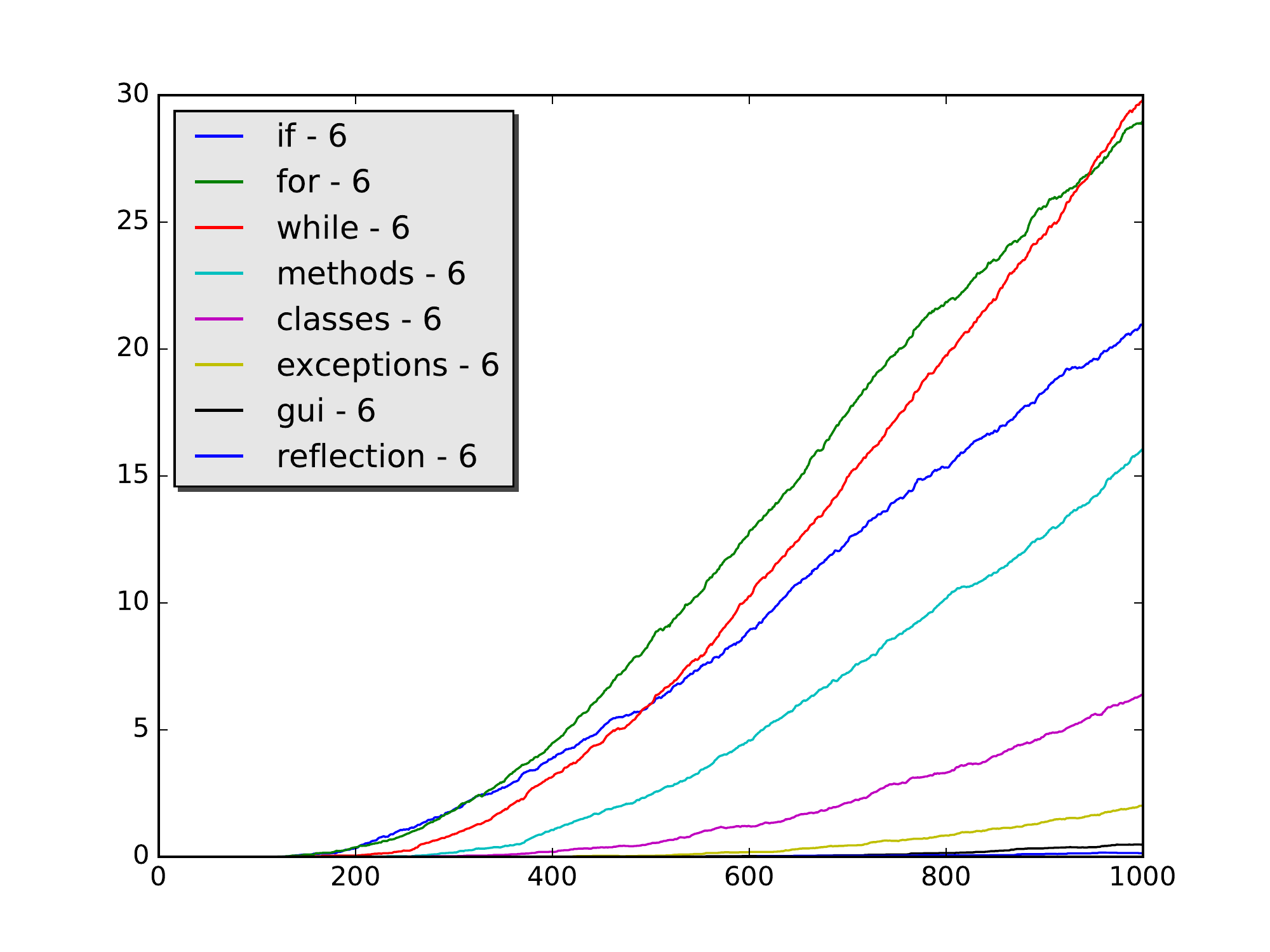} &
  \hspace{-1.6em}\includegraphics[width=35mm]{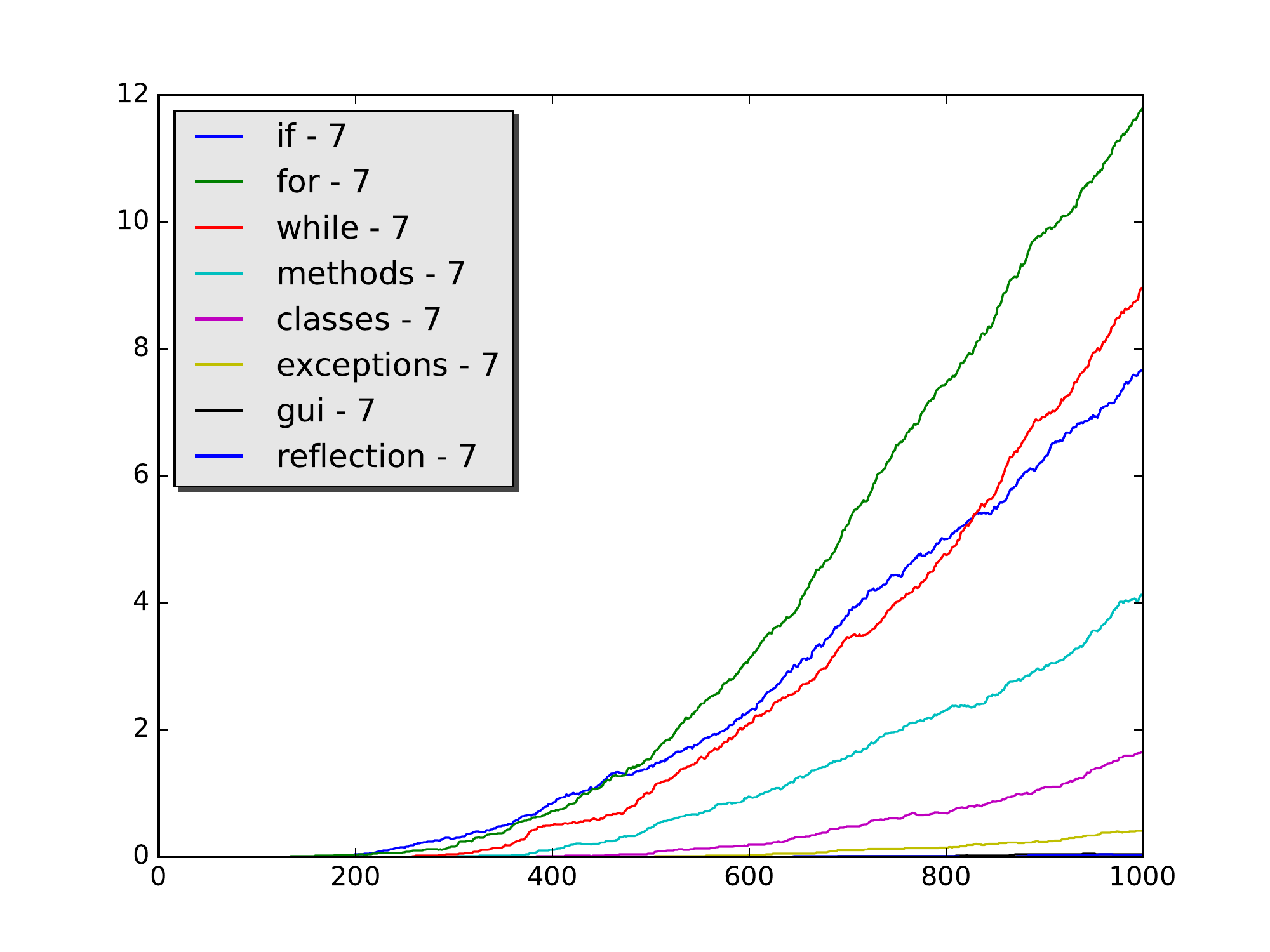} &
  \hspace{-1.6em}\includegraphics[width=35mm]{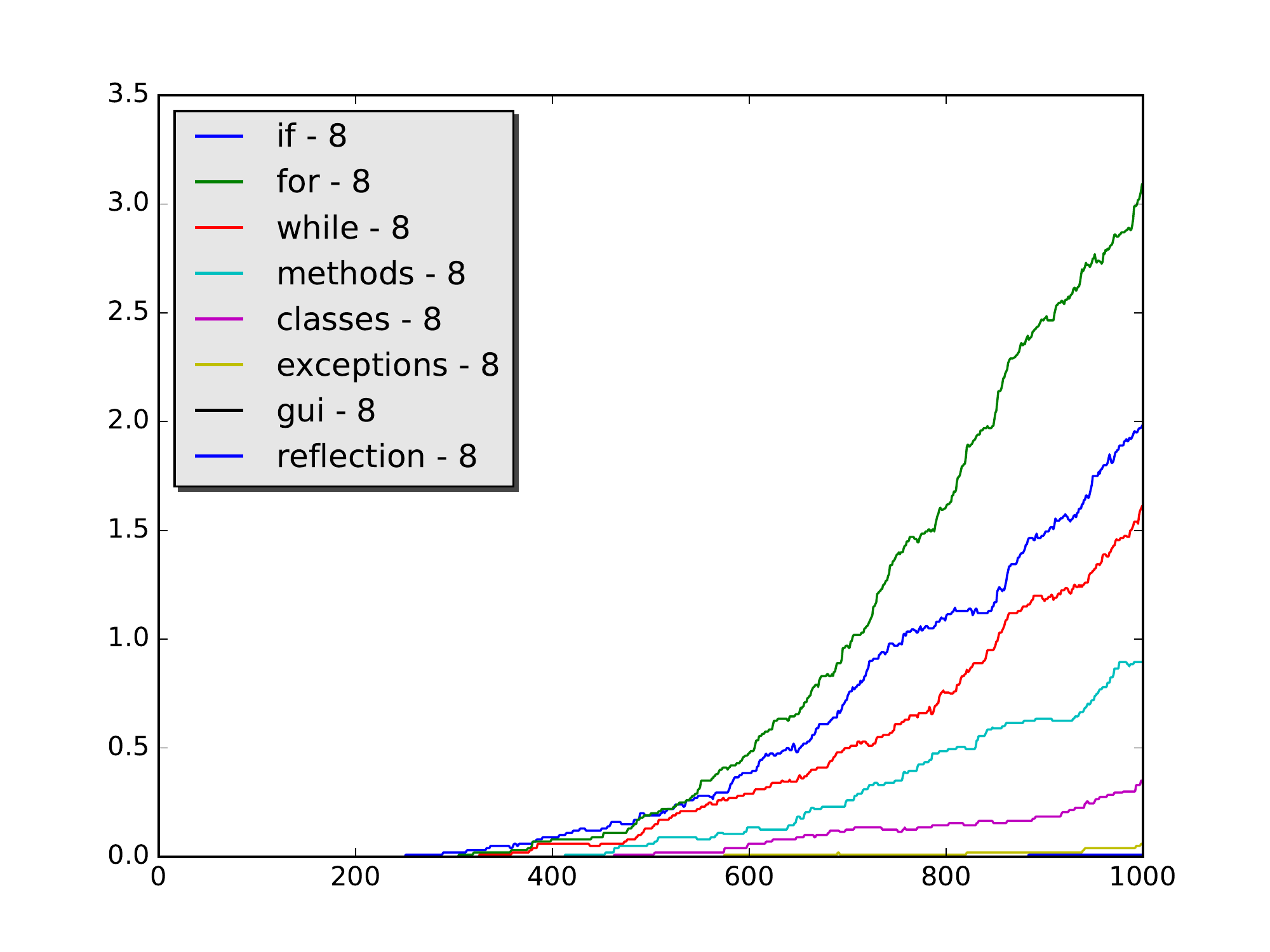} &
  \hspace{-1.6em}\includegraphics[width=35mm]{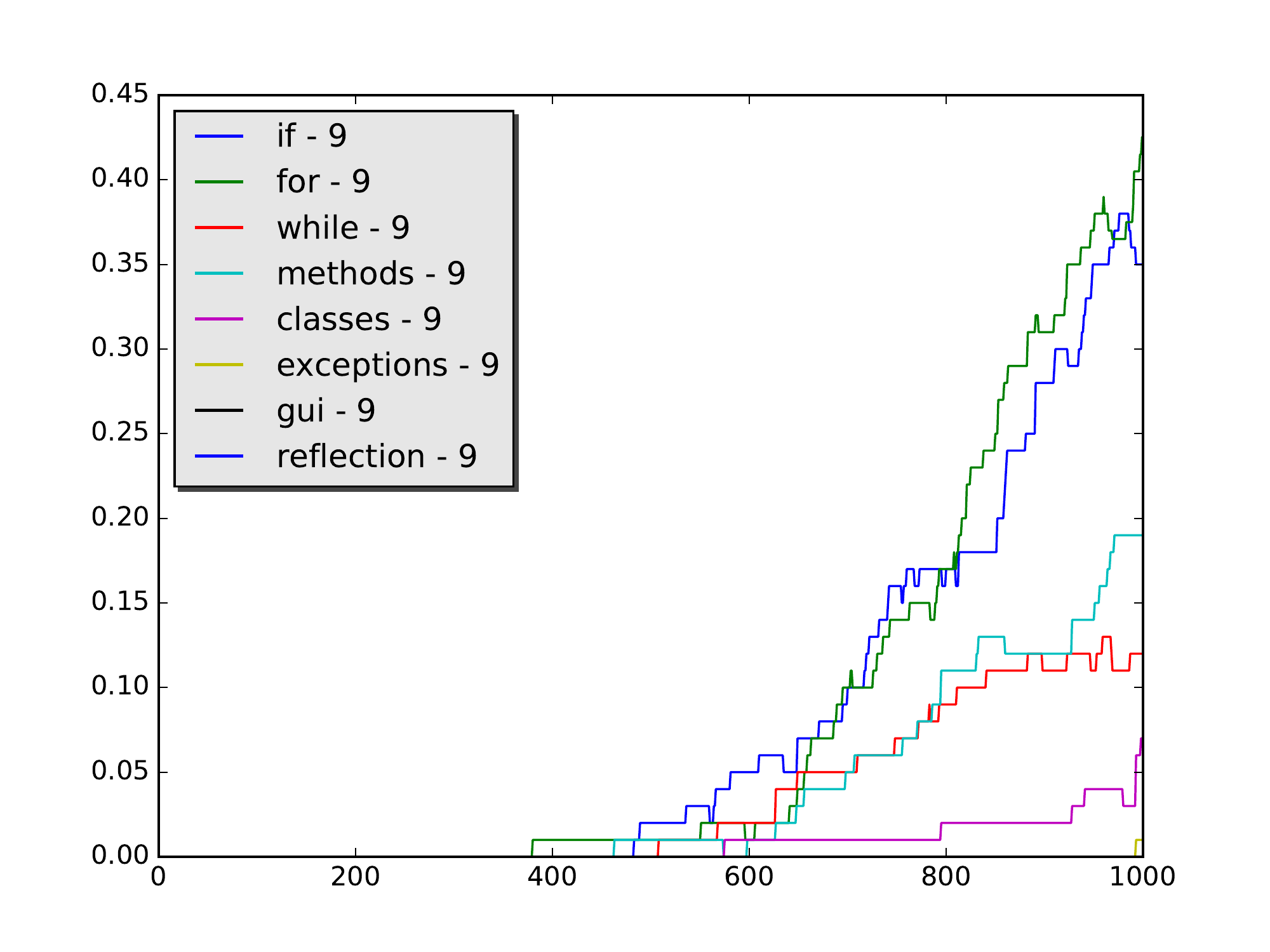} \\[-12pt]
  Level=6&
  \hspace{-1.6em} Level=7&
  \hspace{-1.6em} Level=8&
  \hspace{-1.6em} Level=9&
  \hspace{-1.6em} Level=10\\

\end{tabular}

\caption{Static Environment - Good Student}
\label{fig:Good-static-student}
\end{figure*}
}
    
%


    \subsection{Case 2 - Static epsilon}
     \textbf{Parameters - 1000 students, 200 task-sets, 100 iterations}

    The next test is based on a static epsilon value, where we give the student a static chance of success on exploration. This test gives us a bit more information compared to test-1, because it will utilize the student matrix, allowing to account for knowledge gained from previous tasks. In this test the percentages are the same as the previous (20\% for the bad and 70\% for the good), because we wanted to research if the student progressed more when he gained the ability to use previous knowledge.

    Figure \ref{fig:Bad-static-student-eps} and Figure \ref{fig:Good-static-student-eps} clearly show the student improves slightly the learning rate. We now visualize a student that is able to use the previous knowledge and also has a fairly good chance to ``guess" when he needs to explore (in MAB terms) for an optimal solution. The same goes for the ``bad-student", which drastically increased the final outcome.

{\renewcommand{\arraystretch}{2.0}
\setlength{\textfloatsep}{10pt plus 1.0pt minus 2.0pt}
\begin{figure*}[ht]
\begin{tabular}{*{5}{c}}
  \includegraphics[width=35mm]{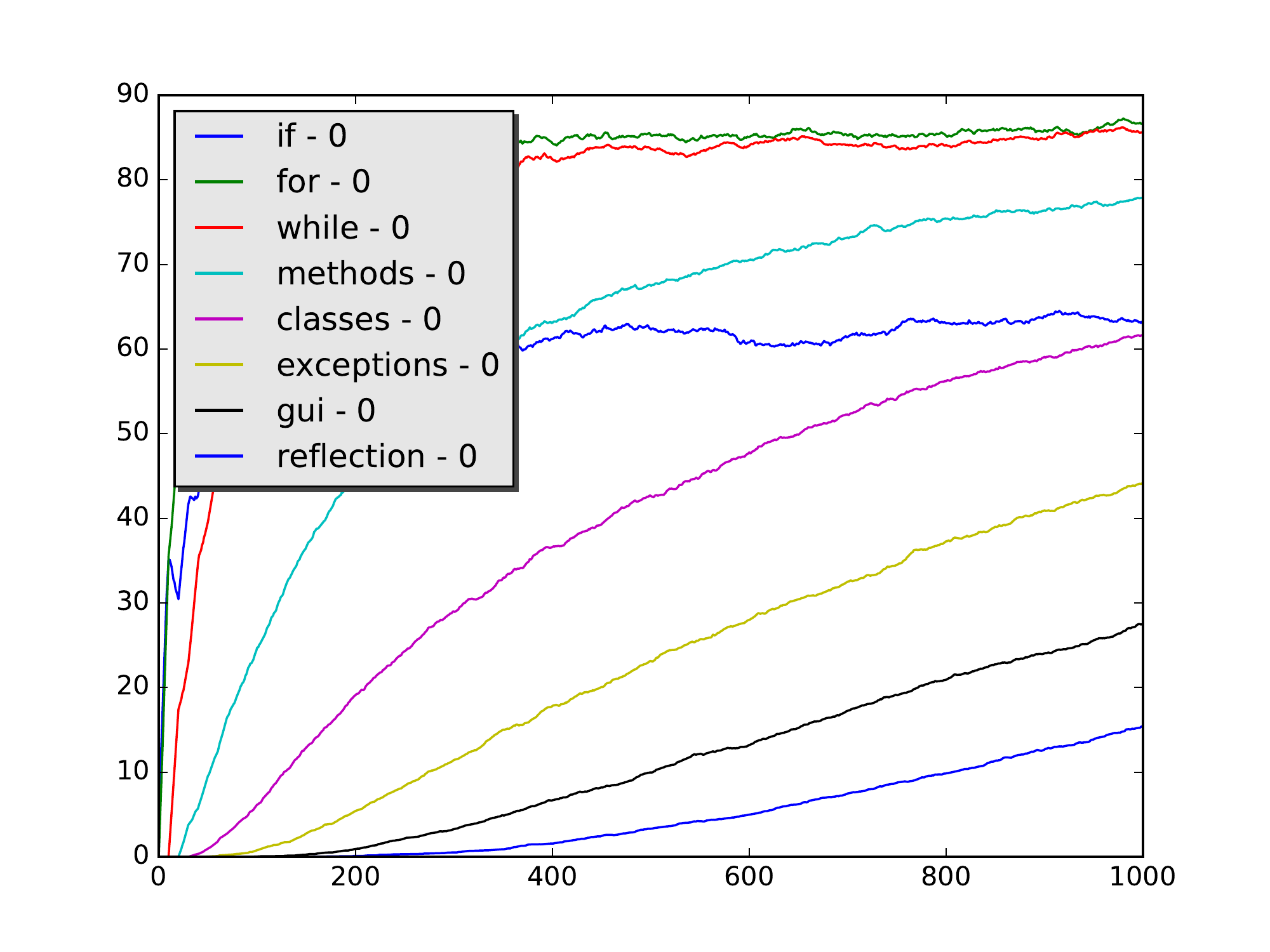} &
  \hspace{-1.6em}\includegraphics[width=35mm]{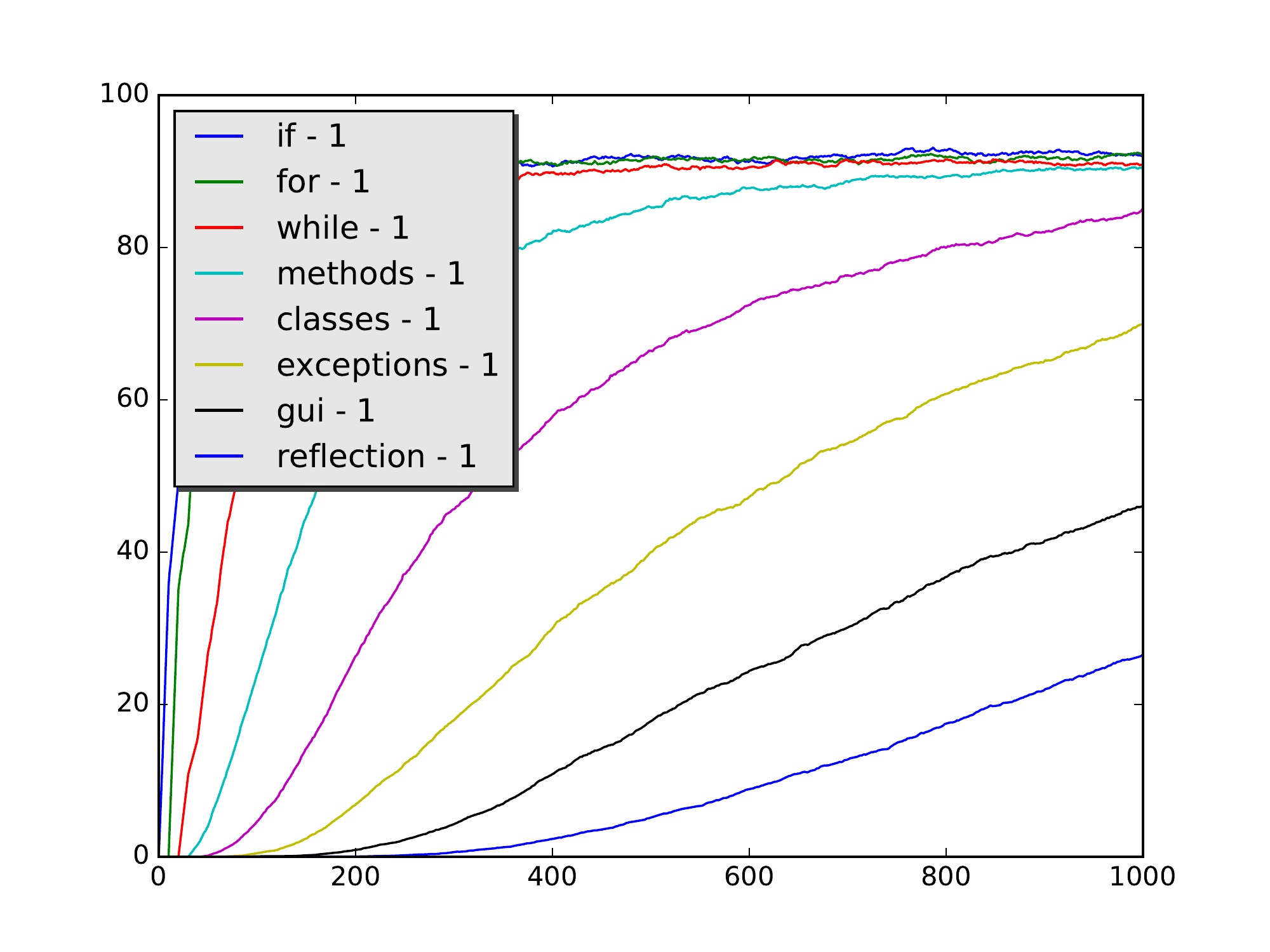} &
  \hspace{-1.6em}\includegraphics[width=35mm]{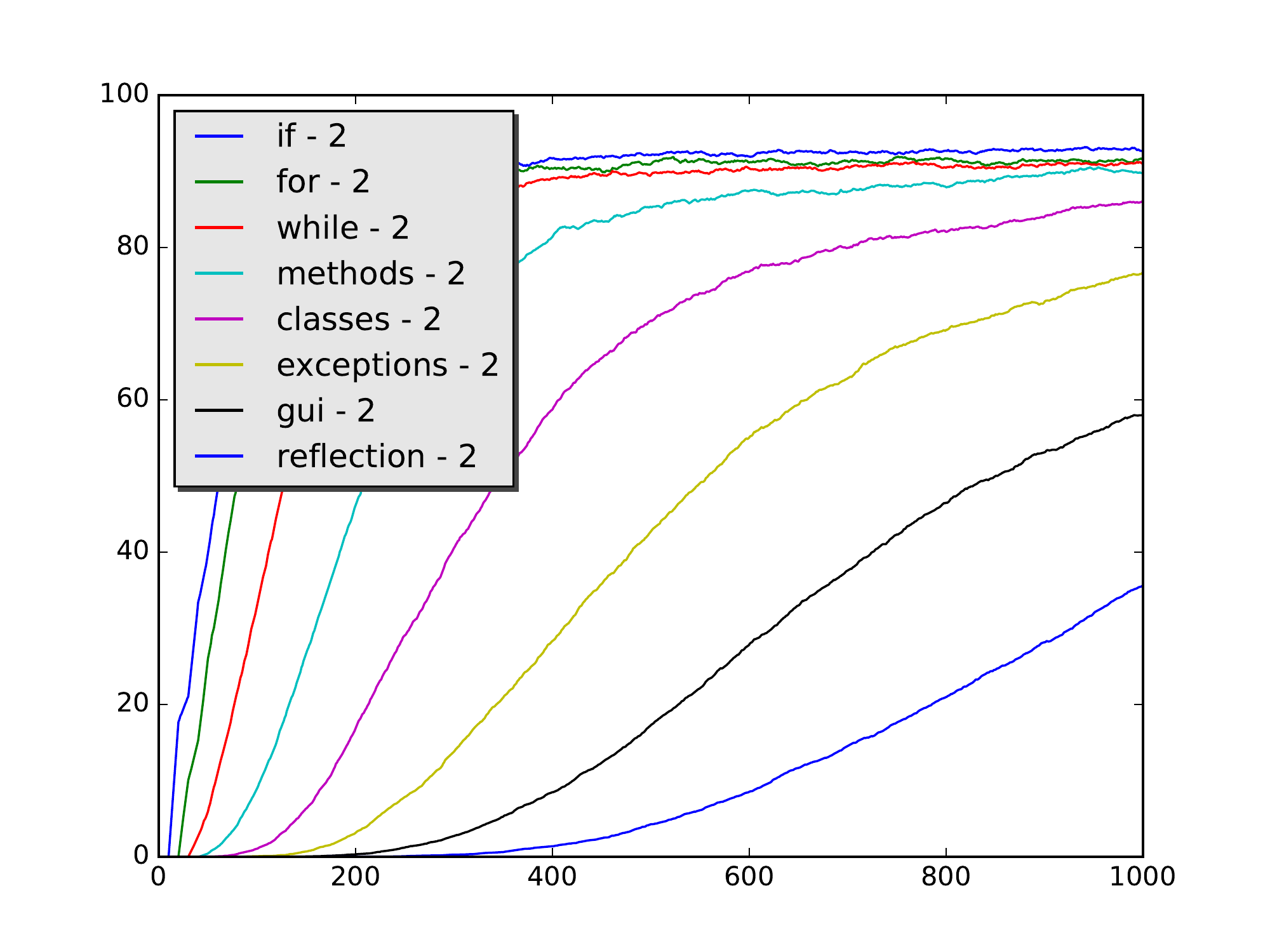} &
  \hspace{-1.6em}\includegraphics[width=35mm]{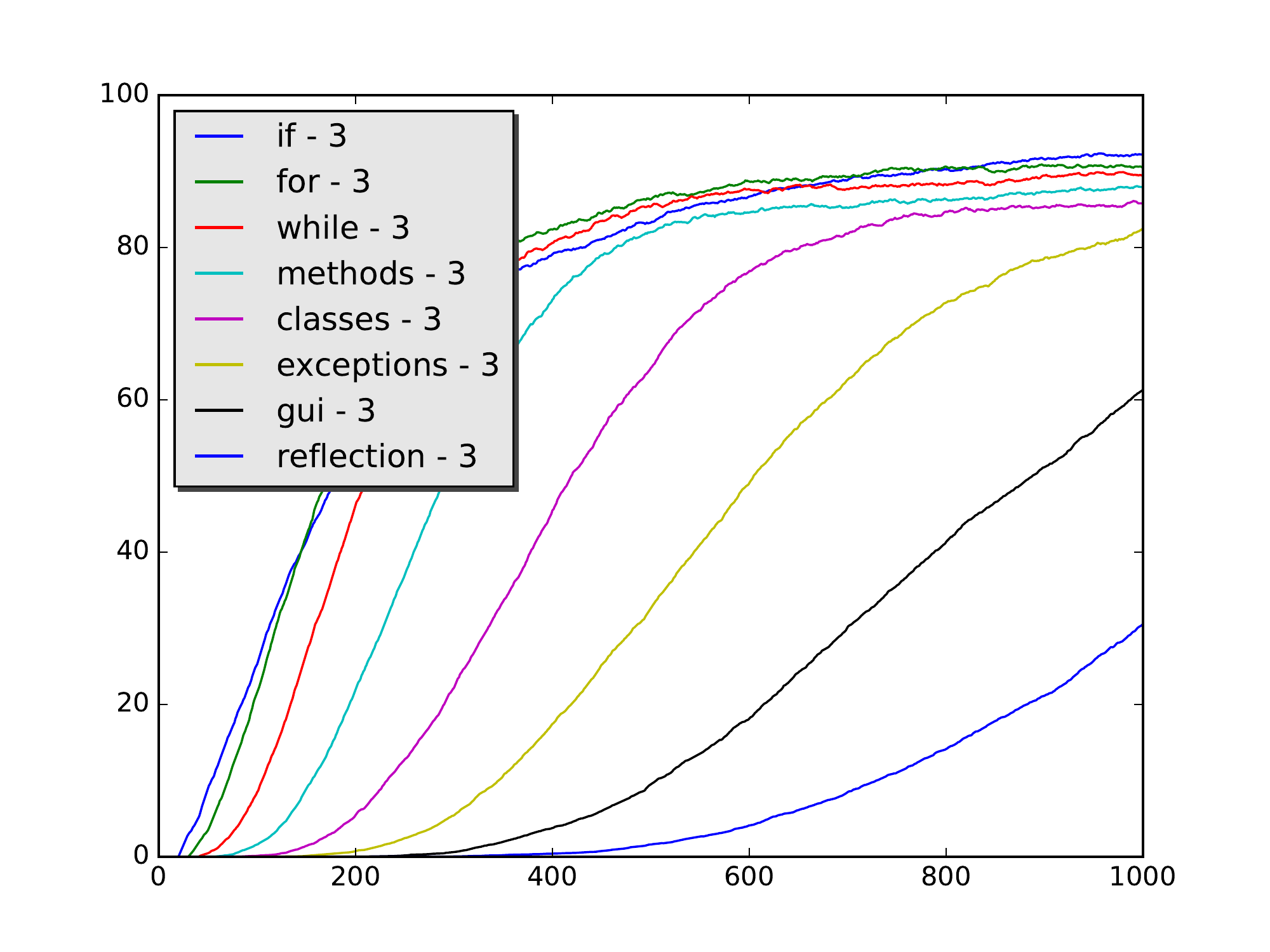} &
  \hspace{-1.6em}\includegraphics[width=35mm]{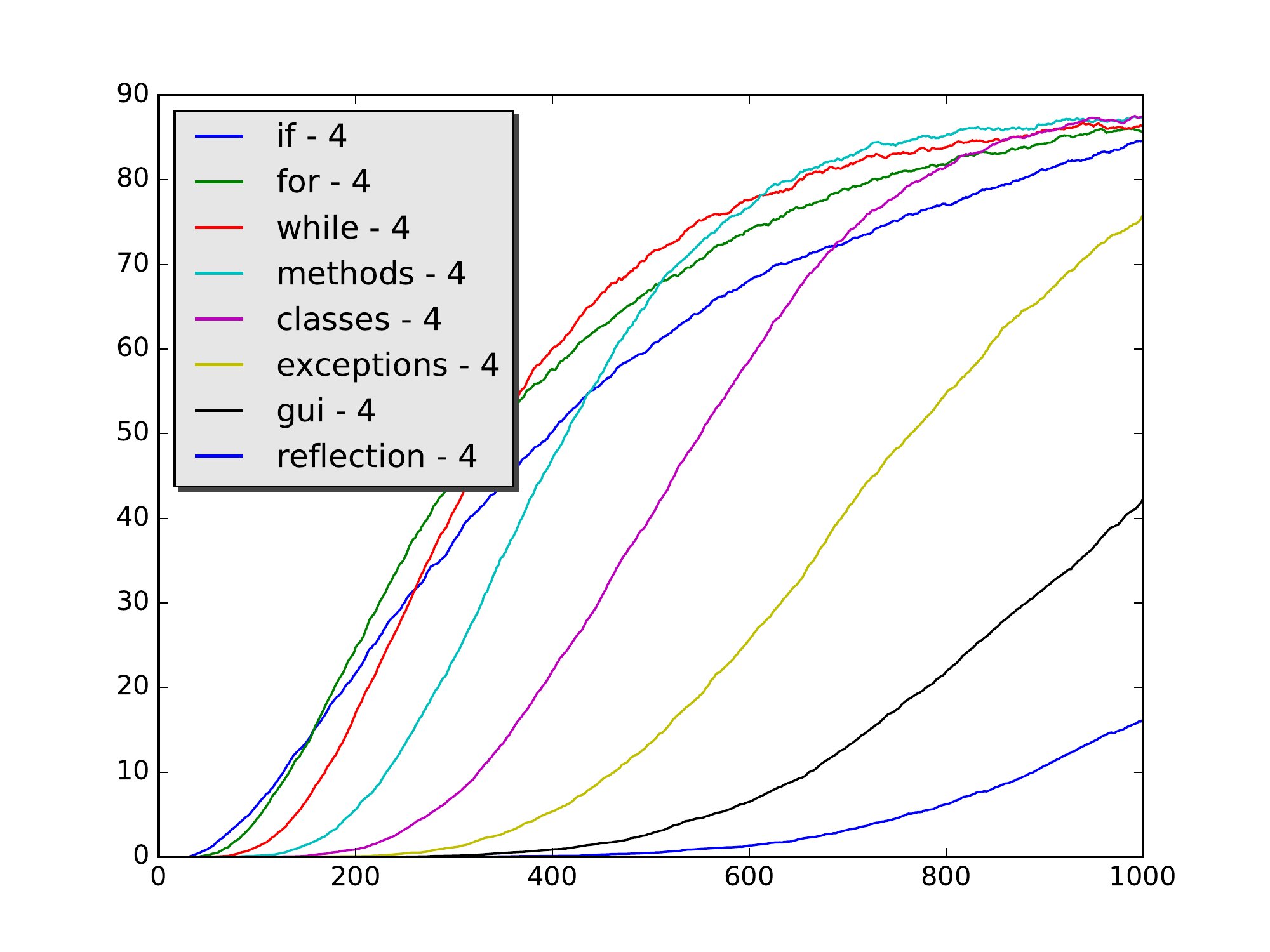} \\[-12pt]
  Level=1&
  \hspace{-1.6em} Level=2&
  \hspace{-1.6em} Level=3&
  \hspace{-1.6em} Level=4&
  \hspace{-1.6em} Level=5\\
  \includegraphics[width=35mm]{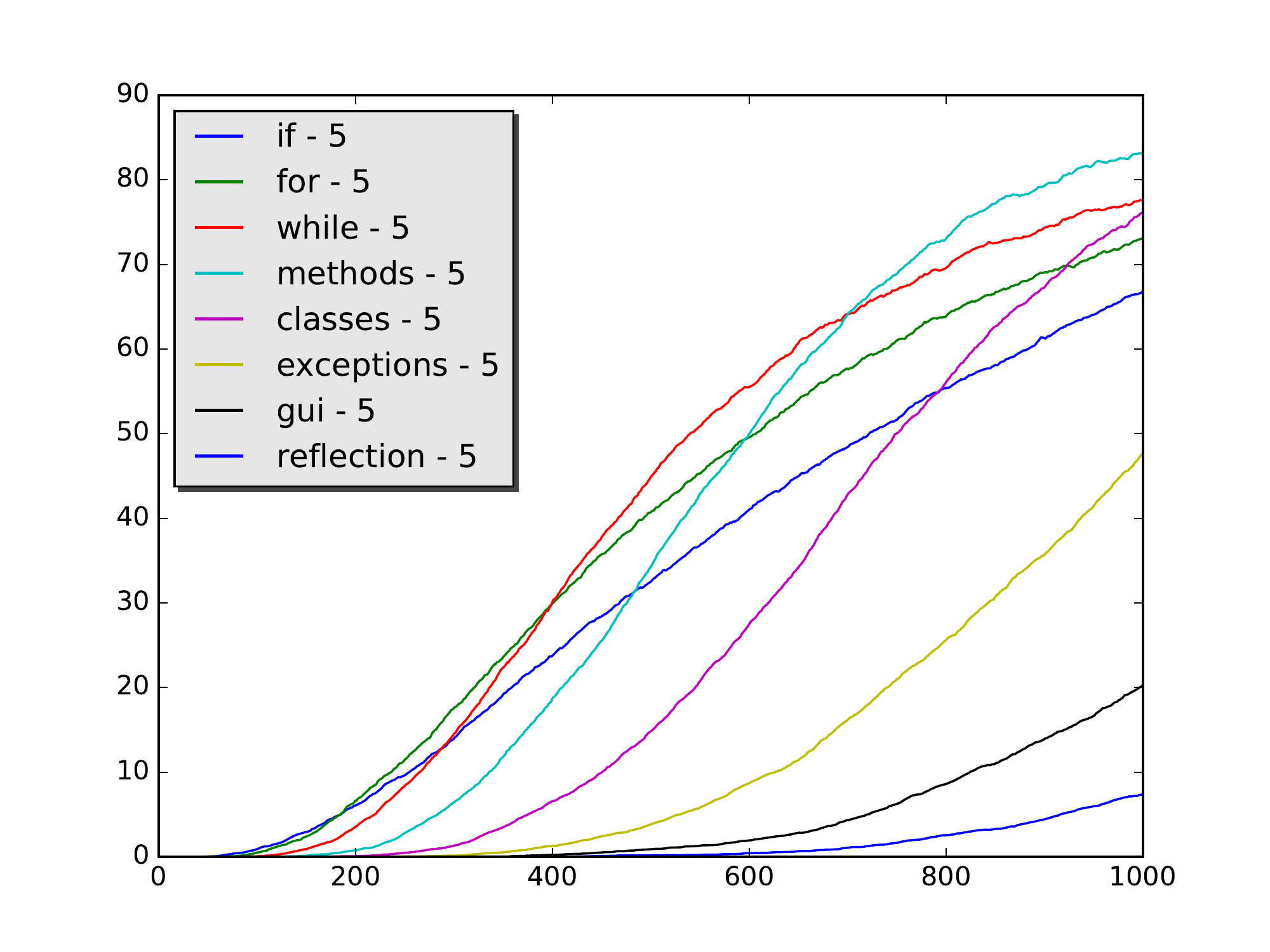} &
  \hspace{-1.6em}\includegraphics[width=35mm]{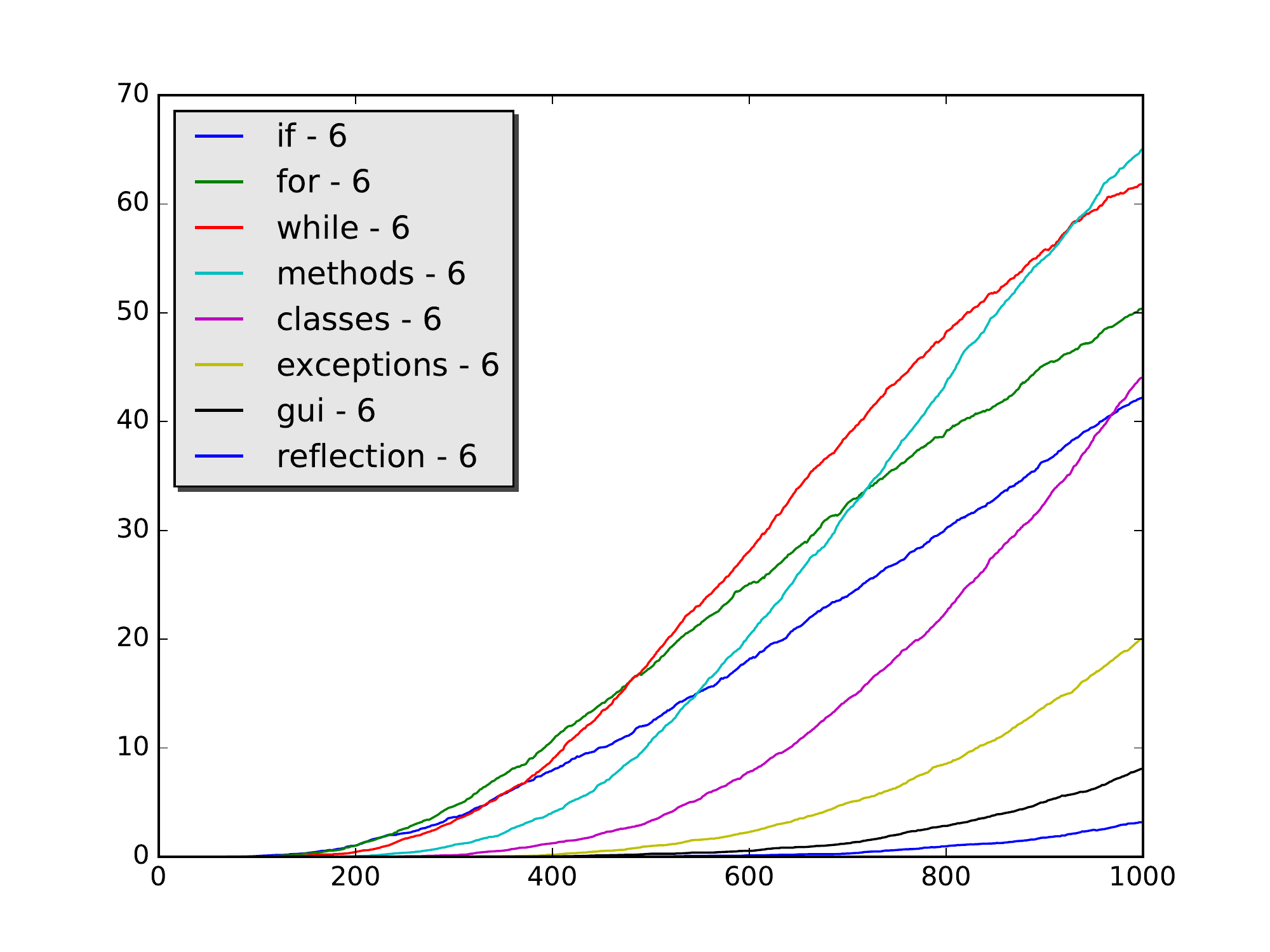} &
  \hspace{-1.6em}\includegraphics[width=35mm]{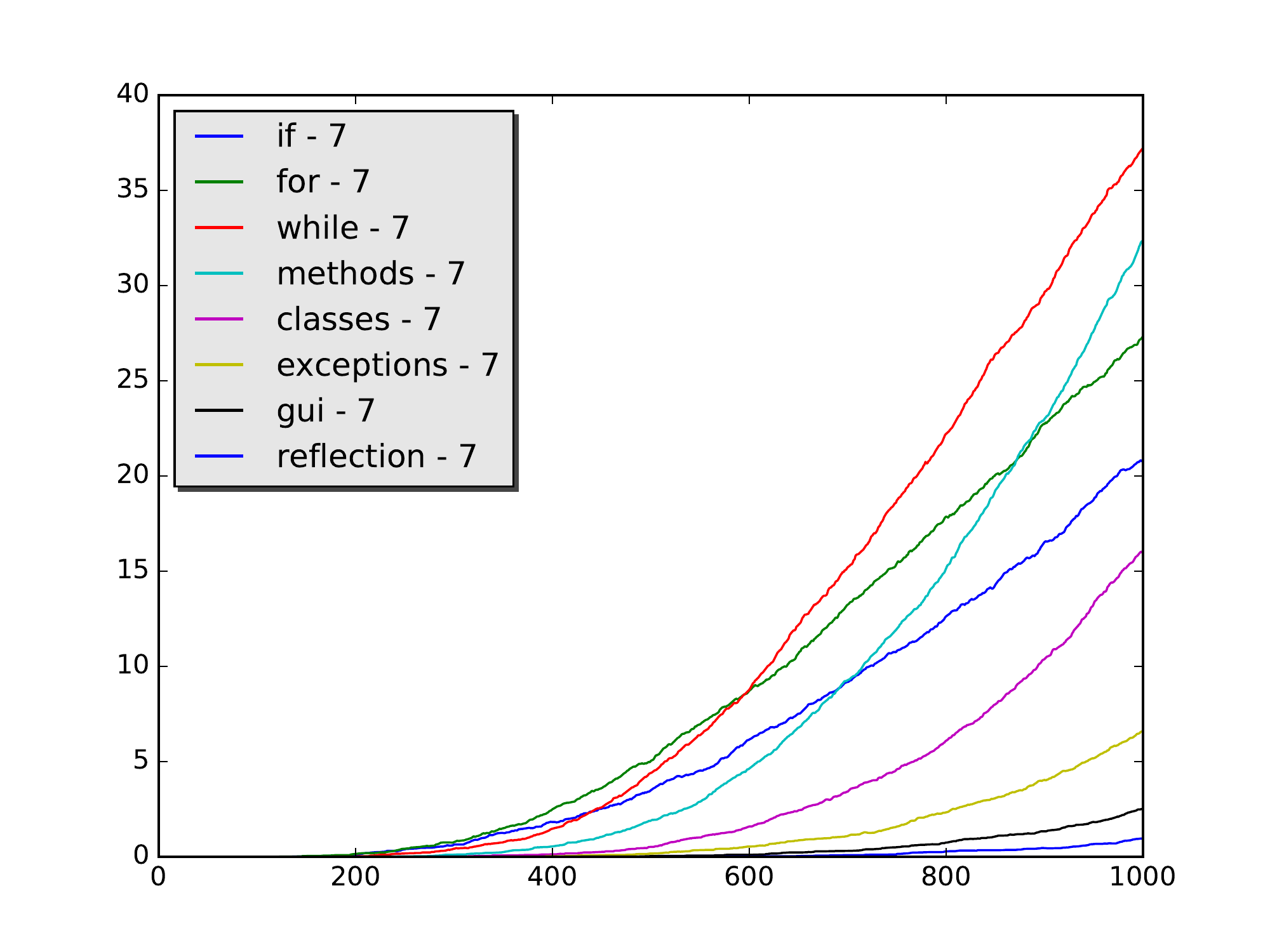} &
  \hspace{-1.6em}\includegraphics[width=35mm]{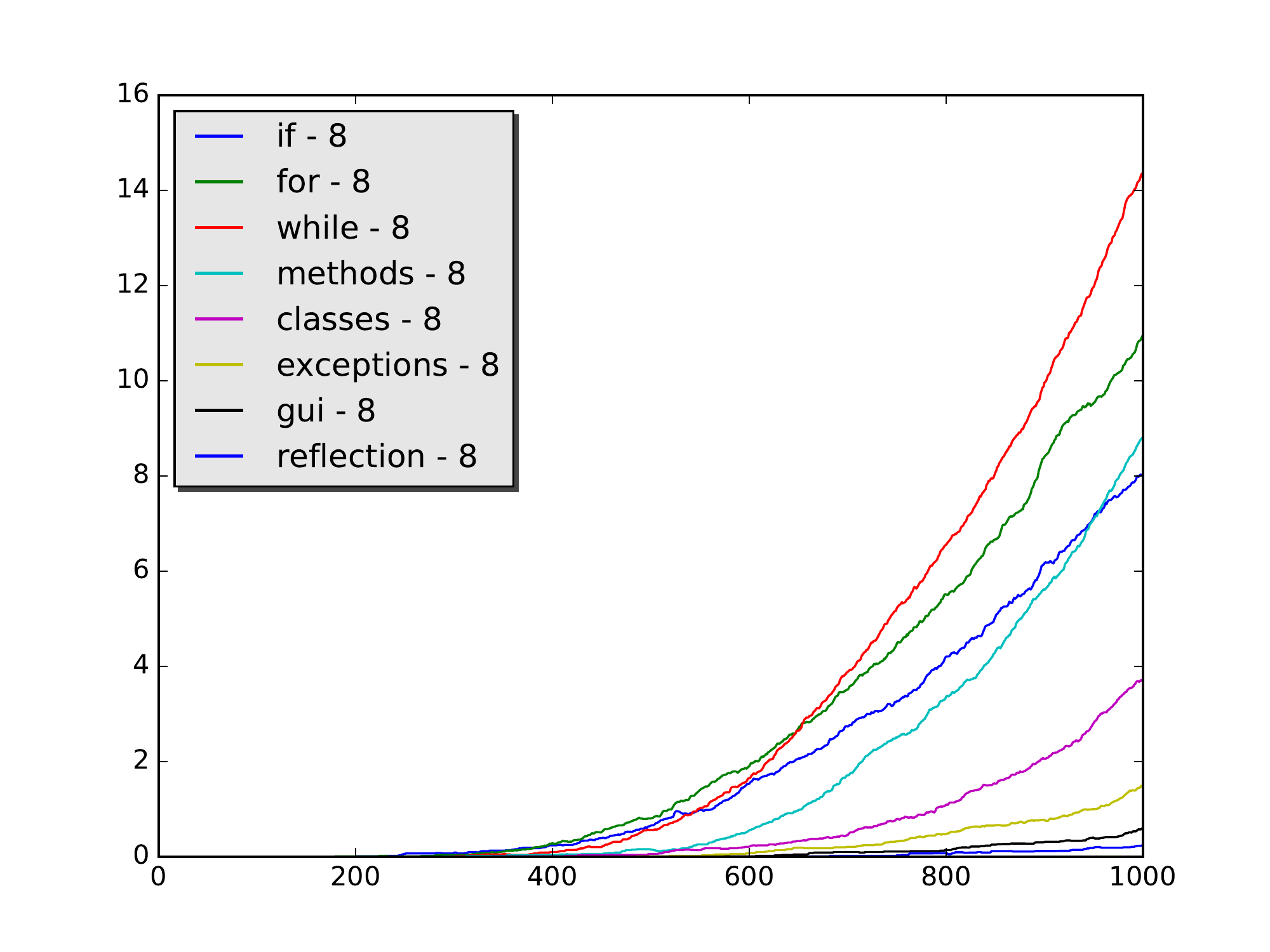} &
  \hspace{-1.6em}\includegraphics[width=35mm]{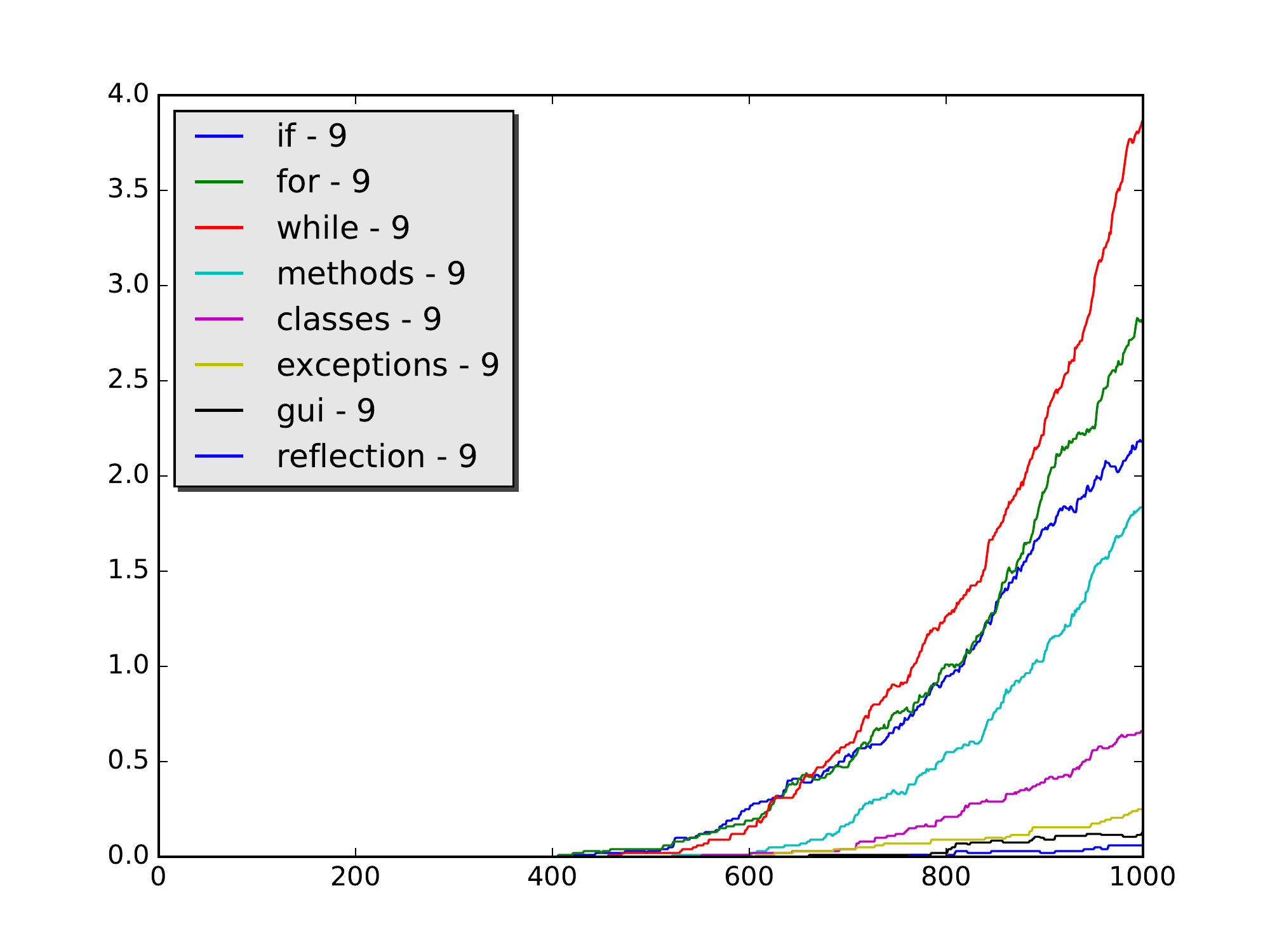} \\[-12pt]
  Level=6&
  \hspace{-1.6em} Level=7&
  \hspace{-1.6em} Level=8&
  \hspace{-1.6em} Level=9&
  \hspace{-1.6em} Level=10\\

\end{tabular}

\caption{Static Epsilon - Bad Student}
\label{fig:Bad-static-student-eps}
\end{figure*}
}


{\renewcommand{\arraystretch}{2.0}
\setlength{\textfloatsep}{10pt plus 1.0pt minus 2.0pt}
\begin{figure*}[ht]
\begin{tabular}{*{5}{c}}
  \includegraphics[width=35mm]{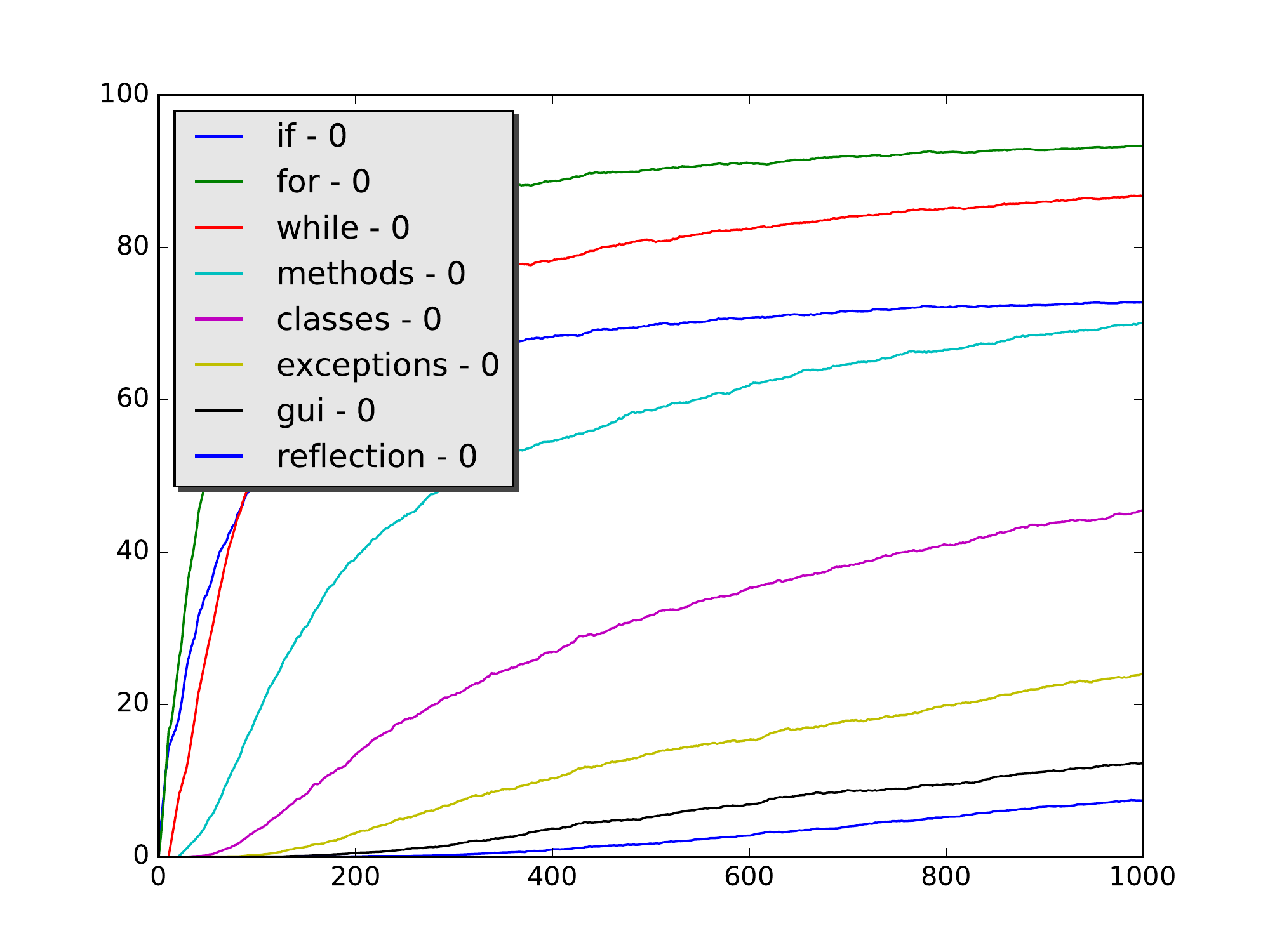} &
  \hspace{-1.6em}\includegraphics[width=35mm]{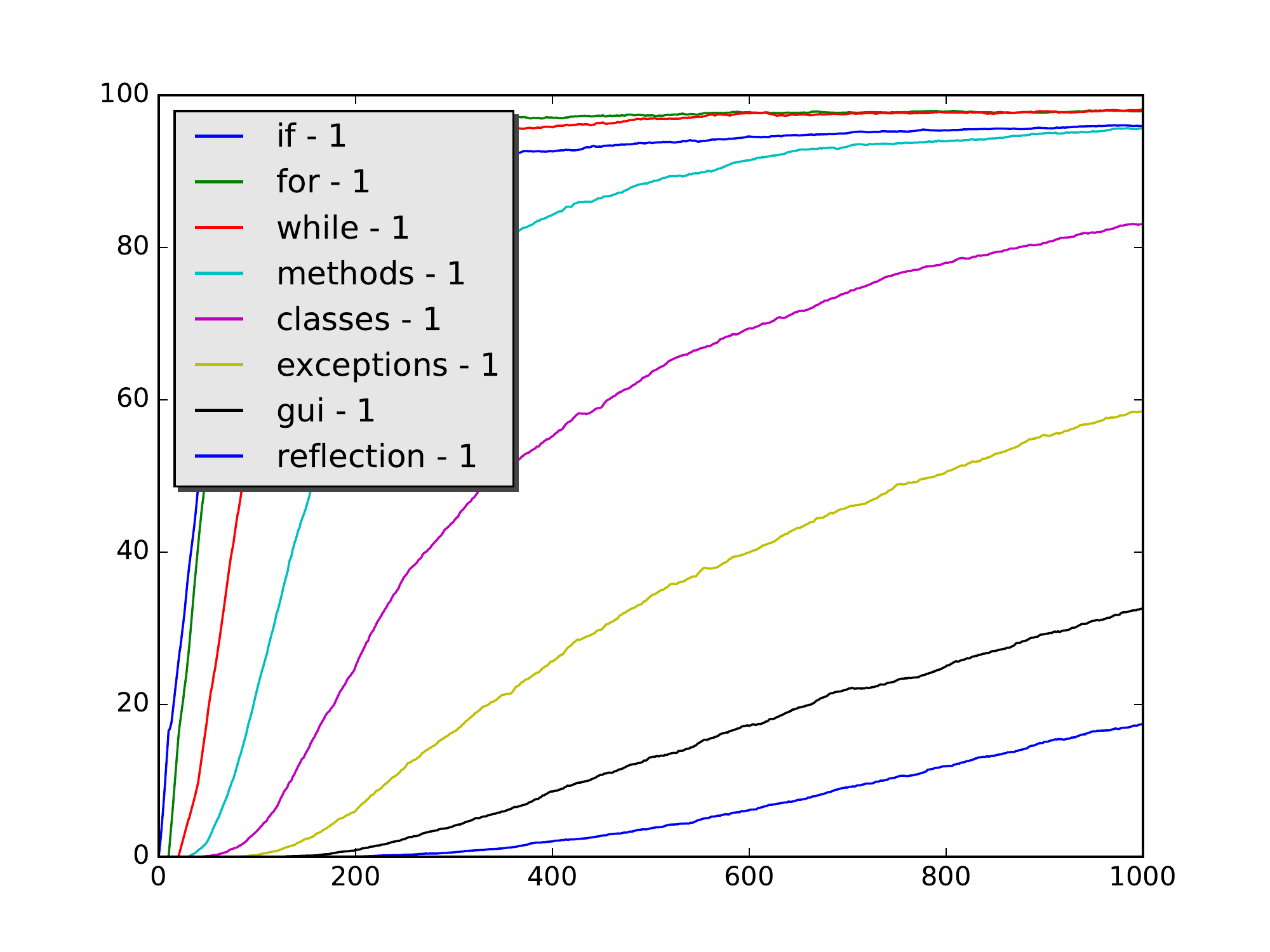} &
  \hspace{-1.6em}\includegraphics[width=35mm]{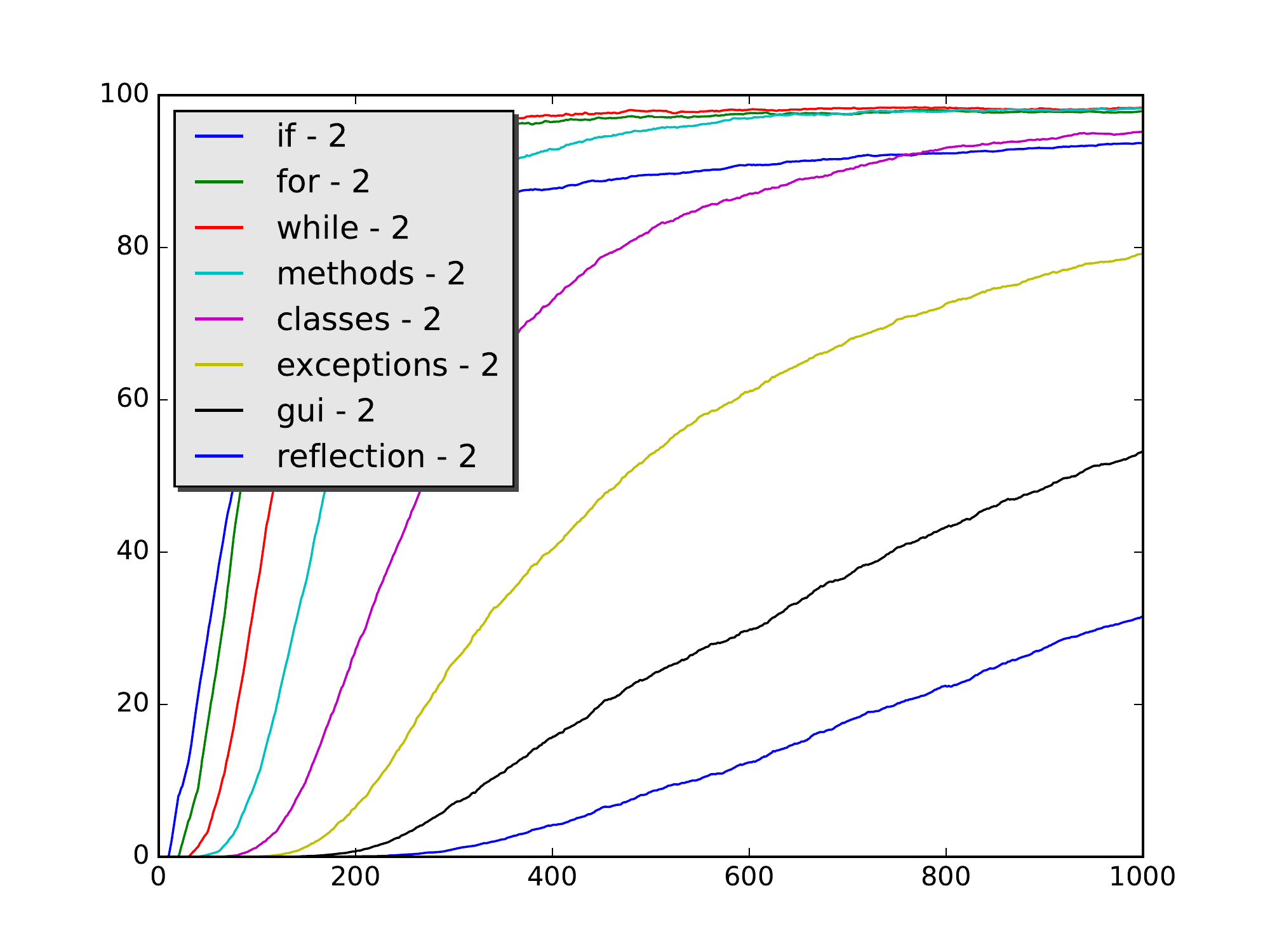} &
  \hspace{-1.6em}\includegraphics[width=35mm]{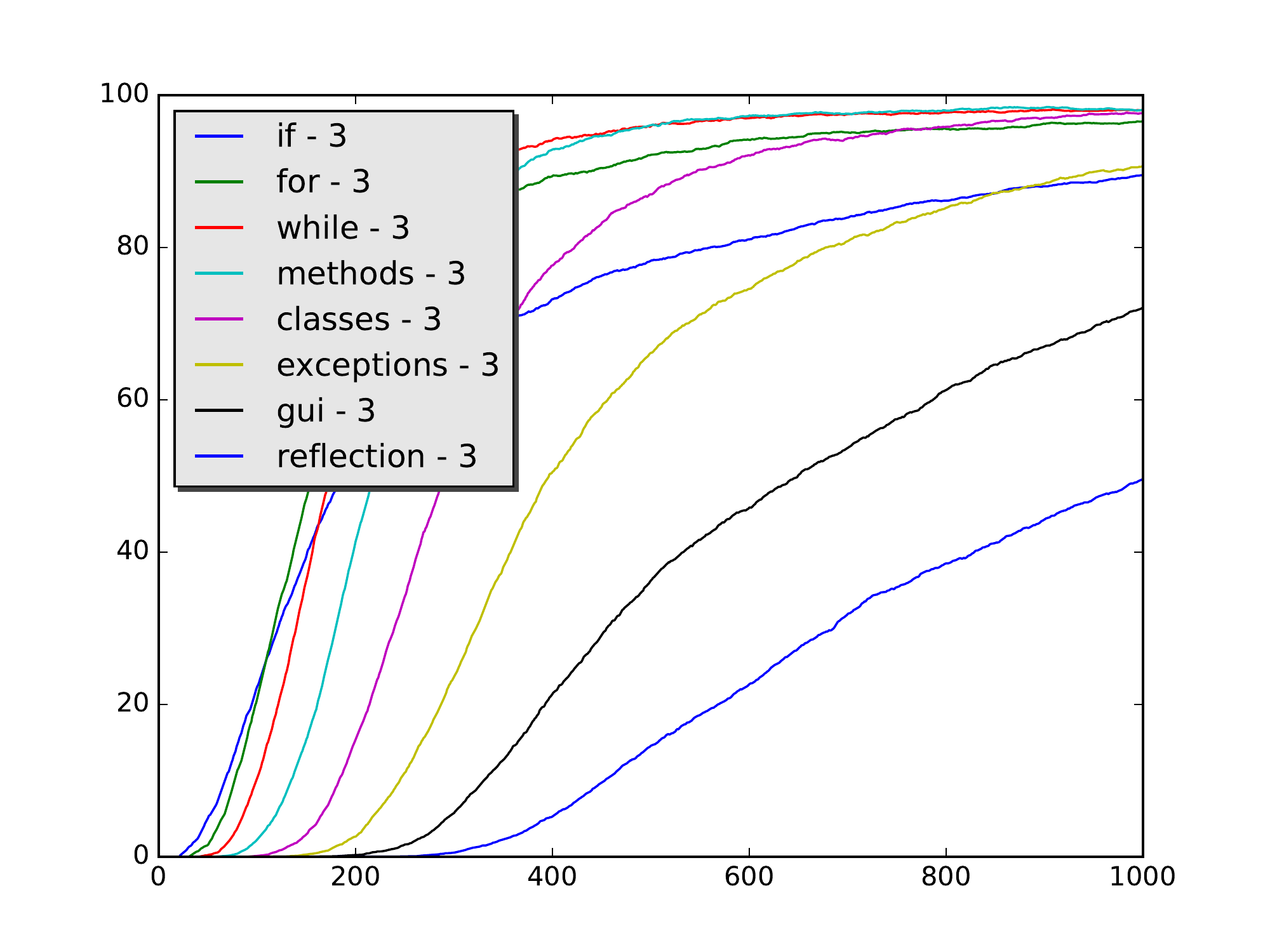} &
  \hspace{-1.6em}\includegraphics[width=35mm]{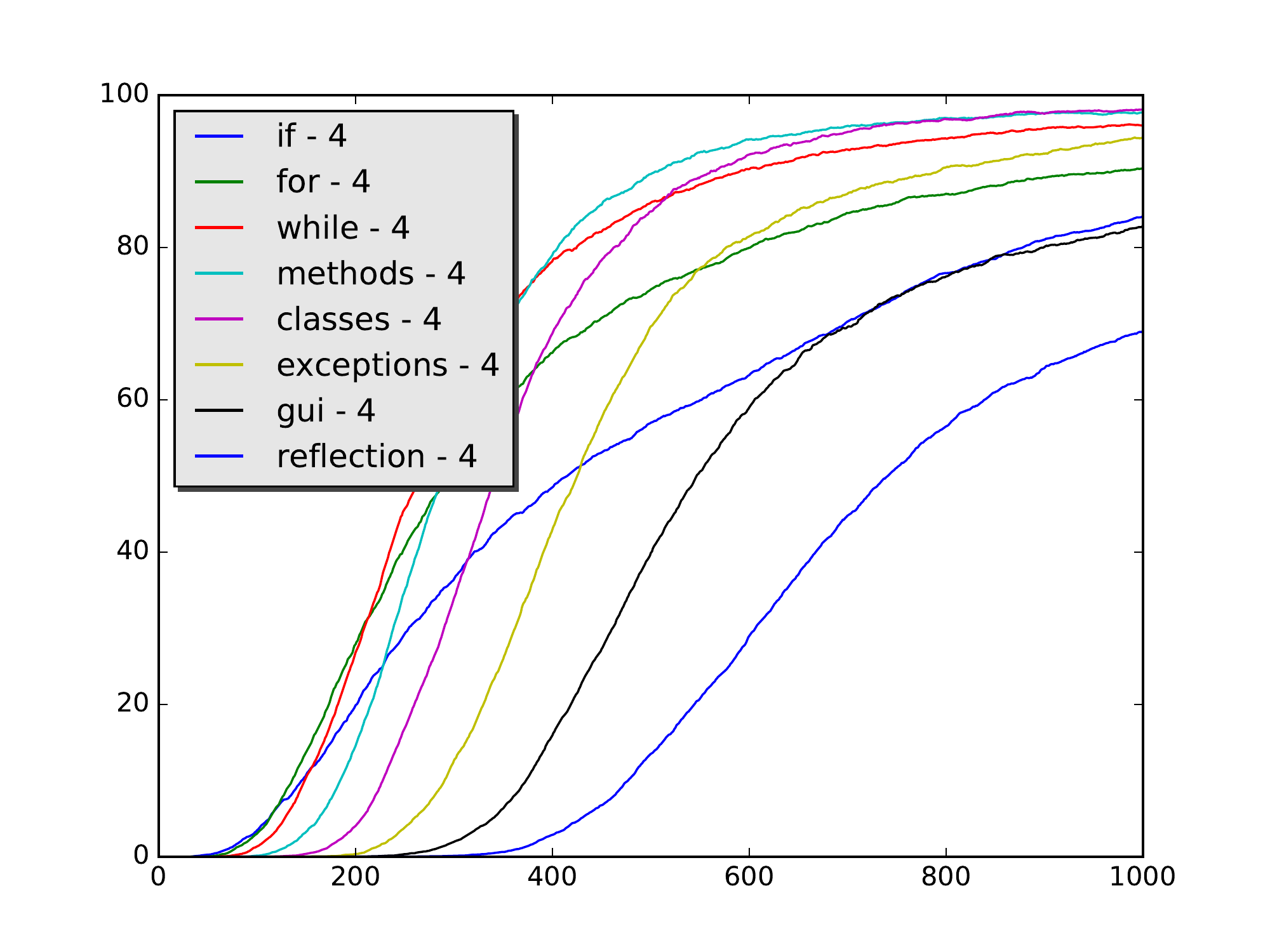} \\[-12pt]
  Level=1&
  \hspace{-1.6em} Level=2&
  \hspace{-1.6em} Level=3&
  \hspace{-1.6em} Level=4&
  \hspace{-1.6em} Level=5\\
  \includegraphics[width=35mm]{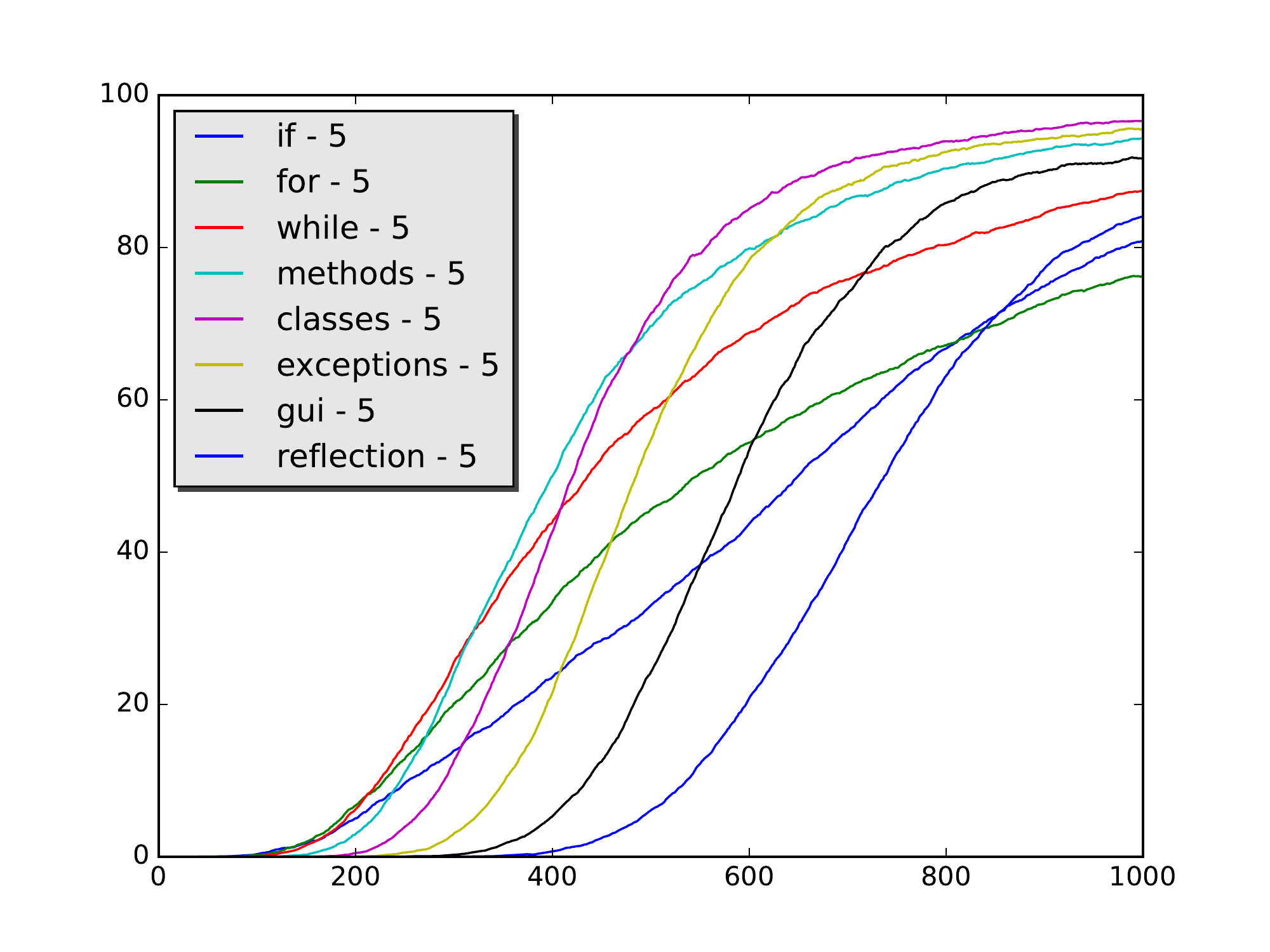} &
  \hspace{-1.6em}\includegraphics[width=35mm]{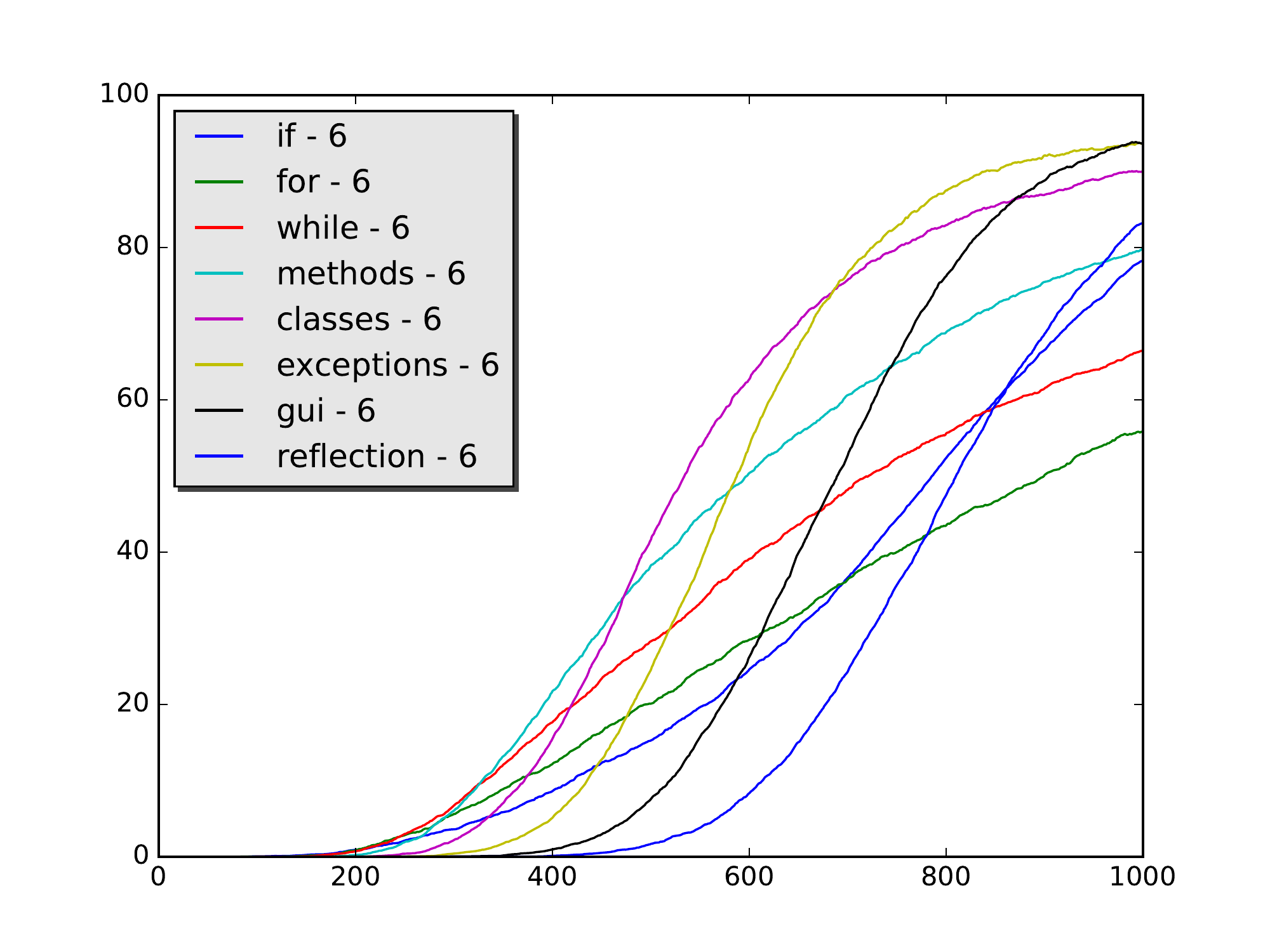} &
  \hspace{-1.6em}\includegraphics[width=35mm]{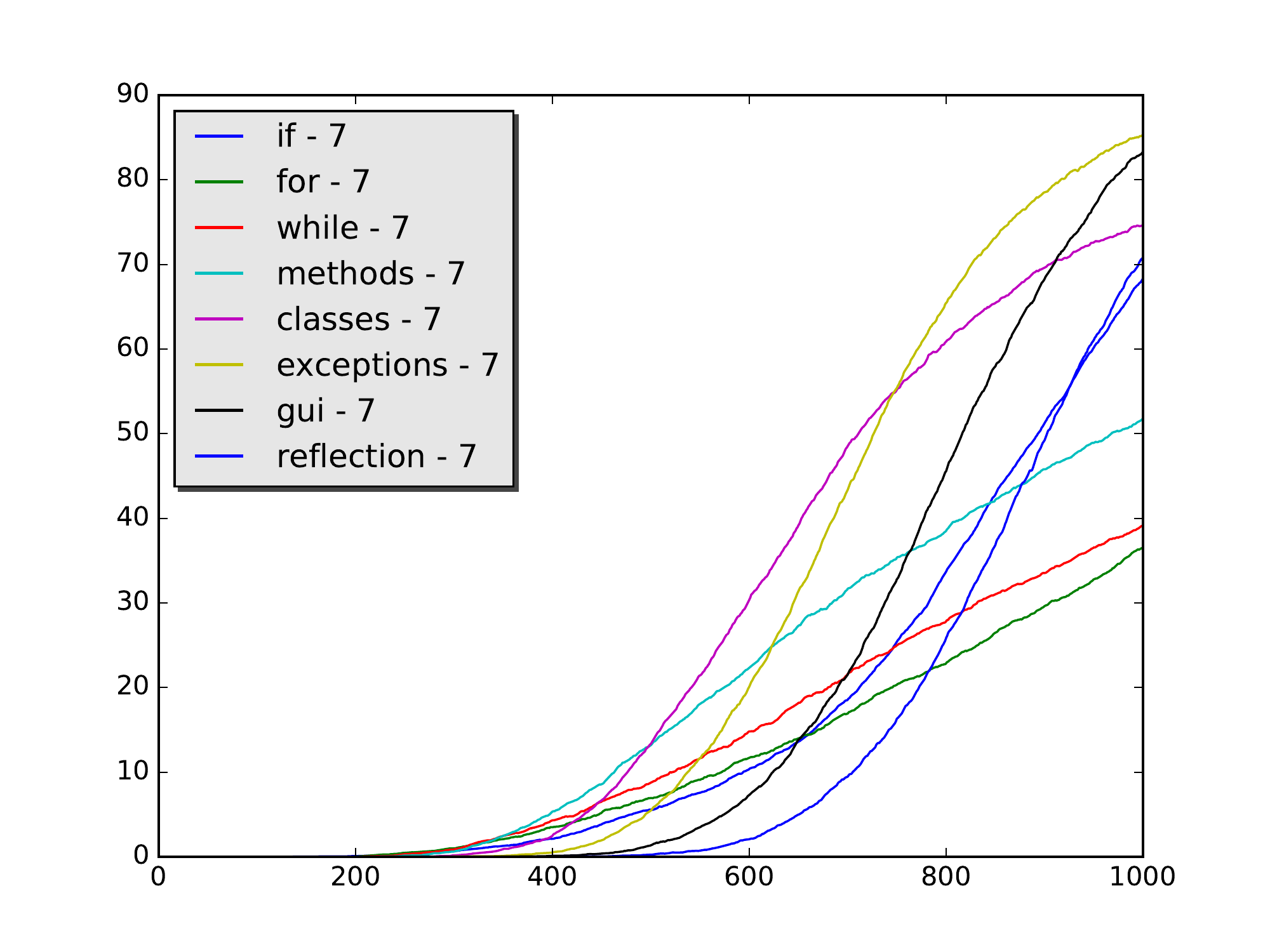} &
  \hspace{-1.6em}\includegraphics[width=35mm]{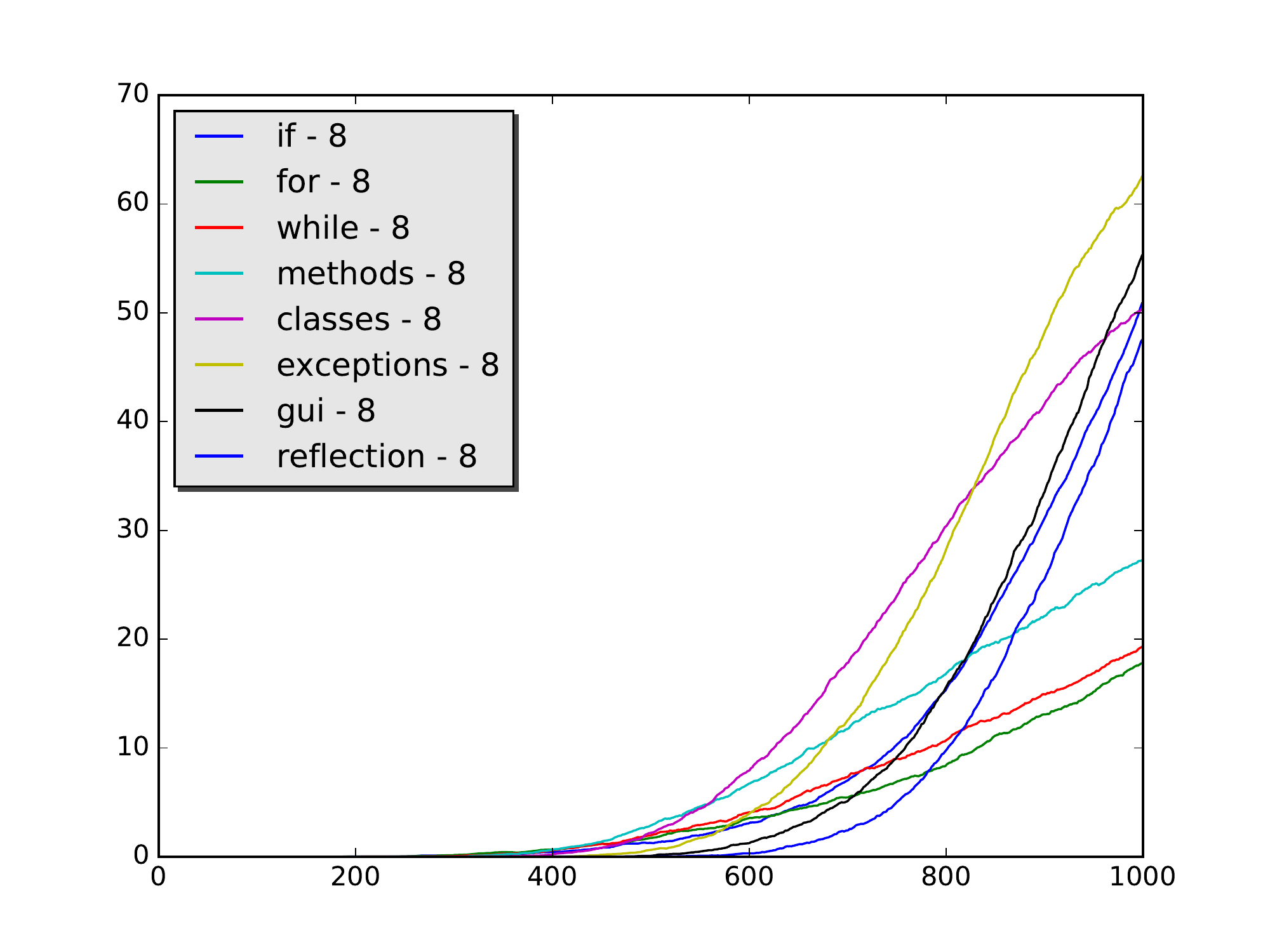} &
  \hspace{-1.6em}\includegraphics[width=35mm]{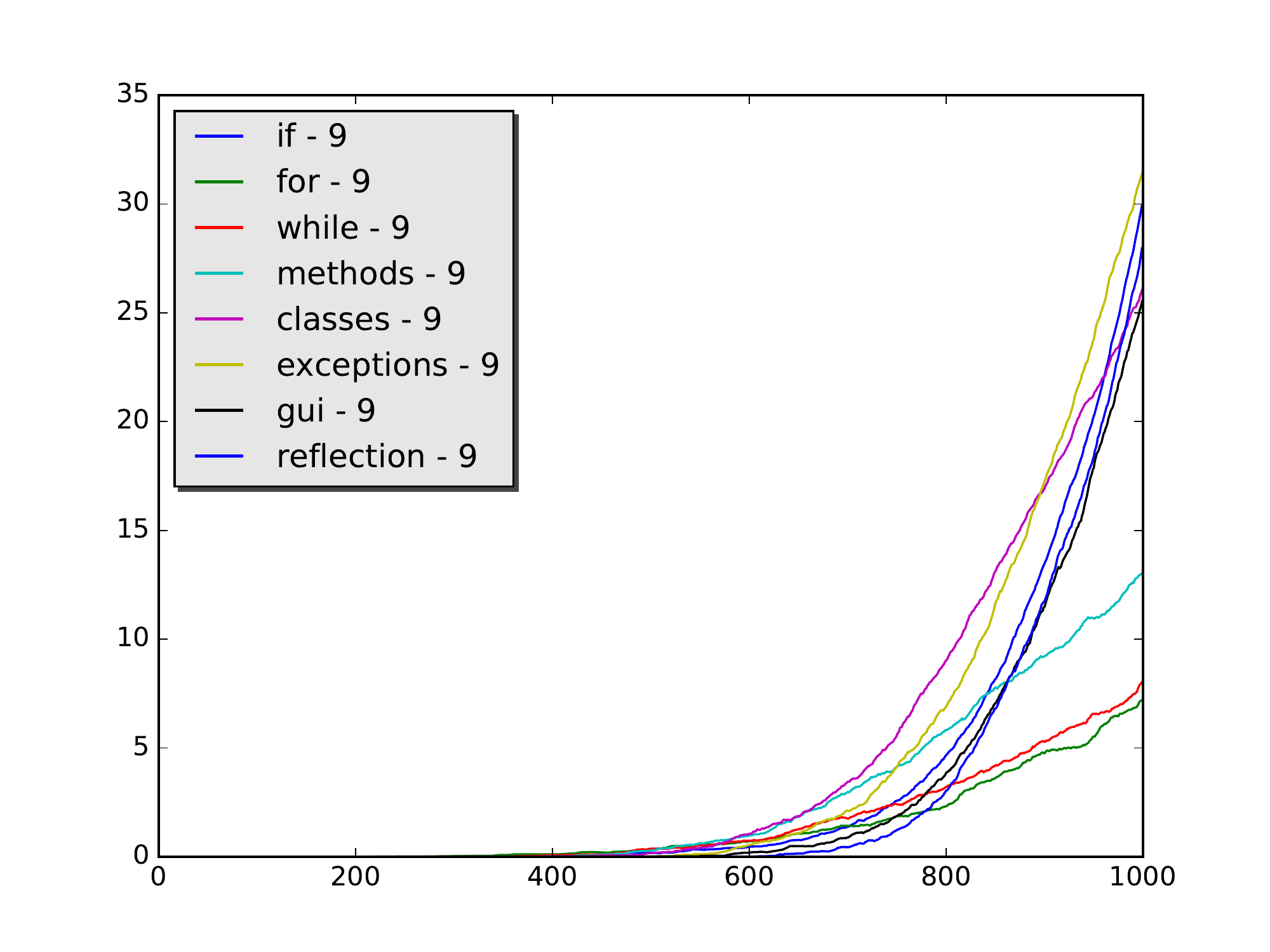} \\[-12pt]
  Level=6&
  \hspace{-1.6em} Level=7&
  \hspace{-1.6em} Level=8&
  \hspace{-1.6em} Level=9&
  \hspace{-1.6em} Level=10\\

\end{tabular}

\caption{Static Epsilon - Good Student}
\label{fig:Good-static-student-eps}
\end{figure*}
}
    
%
%

    \begin{center}
      \begin{tabular}{ | l | l | l | l  |}
        \hline
        Student    &  Static &  Static epsilon &  Change           \\ \hline
        Good   &  7.8    & \textbf{9.5}    &  ~+22\%   \\ \hline
        Bad    &  3.0    & \textbf{8.5}    &  ~+183\%  \\ \hline

    \end{tabular}
    \\
    \label{tbl:increase_vs}
    \tiny{A table comparing results form the two implementations}
    \end{center}

    Table ~\ref{tbl:increase_vs} proves the increase in efficiency compared to a static variant. This makes perfect sense. As one can see, the increase for the bad student is massive, this is explainable because the good student already had a good chance of accomplishing the tasks based on his good static percentage value. 

\newpage 
    \subsection{Case 3 - Dynamic epsilon greedy}
    
    \textbf{Parameters - 1000 students, 200 task-sets, 100 iterations}
    
    In this example, we use a high number of tasks and students to see how far the average student will get on each level. It is important to remember that due to the random state of the epsilon selection, the test will contain both good and bad students. Thus explaining why some will be able to reach a higher level than others.

    Figure \ref{fig:Dynamic} illustrates that students adapt and develop as each task is answered. Every student starts at skill level zero and moves towards higher difficulties. On average, each student gets a score higher than \textit{70\%} in each of the early topics. The curve-shape and smoothness changes throughout the different graphs, indicating several important factors. One thing worth noticing is the lack of smoothness throughout skill-level \textit{7-9}. These graphs are curvy and not as smooth as \textit{1-6}. The reason for this is the lack of data due to students struggling with earlier topics. Some students will use more time to progress, compared to others, thus resulting in only some making it to the top levels. This is clearly visible in the last graph \textit{9}.

{\renewcommand{\arraystretch}{2.0}
\setlength{\textfloatsep}{10pt plus 1.0pt minus 2.0pt}
\begin{figure*}[h!t!]
\begin{tabular}{*{3}{c}}
  \includegraphics[width=55mm]{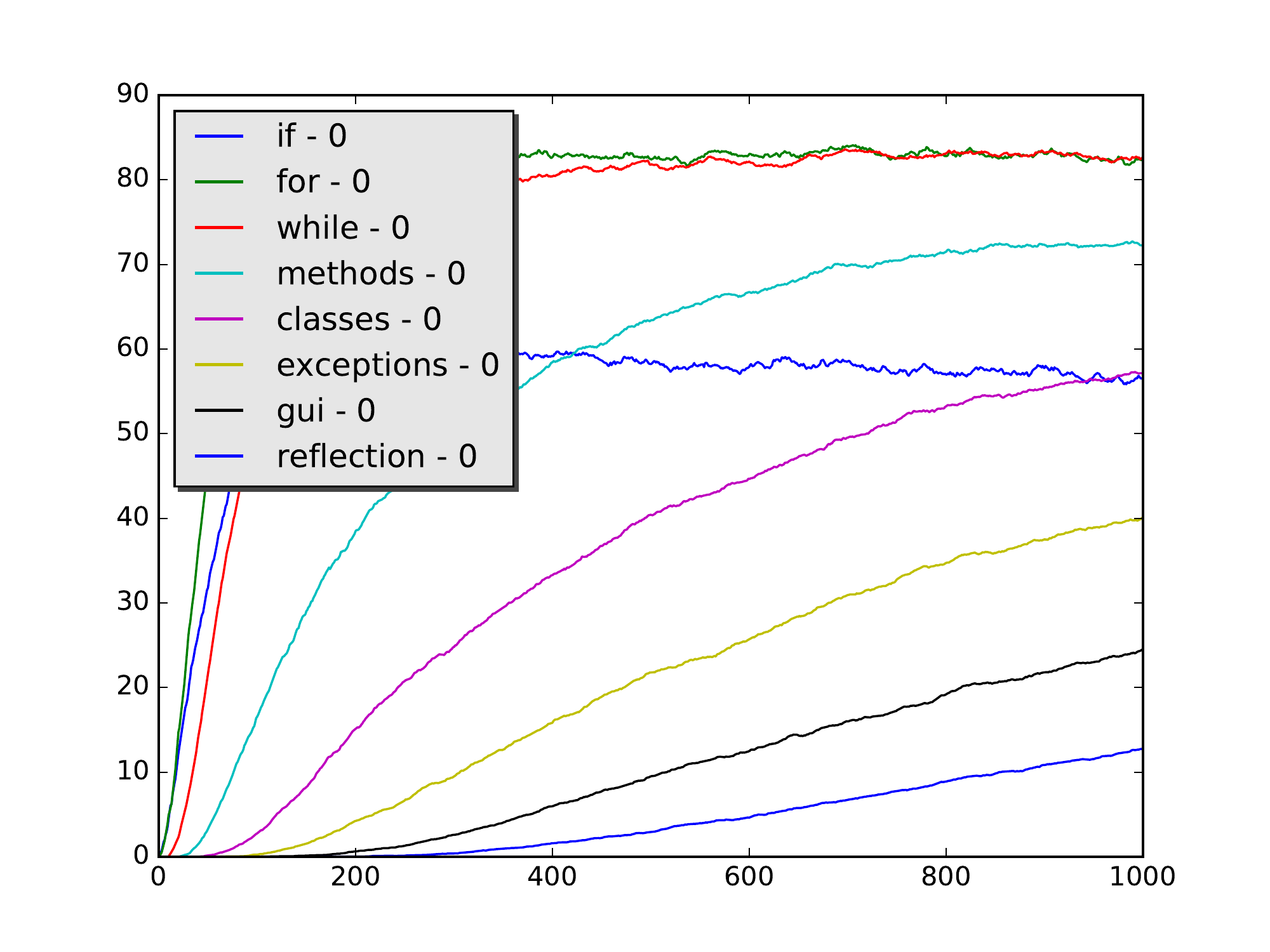} &
  \hspace{-1.6em}\includegraphics[width=55mm]{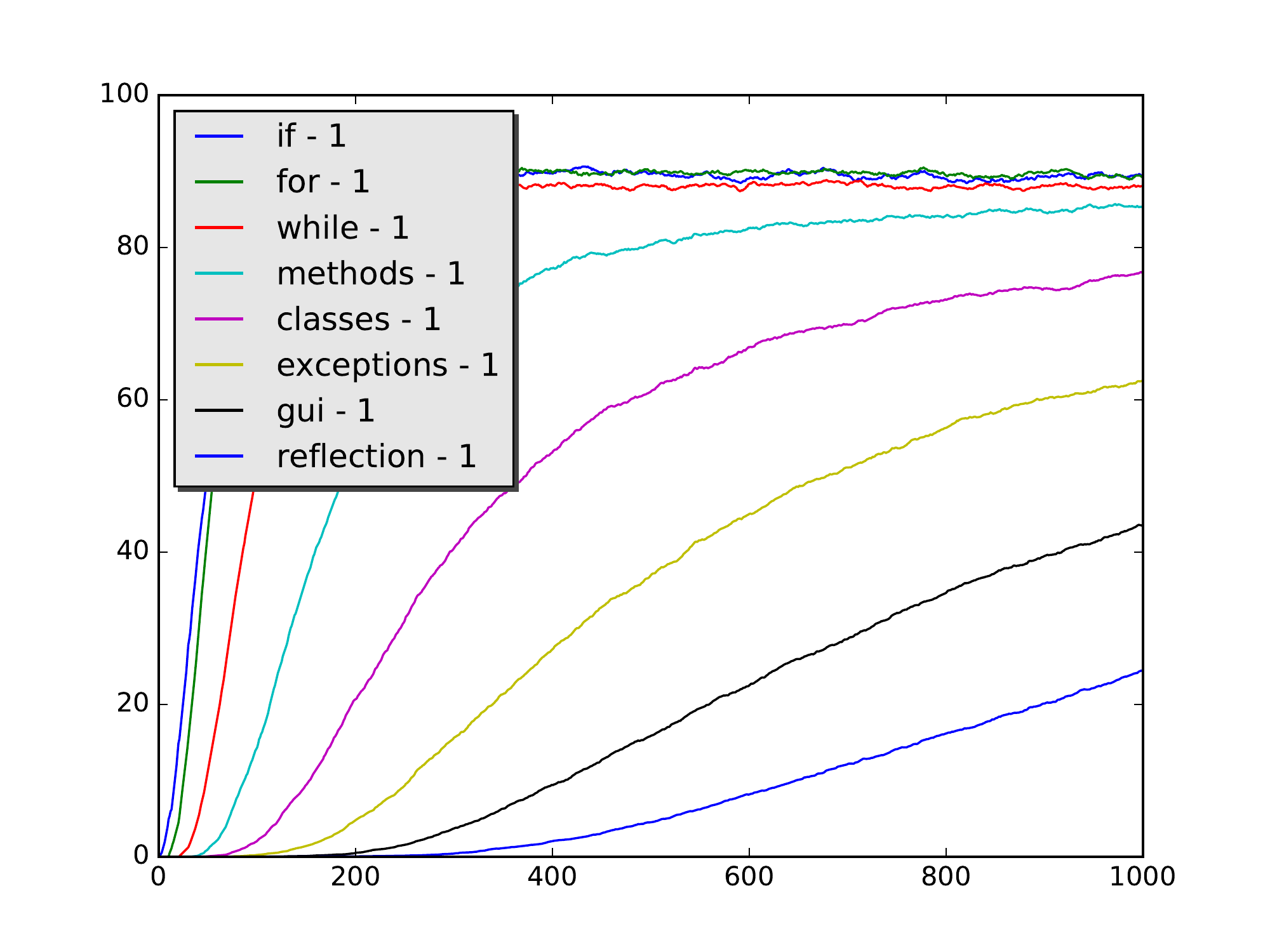} &
  \hspace{-1.6em}\includegraphics[width=55mm]{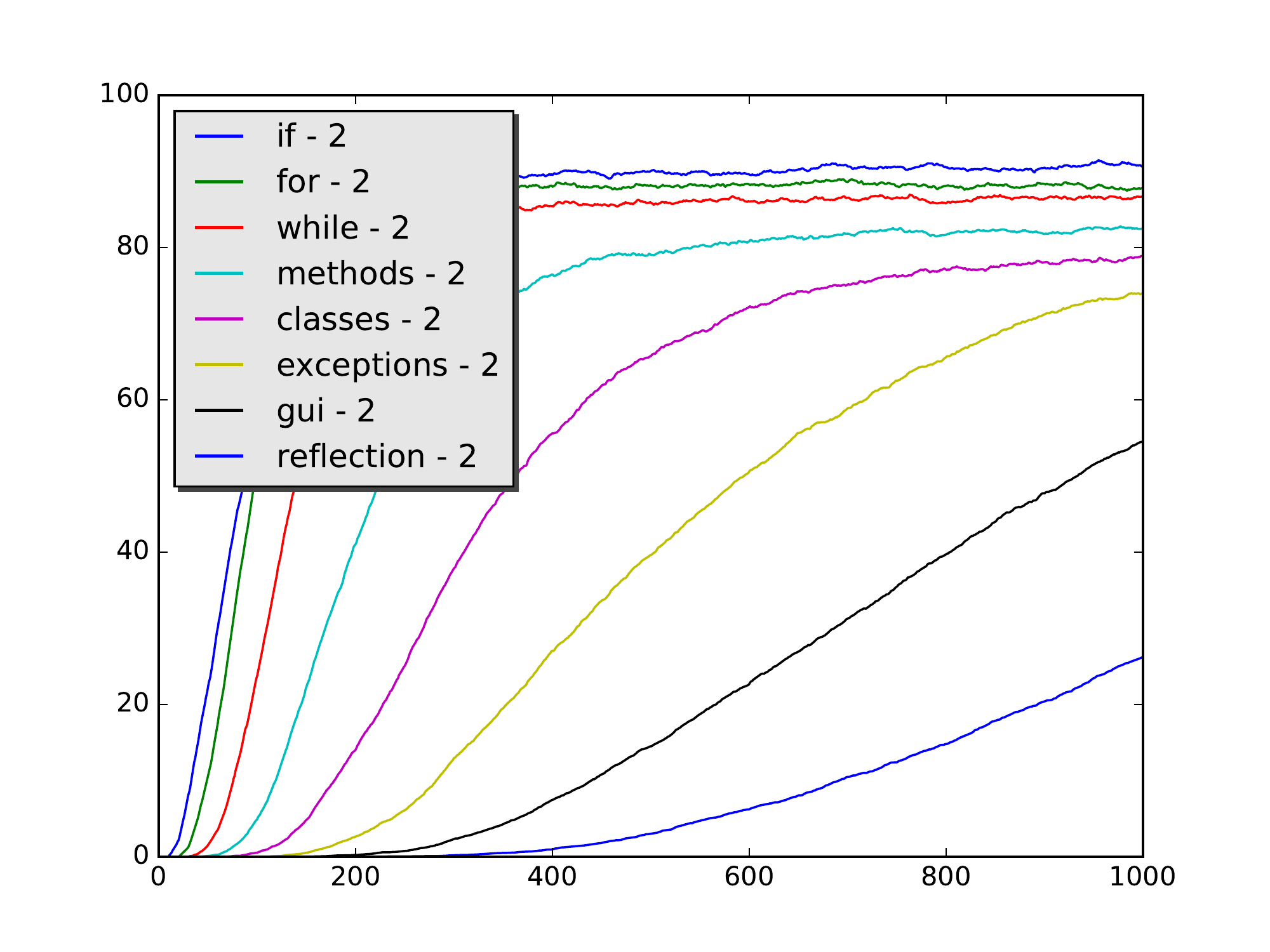} \\[-12pt]
  Level=1&
  \hspace{-1.6em} Level=2&
  \hspace{-1.6em} Level=3\\
  \includegraphics[width=55mm]{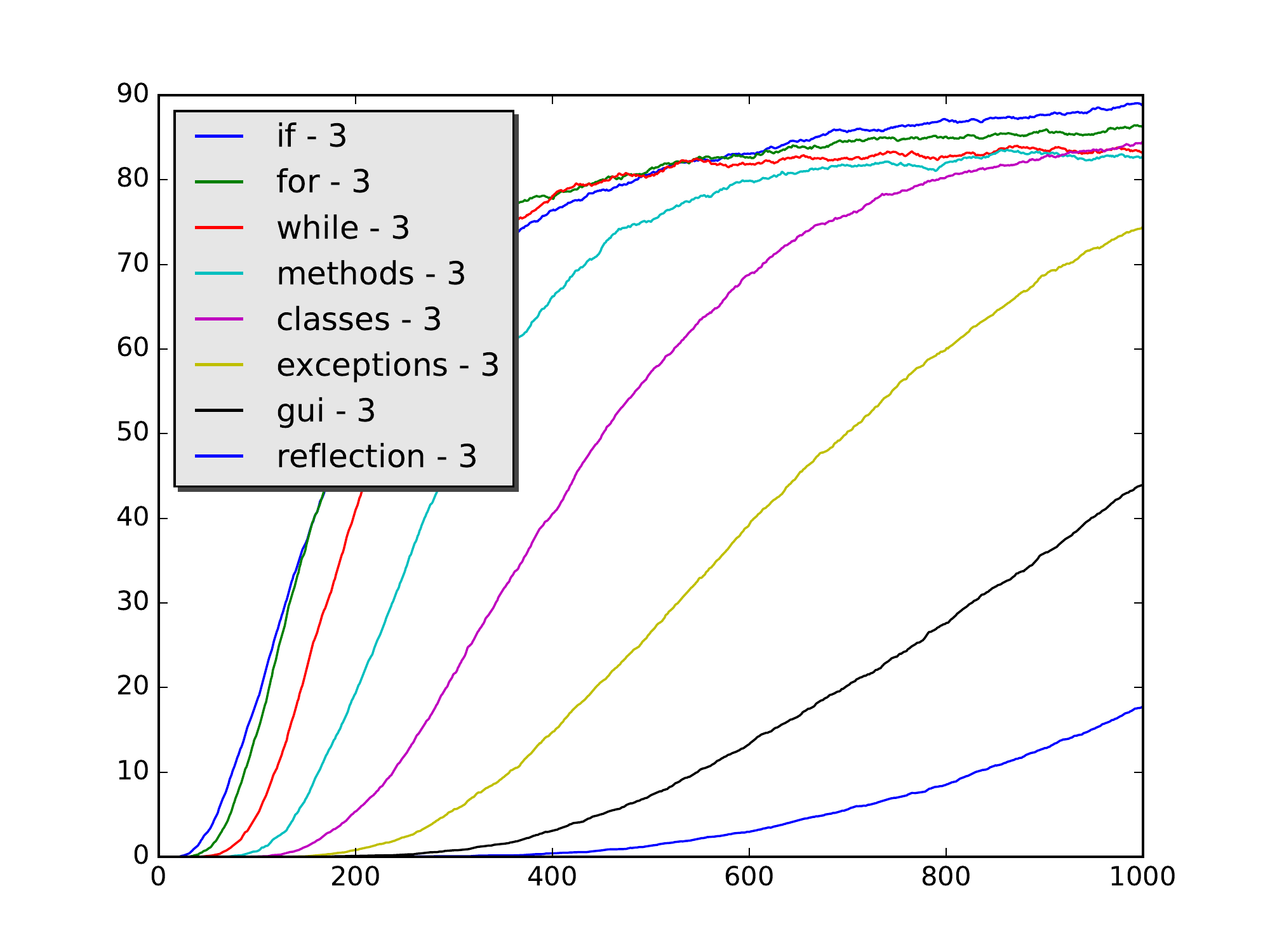} &
  \hspace{-1.6em}\includegraphics[width=55mm]{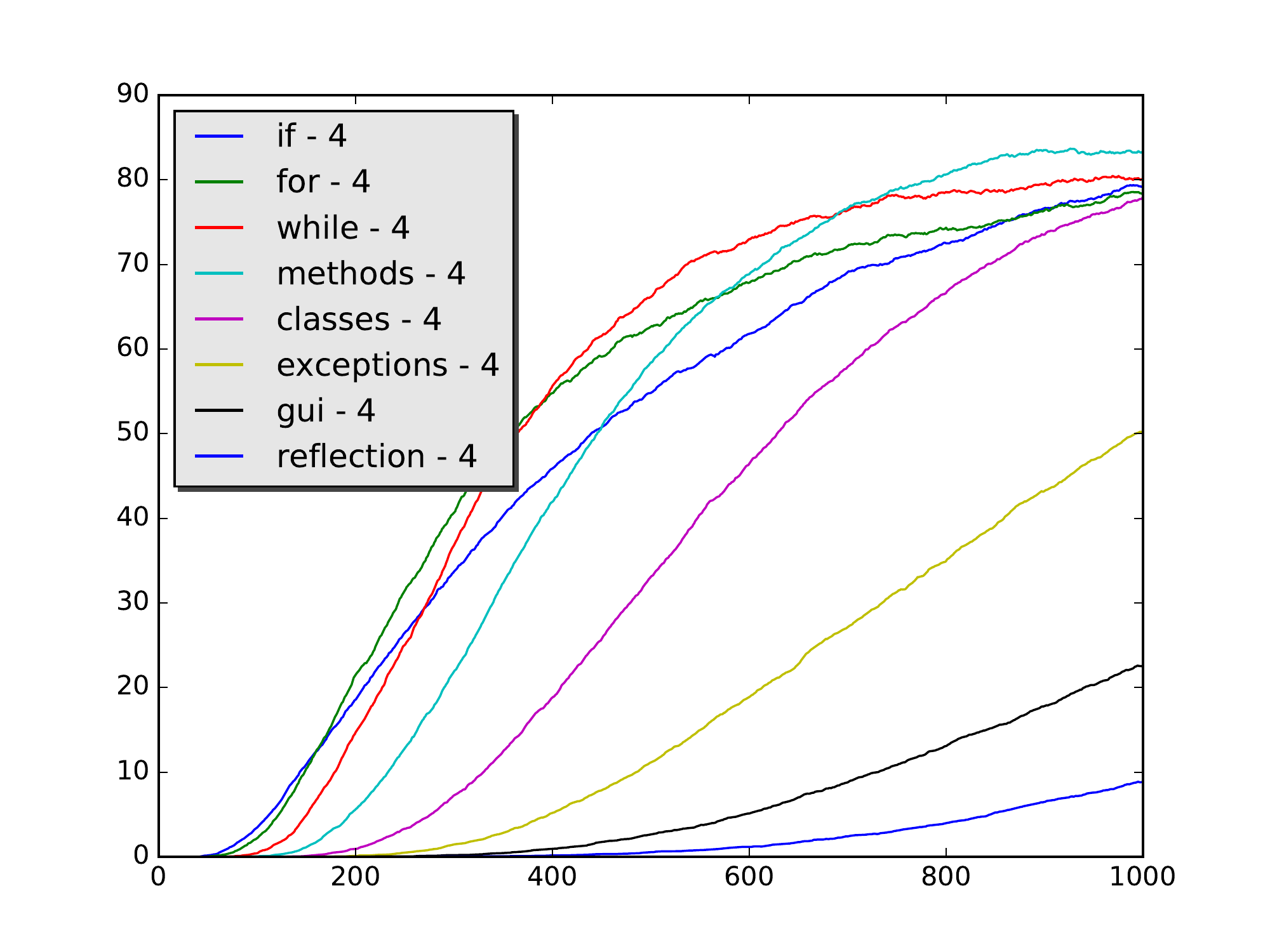} &
  \hspace{-1.6em}\includegraphics[width=55mm]{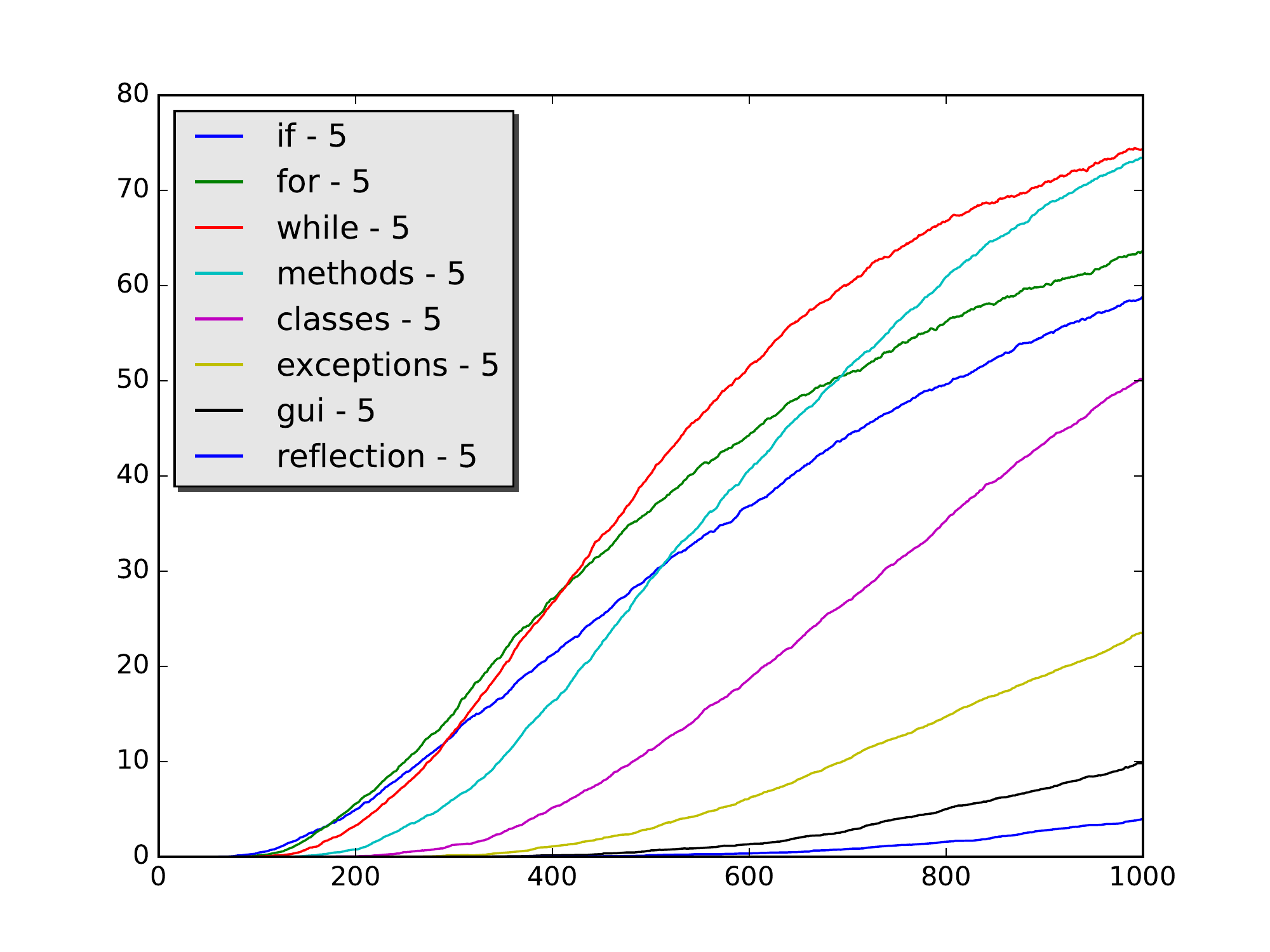} \\[-12pt]
  Level=4&
  \hspace{-1.6em} Level=5&
  \hspace{-1.6em} Level=6\\
  \includegraphics[width=55mm]{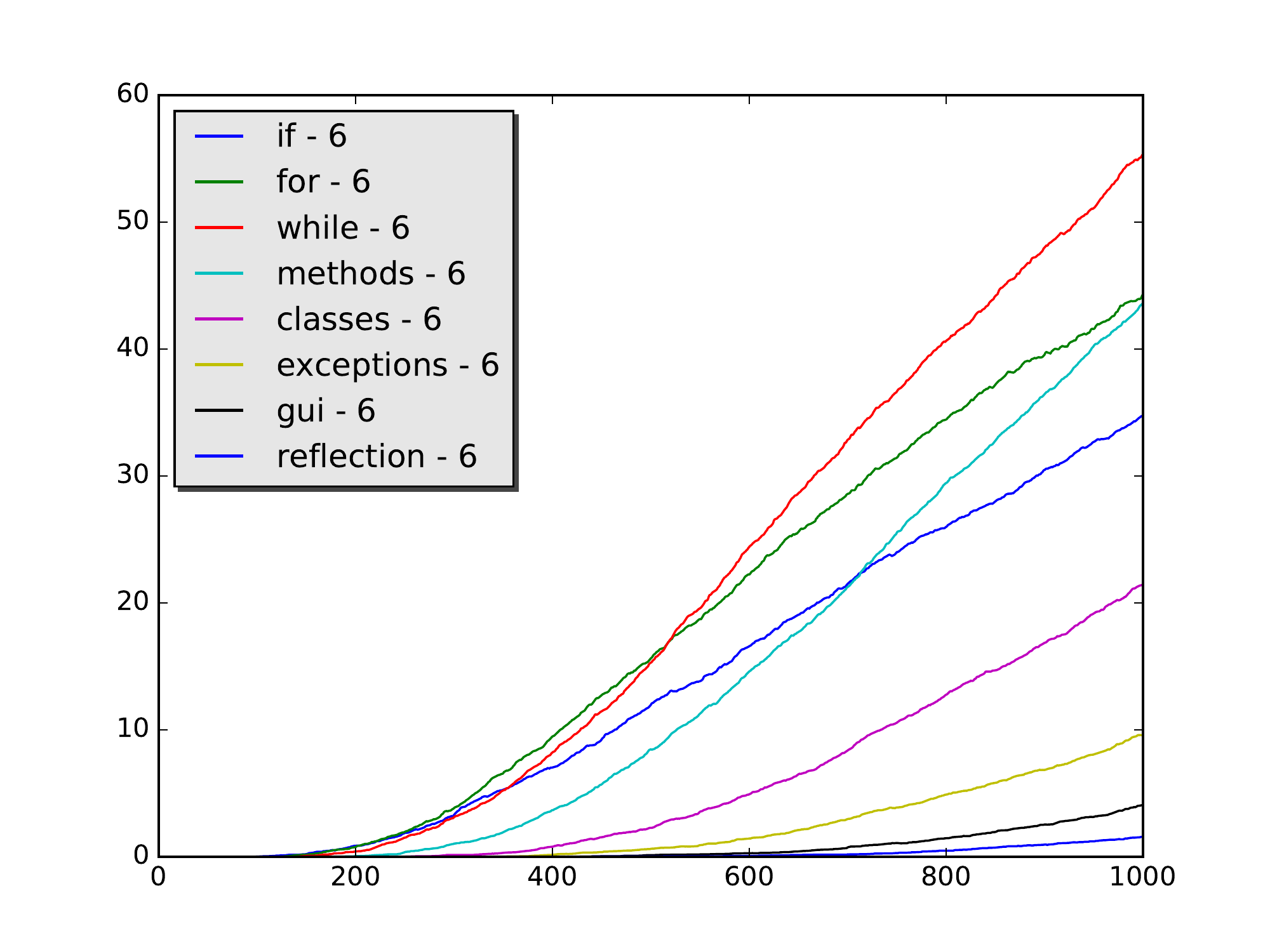} &
  \hspace{-1.6em}\includegraphics[width=55mm]{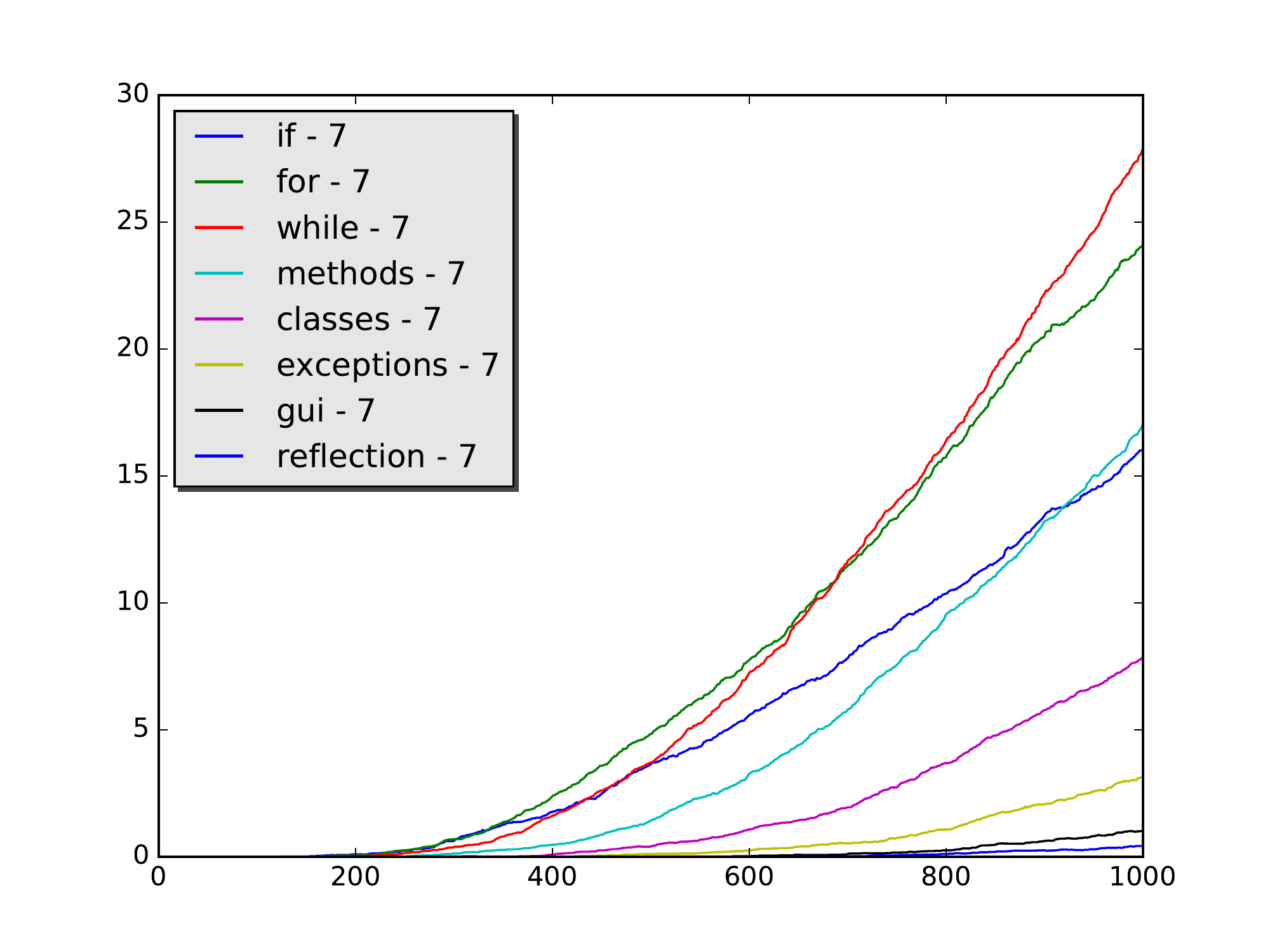} &
  \hspace{-1.6em}\includegraphics[width=55mm]{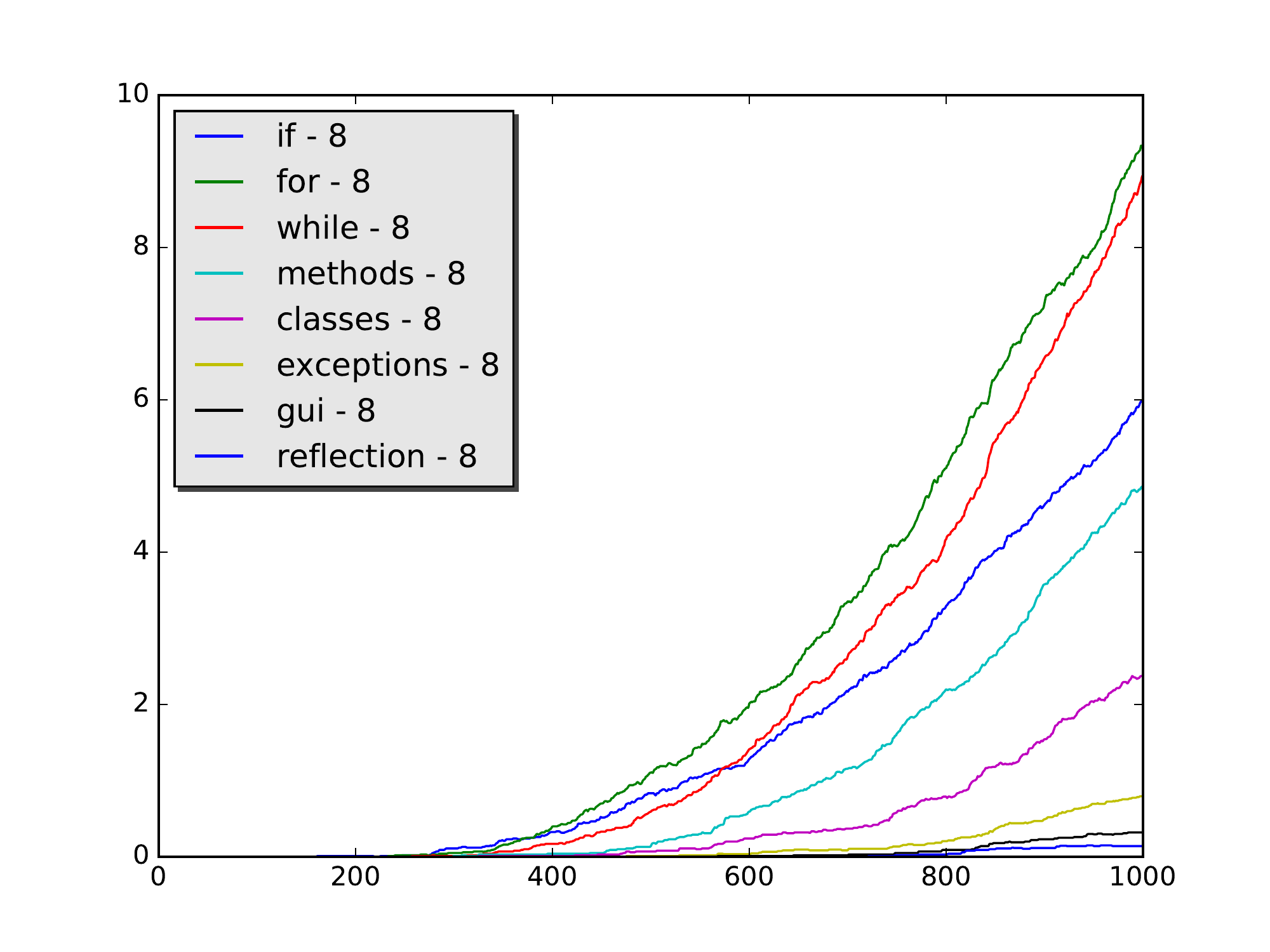} \\[-12pt]
  Level=7&
  \hspace{-1.6em} Level=8&
  \hspace{-1.6em} Level=9\\
  \includegraphics[width=55mm]{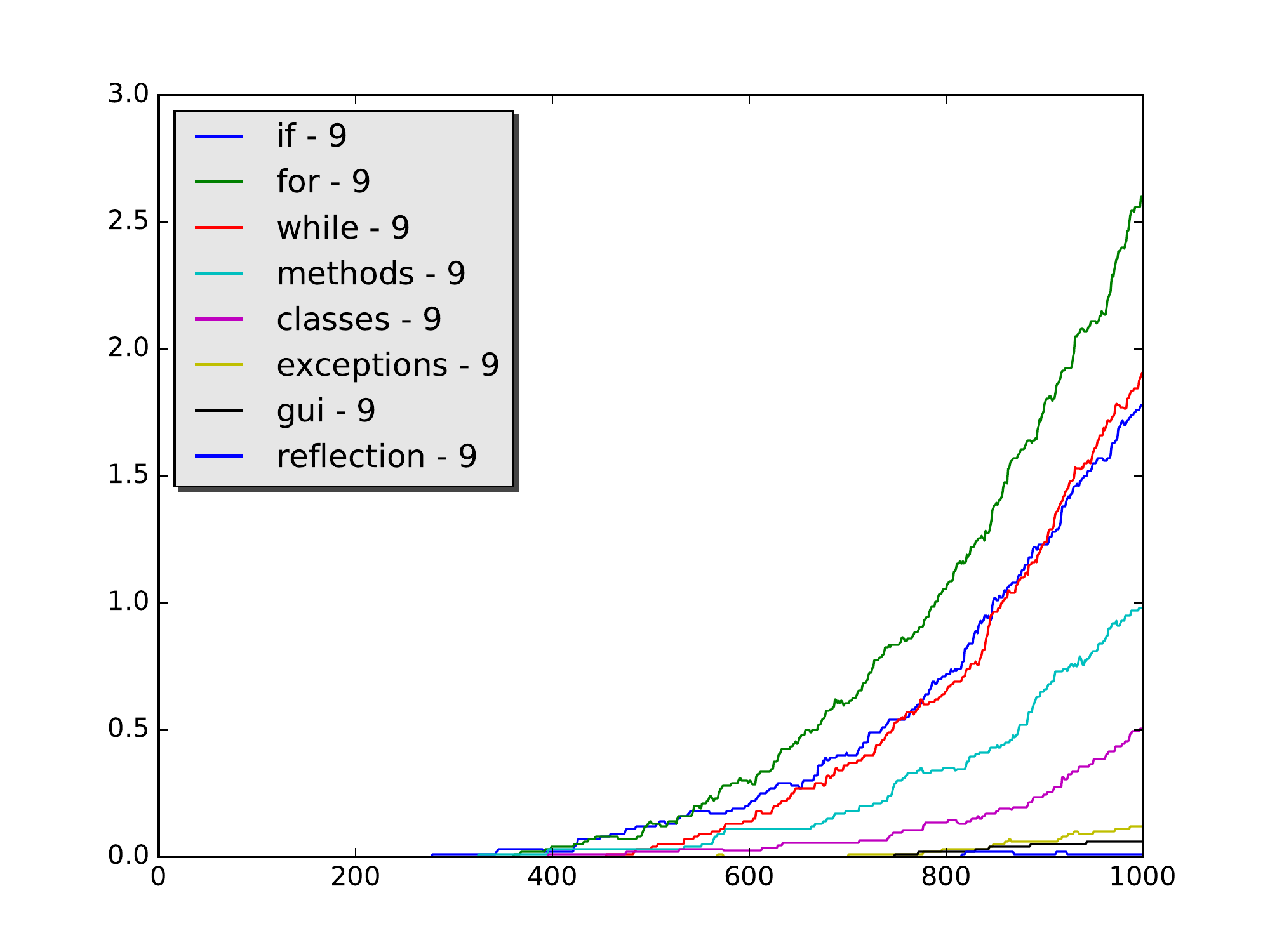} \\[-12pt]
  Level=10\\

\end{tabular}
            \caption{Dynamic - Neutral student}
            \label{fig:Dynamic}
\end{figure*}
}

%
%

    Another interesting fact is the different slopes. As expected, most students pass the more easy levels with a fairly steep graph, indicating a quick learning rate. Through level \textit{3-5} the graph turns into a fairly more moderate version, and at 6 through 9 we experience the slowest growth as expected, since these are the most challenging tasks. 
    
    The graphs also show when the average student started at each level. In our example the numbers would be :

    \begin{center}
      \begin{tabular}{ | l | l |}
        \hline
       Level 1 & 0-100    \\ \hline
       Level 2 & 0-100    \\ \hline
       Level 3 & 0-200    \\ \hline
       Level 4 & 50-300   \\ \hline
       Level 5 & 50-350   \\ \hline
       Level 6 & 150-450  \\ \hline  
       Level 7 & 250-450  \\ \hline
       Level 8 & 350-500* \\ \hline
       Level 9 & 450-750* \\ \hline
       Level 10 & 450-750* \\ \hline
    \end{tabular}
    \\
    \tiny{* Limited amount of students}
    \end{center}

    As shown by this table, almost every student has started the last levels after 500 tasks. After this, all time used, is at progressing in the last tiers of skill-levels. We think this is appropriate, because these tasks are made to be the most challenging. All this can be adapted to a specific case, with the altering of parameters described earlier in the text.

   \begin{center}
      \begin{tabular}{ | l | l | l | l | l |}
        \hline
        Student    &  Static &  Static epsilon &  Dynamic    &     Change        \\ \hline
        Neutral    &  7.8    &    9.5          &    \textbf{6.7}      & ~-30\%*   \\ \hline
         
    \end{tabular}
    \label{tbl:dynamic-table}
    \\
    \tiny{*Compared to Static Epsilon}
    \end{center}
    
    As seen in table ~\ref{tbl:dynamic-table}, the total skill level significantly decreases while using the dynamic epsilon approach. This is because the student model accounts for all types of students, good and bad. In addition to  this, there are no set percentages thus making this model autonomous. This is our attempt to implement a real-world scenario student model. It is however not expected that the implemented student model should perform better than a static high value percentage based model.

    \subsection{Future Improvements}
    
    
    
    \textbf{Decay} is a method where a task gets a decreased probability when a task is selected during the selection phase of the algorithm. This is believed to greatly improve the selection phase of the algorithm as it would lower the chance of getting recurring tasks.
    Decay is already implemented, but current implementation punish the selected cell significantly more than initially planned. This results in a selected cell to only be chosen once, and makes the student progress less than it is expected of the algorithm to achieve.
    
    The challenge here is to find a good formula to punish the selected cell and reward the surrounding cells without noticeably obscuring the skill matrix.
    
    \textbf{Graph based}: As a future direction, we propose to test out current principles in a graphed environment instead of a matrix. This would allow for more advanced relationships between the topics. The principles from the matrix version of the algorithm would easily be adopted to a graph based one.
    Using skill levels on edges makes it possible to determine which cells to reward and punish. 

    \section{Conclusion}
    
    
    In this paper, we tackled the problem of personalized task assignment in online learning environment. We proposed two algorithms for enhancing ITS system which are inspired by the \textit{MAB} problem, namely one algorithm for understanding student behavior and in turn giving appropriate tasks, and another algorithm for simulating the complexity of student's behaviour. Our paradigm involves creating a random walk on a grid in order to choose the assignments.
    
    
    Our solution is believed to be ready for practical trials in a classroom setting. This would give valuable test and validation data for further research.

   \bibliographystyle{abbrv}
\bibliography{biblo}

\begin{thebibliography}{10}

\bibitem{barr2015advice}
V.~Barr and M.~Guzdial.
\newblock Advice on teaching cs, and the learnability of programming languages.
\newblock {\em Communications of the ACM}, 58(3):8--9, 2015.

\bibitem{aibook}
R.~E. Bellman et~al.
\newblock {\em An introduction to artificial intelligence: Can computers
  think?}
\newblock Boyd \& Fraser Publishing Company, 1978.

\bibitem{mabits}
P.-Y.~O. Benjamin~Clement, Didier~Roy.
\newblock Multi-armed bandits for intelligent tutoring systems.
\newblock {\em Journal of Educational Data Mining}, 7(2), 2015.

\bibitem{bloom19842}
B.~S. Bloom.
\newblock The 2 sigma problem: The search for methods of group instruction as
  effective as one-to-one tutoring.
\newblock {\em Educational researcher}, 13(6):4--16, 1984.

\bibitem{umap}
P.~Brusilovsky and E.~Mill{\'a}n.
\newblock User models for adaptive hypermedia and adaptive educational systems.
\newblock In {\em The adaptive web}, pages 3--53. Springer-Verlag, 2007.

\bibitem{oopt}
B.~Clement, D.~Roy, P.-Y. Oudeyer, and M.~Lopes.
\newblock Online optimization of teaching sequences with multi-armed bandits.
\newblock In {\em 7th International Conference on Educational Data Mining},
  2014.

\bibitem{goodwin2015educating}
M.~Goodwin, C.~Auby, R.~Andersen, and V.~Barstad.
\newblock Educating programming students for the industry.
\newblock {\em Digital Media in Teaching and its Added Value}, page 100, 2015.

\bibitem{goodwin2016teaching}
M.~Goodwin and T.~Drange.
\newblock Teaching programming to large student groups through test driven
  development.
\newblock In {\em the International Conference on Computer Supported Education
  (CSEDU)}, 2015.

\bibitem{jacobs2015experiences}
C.~T. Jacobs, G.~J. Gorman, and L.~Craig.
\newblock Experiences with efficient methodologies for teaching computer
  programming to geoscientists.
\newblock {\em arXiv preprint arXiv:1505.05425}, 2015.

\bibitem{askelle}
J.~Jeuring, A.~Gerdes, and B.~Heeren.
\newblock Ask-elle: A haskell tutor.
\newblock In {\em 21st Century Learning for 21st Century Skills}, pages
  453--458. Springer, 2012.

\bibitem{iism}
J.~I. Lee and E.~Brunskill.
\newblock The impact on individualizing student models on necessary practice
  opportunities.
\newblock {\em International Educational Data Mining Society}, 2012.

\bibitem{smtl}
L.~S. Lumsden.
\newblock Student motivation to learn. eric digest, number 92.
\newblock 1994.

\bibitem{plins}
D.~Miliband.
\newblock Personalised learning: building a new relationship with schools.
\newblock In {\em Speech by the Minister of State for School Standards to the
  North of England Education Conference}, 2004.

\bibitem{baynet}
E.~Mill{\'a}n, T.~Loboda, and J.~L. P{\'e}rez-de-la Cruz.
\newblock Bayesian networks for student model engineering.
\newblock {\em Computers \& Education}, 55(4):1663--1683, 2010.

\bibitem{milne2002difficulties}
I.~Milne and G.~Rowe.
\newblock Difficulties in learning and teaching programming views of students
  and tutors.
\newblock {\em Education and Information technologies}, 7(1):55--66, 2002.

\bibitem{aits}
R.~Nkambou, R.~Mizoguchi, and J.~Bourdeau.
\newblock {\em Advances in intelligent tutoring systems}, volume 308.
\newblock Springer Science \& Business Media, 2010.

\bibitem{bktbook}
N.~T. Pardos, Z. A.~Heffernan.
\newblock {\em Modeling Individualization in a Bayesian Networks IMplementation
  of Knowledge Tracing}.
\newblock Springer, 2010.

\bibitem{itsbook}
J.~Psotka, L.~D. Massey, and S.~A. Mutter.
\newblock {\em Intelligent tutoring systems: Lessons learned}.
\newblock Psychology Press, 1988.

\bibitem{schulte2006teachers}
C.~Schulte and J.~Bennedsen.
\newblock What do teachers teach in introductory programming?
\newblock In {\em Proceedings of the second international workshop on Computing
  education research}, pages 17--28. ACM, 2006.

\bibitem{itsold}
D.~Sleeman and J.~S. Brown.
\newblock {\em Intelligent tutoring systems}.
\newblock London: Academic Press, 1982.

\bibitem{ibktm}
M.~V. Yudelson, K.~R. Koedinger, and G.~J. Gordon.
\newblock Individualized bayesian knowledge tracing models.
\newblock In {\em Artificial intelligence in education}, pages 171--180.
  Springer, 2013.

\bibitem{alen}
R.~Zatarain~Cabada, M.~L. Barron~Estrada, F.~Gonzalez~Hernandez, and
  R.~Oramas~Bustillos.
\newblock An affective learning environment for java.
\newblock In {\em 2015 International Conference on Advanced Learning
  Technologies (ICALT)}, pages 350--354. IEEE, 2015.

\end{thebibliography}

\end{document}